\documentclass{article}

 \usepackage[preprint, nonatbib]{neurips_2026}

\usepackage[utf8]{inputenc} %
\usepackage[T1]{fontenc}    %
\usepackage{hyperref}       %
\usepackage{url}            %
\usepackage{booktabs}       %
\usepackage{amsfonts}       %
\usepackage{nicefrac}       %
\usepackage{microtype}      %
\usepackage{xcolor}         %
\usepackage{enumitem}
\usepackage{wrapfig}
\usepackage{siunitx}
\usepackage{arydshln}
\usepackage{frcursive}
\usepackage{arydshln}
\usepackage{booktabs}
\usepackage{amssymb}
\usepackage{pifont}

\newcommand{\cmark}{\ding{51}} %
\newcommand{\xmark}{\ding{55}} %
\definecolor{mygreen}{RGB}{0,128,0}
\definecolor{myred}{RGB}{200,0,0}
\definecolor{myyellow}{RGB}{204, 153, 0}

\usepackage{booktabs, makecell, tabularx}
\usepackage{multirow}
\usepackage{duckuments}
\usepackage{amsmath}
\usepackage{graphicx}
\usepackage{caption} %
\usepackage{subcaption} %
\usepackage{arydshln}
\usepackage{mathtools}
\usepackage{cleveref}
\usepackage{tcolorbox}
\usepackage{fvextra}
\tcbuselibrary{breakable, skins}
\usepackage{soul}
\definecolor{MyDarkBlue}{rgb}{0,0.08,1}
\definecolor{MyDarkGreen}{rgb}{0.02,0.6,0.02}
\definecolor{MyDarkRed}{rgb}{0.8,0.02,0.02}
\definecolor{MyDarkOrange}{rgb}{0.40,0.2,0.02}
\definecolor{MyPurple}{RGB}{111,0,255}
\definecolor{MyRed}{rgb}{1.0,0.0,0.0}
\definecolor{MyGold}{rgb}{0.75,0.6,0.12}
\definecolor{MyDarkgray}{rgb}{0.66, 0.66, 0.66}
\definecolor{MyDarkCyan}{rgb}{0.05, 0.55, 0.45}
\definecolor{MyBlack}{rgb}{0., 0., 0.}
\definecolor{MyMagenta}{rgb}{1., 0., 1.}
\definecolor{BerkeleyYellow}{RGB}{255,204,41}
\definecolor{BerkeleyLightBlue}{RGB}{94,146,221}
\definecolor{BkDarkBlue}{rgb}{.05,.07,.353}

\newcommand{\updated}[1]{{#1}}
\newcommand{\arxiv}[1]{{#1}}

\newcommand{\myparagraph}[1]{\vspace{0pt} \noindent \textbf{#1}}

\newcommand{\etal}{\textit{et al.}}
\newcommand{\ignorethis}[1]{}

\newcommand*{\menlo}{\fontfamily{lmtt}\fontsize{9}{9}\selectfont }
\title{The Many Senses of Visual Similarity: \\ A Text-Prompted Image Perceptual Metric}

\author{Sheng-Yu Wang$^{1}$\hspace{2mm} Yotam Nitzan$^{2}$ \hspace{2mm}  Aaron Hertzmann$^{2}$\hspace{2mm} Jun-Yan Zhu$^{1}$ \\  \textbf{Eli Shechtman$^{2}$\hspace{2mm}Alexei A. Efros$^{3}$\hspace{2mm} Richard Zhang$^{2}$} \\
$^{1}$Carnegie Mellon University \hspace{5mm} $^{2}$Adobe Research \hspace{5mm} $^{3}$UC Berkeley}

\begin{document}

\maketitle

\begin{abstract}
Human visual similarity judgments are context-dependent. For example, two images may be similar in shape but distinct in color. Existing perceptual similarity metrics, however, collapse these nuances into a single scalar value, offering no mechanism to condition on specific aspects. To bridge this gap, we introduce a large-scale dataset of human similarity judgments over image triplets, where each triplet is annotated across multiple, free-form semantic aspects of similarity. Benchmarking a broad range of frontier vision-language models (VLMs) reveals a considerable performance gap compared to human annotators' consensus. Leveraging our data, we fine-tune a VLM to produce our Text-Prompted Image Perceptual Similarity (TPIPS) metric, capturing multiple senses of visual similarity depending on the specified text prompt. 
We demonstrate that TPIPS aligns more closely with human perception and generalizes reliably beyond the training distribution. Finally, we show that TPIPS unlocks new capabilities in text-guided retrieval, compositional search, and the fine-grained evaluation of generative models. Our code, data, and trained models are at \url{https://peterwang512.github.io/TPIPS}.
\end{abstract}

\section{Introduction}

\begin{figure}[t]
    \centering
    \includegraphics[width=1.0\linewidth]{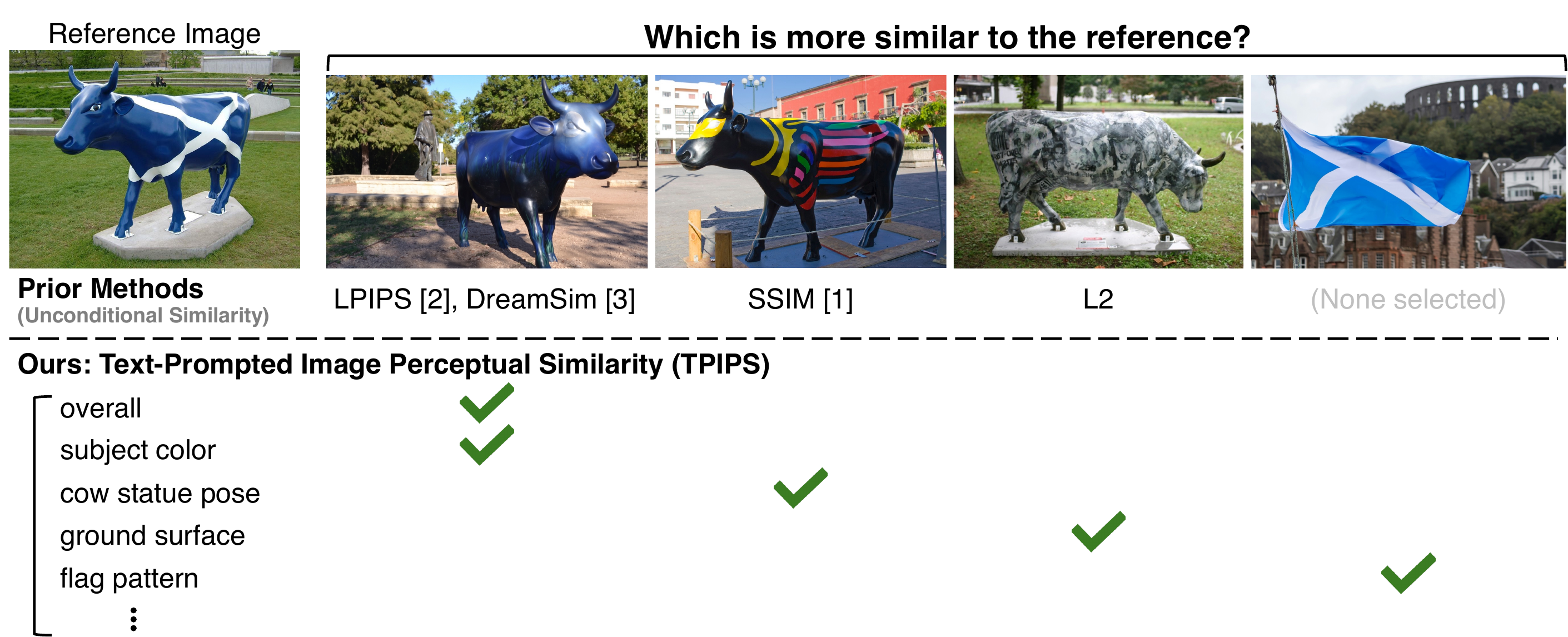}
    \caption{\textbf{The Many Senses of Similarity.} Given a reference image (top left), the four candidate images (top right) are similar to the reference in different ways, such as subject color, pose, ground surface, or patterns.  But existing visual similarity measures (L2, SSIM~\cite{wang2004ssim}, LPIPS~\cite{zhang2018lpips}, DreamSim~\cite{fu2023dreamsim}) collapse similarity into a single scalar, and thus producing ambiguous and inconsistent results. 
    Our \textit{Text-Prompted Image Perceptual Similarity (TPIPS)} \arxiv{is a VLM fine-tuned to assess pairwise similarity for specific visual aspects using free-form text prompts.}
    With conditioning, this visual similarity problem is no longer ambiguous, and our method selects the most similar image aligned with human perceptual judgements.}
    \label{fig:teaser}
\end{figure}

A long line of psychology research has established that human similarity judgments are context-dependent~\cite{tversky1977features, medin1993respects}. The same holds for \arxiv{comparing two} images: there are many senses of visual similarity, depending on which aspect (color, texture, illumination, semantics, style, etc.) is being compared. \updated{Consider Figure~\ref{fig:teaser}: given 
the reference cow statue image (left), it is not immediately obvious 
which of the four candidate images on the right is the most similar overall to the reference. However, the answer becomes clear if the similarity judgement is \textit{conditioned on a specific visual aspect}, such as the cow's color, pose, or the image background.}

The image processing and computer vision communities have spent decades making progress on capturing human perceptual similarity, from hand-engineered distances like SSIM and others~\cite{wang2004ssim, zhang2011fsim, mantiuk2011hdrvdp2} to learned metrics trained on annotated human similarity judgments, such as LPIPS~\cite{zhang2018lpips} or DreamSim~\cite{fu2023dreamsim}. These methods, however, model similarity as a single, scalar quantity, without conditioning on the aspect of the comparison.
While previous methods add a notion of conditioning by leveraging vision-language models~\cite{vaze2023genecis, hsieh2025focallens, chen2025omniattribute},
to our knowledge, no dataset collected with human perceptual similarity judgments exists to serve as validation.

To fill this gap, we collect a large-scale dataset of one million human similarity judgments over 25K image triplets.  Following triplet-based protocols standard in perceptual similarity studies~\cite{zhang2018lpips, fu2023dreamsim, hebart2019things}, annotators provide relative perceptual judgements by comparing three images and indicating which are most or least similar. Different from prior work, we extend this protocol to perceptual judgements conditioned on various visual aspects. We collect 260K triplet-aspect condition combinations, where each triplet is annotated conditioned on multiple \textit{free-form} visual aspects, each aspect yielding a separate set of similarity judgments.  The triplets are generated using text-to-image models and designed to be challenging. If the differences between candidates are too subtle, humans cannot reliably judge similarity; if they are too coarse, the task collapses to judging overall similarity, which current models already handle. We target the middle ground, with images that differ along multiple aspects simultaneously --- through fine-grained variations in properties such as color, lighting, and texture.

We use this synthetic dataset to construct a benchmark and evaluate a broad set of baselines, including recent open-source and proprietary VLMs~\cite{openai2023gpt4v, geminiteam2023gemini, bai2025qwen3vl, chen2024internvl} and state-of-the-art multimodal embedding models~\cite{jiang2025vlm2vec, li2026qwen3vlembedding}, revealing room for improvement.
We find that even the strongest current VLM systems often do not agree with human similarity judgments on specified aspects.
\arxiv{To address this,} we carefully propose and analyze different architecture choices for tuning a VLM into an aspect-conditioned perceptual metric on our collected perceptual data.
Our model, \textit{Text-Prompted Image Perceptual Similarity} (TPIPS), narrows the performance gap considerably. \arxiv{To test whether our method generalizes beyond the training distribution, }
we collect a second set of human similarity judgments over outputs generated by \arxiv{17 unseen} external algorithms, designed for four different vision and graphics tasks, spanning both generative and non-generative pipelines. \arxiv{In addition, we test our method on other unconditional perceptual similarity and text-conditional retrieval benchmarks.} Our model outperforms baselines in these settings too, suggesting that it can serve as a general-purpose perceptual metric for \arxiv{overall perceptual judgement, conditional retrieval, and evaluating vision and graphics algorithm performance.}

\arxiv{In summary, our contributions are (1) a novel large-scale dataset of text-conditioned image perceptual judgements, (2) extensive benchmarking against existing methods, including frontier VLMs, showing our method outperforms them by fine-tuning a VLM on our dataset, and (3) showing} several downstream applications: evaluation of generative vision models, where output and reference similarity is measured along specified visual axes, nearest-neighbor retrieval, where the same query image returns different neighbors under different aspects, and compositional retrieval, where multiple query images, each conditioned on a different visual aspect, can be combined to find images that match the criteria. Our code, data, and trained models are at \url{https://peterwang512.github.io/TPIPS}.

\section{Related Work}

\myparagraph{Perceptual similarity and preferences.}
Comparing two images is a fundamental problem in vision, with early approaches relying on hand-engineered, patch-based metrics~\cite{wang2004ssim,zhang2011fsim,mantiuk2011hdrvdp2}. With advances in deep image recognition~\cite{krizhevsky2012alexnet, simonyan2015vgg, he2016resnet, dosovitskiy2021vit, radford2021clip, oquab2024dinov2}, learned deep features have been found to provide a more useful similarity signal than hand-engineered descriptors~\cite{zhang2018lpips,johnson2016perceptualloss, gatys2016styletransfer,amir2021understanding,sargent2026vlic}. LPIPS~\cite{zhang2018lpips}, PieAPP~\cite{prashnani2018pieapp}, and DISTS~\cite{ding2020dists} directly optimize deep features to align better with human judgments on low-level distortions. DreamSim~\cite{fu2023dreamsim} learns human perceptual judgments on synthesized images with high-level variations. Muttenthaler et al.~\cite{muttenthaler2023humanalignment} study perceptual judgments across object categories with THINGS~\cite{hebart2019things}. Other works improve metric robustness to spatial shifts~\cite{ghildyal2022shifttolerantlpips}, adversarial perturbations~\cite{ghazanfari2024lipsim}, and arbitrary transformations~\cite{kettunen2019elpips}. \updated{Alternative metrics isolate specific attributes like style~\cite{garces2014illustrationsim,somepalli2024csd} or relationships between visual elements~\cite{nguyen2026relational}.} While these methods predict a single similarity score, similarity is inherently multi-faceted~\cite{tversky1977features}. We address this by introducing a benchmark and model for human similarity judgements conditioned on various visual aspects. A separate but adjacent line of work uses pairwise human comparisons over generated or edited images to learn a scalar quality score over a single image~\cite{xu2023imagereward, kirstain2023pickapic, wu2023human, wu2026editreward}. In contrast, we learn from triplet judgments to model pairwise similarity conditioned on visual aspects.

\myparagraph{Conditional representation.}
Various research has studied learning representations of an input conditioned on different aspects. A line of work explores learning conditional representation given an explicit aspect, but assumes a fixed, finite vocabulary of conditions~\cite{amid2015mvte,veit2017csn,vasileva2018typeaware,thong2019cooperative,tan2019learningsimilarity}. More recent work generalizes the condition input to free-form text, including training-free methods via vision-language models~\cite{kawarada2025dior, li2025promptableembeddings,wei2026qare} and training-based methods that curate conditional supervision. GeneCIS~\cite{vaze2023genecis} and FocalLens~\cite{hsieh2025focallens} mine attribute-difference triplets from existing vision-language datasets, while Omni-Attribute~\cite{chen2025omniattribute} bootstraps supervision from VLM-generated labels and TF-QARE~\cite{wei2026qare} prompts for an image's object, background, or style and averages the embedding of the resulting description. A parallel line of work studies context-conditional categorization~\cite{wang2025oak, kwon2024ictc}, and composed image retrieval~\cite{vo2019tirg, liu2021cirr, saito2023pic2word, baldrati2023searle} takes image-text inputs but uses the text to specify how the retrieved images should be \emph{different} from the query image. Unlike prior conditional similarity work, we collect human similarity judgments directly under each aspect condition. This captures fine-grained relative differences not reliably present in image captions and VLM outputs, better aligning our method with human judgments than prior training-free, training-based methods, and state-of-the-art VLMs.

\myparagraph{Vision-language models.}
Early vision-language models targeted a small number of tasks: dual-encoder contrastive models~\cite{radford2021clip,zhai2023siglip,cherti2023openclip} such as CLIP~\cite{radford2021clip} align image and text representations for retrieval and classification, while encoder-decoder models such as BLIP~\cite{li2022blip, li2023blip2} added captioning and visual question answering. Modern instruction-tuned multimodal models including Qwen-VL series~\cite{bai2023qwenvl, wang2024qwen2vl, bai2025qwen3vl, qwen2026qwen35} and others~\cite{alayrac2022flamingo,li2024llavaonevision,chen2024internvl,wu2024deepseekvl2}, and frontier proprietary systems such as Gemini~\cite{geminiteam2023gemini} and GPT~\cite{openai2023gpt4v}---are trained on a much broader task mixtures~\cite{li2024llavaonevision, wang2024qwen2vl, chen2024internvl} that, in some cases, also includes similarity benchmarks such as NIGHTS~\cite{fu2023dreamsim} and GeneCIS~\cite{vaze2023genecis}. A parallel line of universal multimodal embedders, such as VLM2Vec~\cite{jiang2025vlm2vec, meng2025vlm2vecv2} and Qwen3-VL-Embedding~\cite{li2026qwen3vlembedding}, repurposes these backbones for instruction-conditioned retrieval. While VLMs can solve various vision tasks, we find that the performance gap remains for VLMs to yield human-aligned pairwise similarity judgments conditioned on specified aspects. Our benchmark and models aid in closing this gap.

\section{Dataset collection}
\label{sec:dataset}

\begin{figure}
    \centering
    \includegraphics[width=0.85\linewidth]{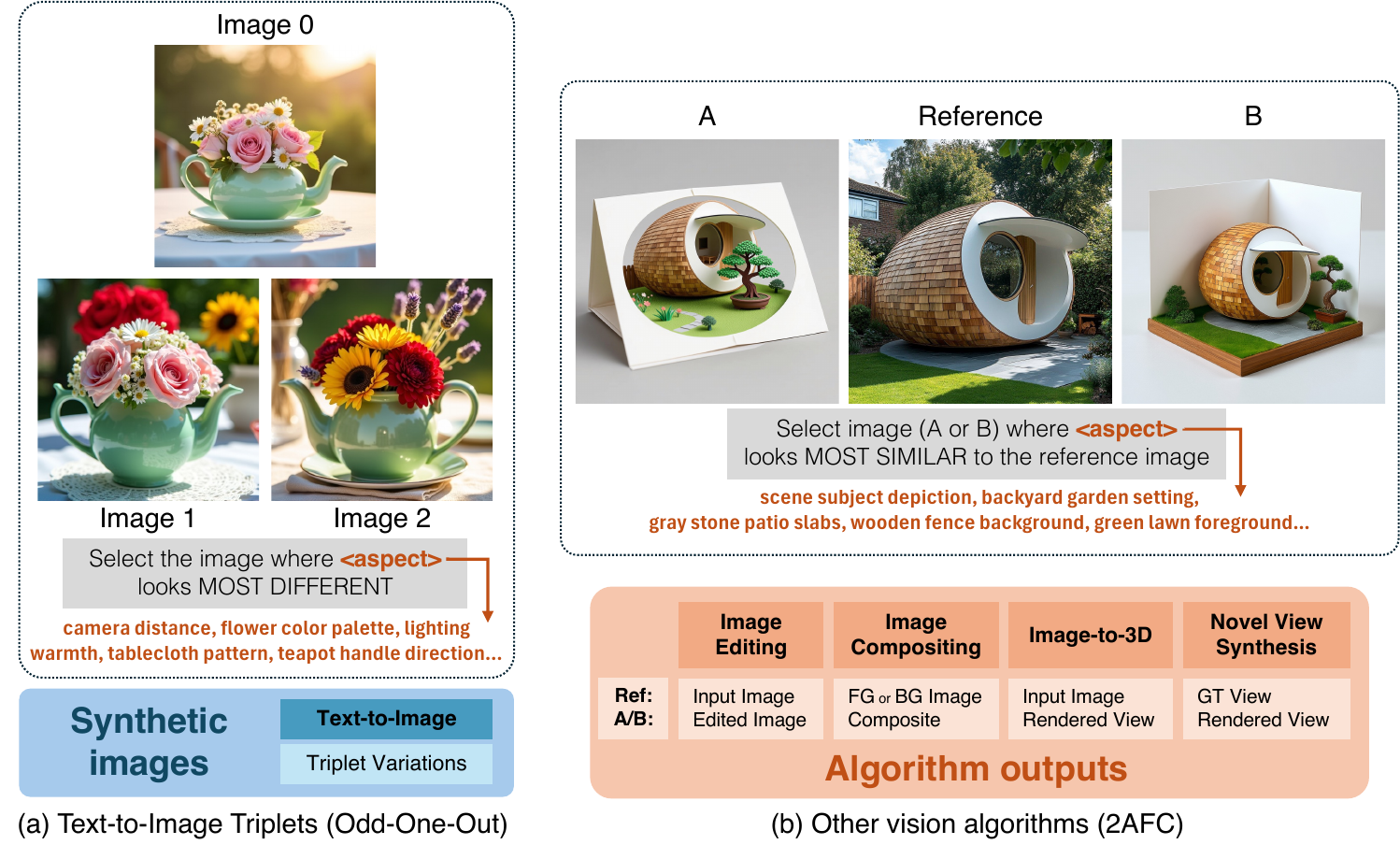}
    \caption{\arxiv{\textbf{Datasets of relative perceptual judgements} for learning and evaluating text-conditioned pairwise similarity.} \textbf{(Left)} We ask users which image within the triplet is the odd-one-out (OOO). The images are synthetically generated using variations of text-to-image models, and are used to train our text-conditioned similarity model. \textbf{(Right)} \arxiv{To test our method as a robust perceptual metric for vision and graphics algorithms, we construct an evaluation set with outputs from 17 unseen external algorithms across 4 task categories.} We ask users which is more similar to a reference image, between two algorithm outputs, in a two-alternative forced choice (2AFC) setup.
    }

    \label{fig:dataset}
\end{figure}

\arxiv{Our goal is to collect a dataset of similarity judgments $\mathcal{D}$ to train and evaluate a pairwise, text-conditioned similarity algorithm.  Because it is difficult for humans to assign consistent similarity scores to image pairs, perceptual similarity is most reliably captured through \textit{relative} comparisons~\cite{tversky1977features}, such as identifying the odd-one-out~\cite{hebart2019things} or the closer pair~\cite{zhang2018lpips,fu2023dreamsim} from an image triplet.
We extend this by conditioning the relative comparisons on specific visual aspects (e.g., lighting, object pose, or background). The relative comparisons implicitly rank the pairs within the triplet, allowing us to learn and evaluate a pairwise similarity metric (detailed in Section~\ref{sec:method}).
As shown in Figure~\ref{fig:dataset}, we collect two datasets: a large, synthetic dataset on odd-one-out judgments for training (Section~\ref{sec:dataset_odd_one_out}), and a smaller dataset designed specifically to test if our text-conditioned similarity metric can reliably evaluate outputs from \textit{unseen} vision and graphics algorithms (Section~\ref{sec:dataset_2afc}).}

\subsection{Odd-one-out dataset}
\label{sec:dataset_odd_one_out}

\arxiv{We build our training dataset by extending the standard odd-one-out task~\cite{hebart2019things} to be conditioned on specific visual aspects. Given an image triplet $\mathbf{X}=\{\mathbf{x}_1,\mathbf{x}_2,\mathbf{x}_3\}$ and a text condition $\mathbf{c}$, we ask annotators to select the image $y$ that is the most different with respect to $\mathbf{c}$. Next, we describe our procedure for generating these triplets $\mathbf{X}$ and text conditionings $\mathbf{c}$.}

\myparagraph{Triplet generation.}
To construct our dataset, we curate a diverse yet controlled set of image variations. A core challenge lies in identifying an optimal difficulty regime: %
\arxiv{we want triplets where visual differences are neither so vast that frontier models trivially match human judgements, nor so negligible that they fail to provide a generalizable learning signal.}
Furthermore, images within a single triplet must exhibit variation across multiple independent text-conditioned aspects to necessitate multi-faceted reasoning. Randomly sampling from large-scale image datasets or subsampling a video dataset typically yields degenerate triplets, where one image is an outlier across all aspects or the visual distance between the images is too great to allow for meaningful comparison.

Inspired by Fu~\etal~\cite{fu2023dreamsim}, we leverage text-to-image (T2I), language (LLM), and vision-language (VLM) models to propose candidate triplets and text conditionings at scale. We generate candidate triplets by sampling prompts from FLUX-Reason-6M~\cite{fang2025fluxreason}. A straightforward approach is to sample images from the same prompt, but this often fails to produce sufficiently diverse variations. As such, we use a language model to generate variations of a given prompt, before rendering the triplet. This produces a controlled yet diverse candidate set,
while allowing humans to judge which of the text-induced variations are actually reflected in the images.

\myparagraph{Text guidance proposal and refinement.}
For each generated triplet, we first construct a candidate set of prompts. During prompt-variation generation, we also prompt an LLM to identify the aspects of the images that are intended to change. If the images were fully described by the input text prompts, this would be sufficient. However, as the mapping between prompts and images is not one-to-one, images may vary from the prompt, or show additional unspecified properties. As such, we pass the triplet and the candidate list to a VLM for refinement, removing unrelated ones and adding visually-varying factors that are not explicit in the prompts (e.g., pose or object size).

The resulting aspect list is intentionally over-complete, and it is not guaranteed to be aligned with how humans describe (or attend to) differences in the triplet. We therefore treat these as proposals and prune. For each triplet, we ask 5 annotators for judgments. We provide an additional ``can't tell'' option, and if marked by 3 or more users, we remove the attribute from the list. 
From the remaining triplets we remove the ``can't tell'' votes and normalize to get probability distribution $\mathbf{y} = (y_1,y_2,y_3) \in \Delta^3$. We also ask for overall similarity, recorded under the ``overall'' condition.

Our final dataset is $\mathcal{D} = \{(\mathbf{X}^{(n)}, \mathbf{c}^{(n)}, \mathbf{y}^{(n)})\}_n$, where each datapoint is a tuple of image triplet $\mathbf{X}^{(n)} = \{\mathbf{x}_1^{(n)}, \mathbf{x}_2^{(n)}, \mathbf{x}_3^{(n)}\}$, aspect condition $\mathbf{c}^{(n)}$, and human vote distribution $\mathbf{y}^{(n)}$. We divide the dataset into train, validation, and test datasets, as summarized in Table~\ref{tab:supp_dataset_stats}.

\begin{table}[t]
\caption{\textbf{Dataset statistics.} We report the number of image triplets, triplet-aspect combinations, and total human votes. Each triplet-aspect combination receives 5 votes; annotators may also select ``can't tell,'' in which case that vote is dropped. To also capture overall similarity, ``overall'' is included as one of the aspects in the odd-one-out dataset for both training and evaluation, whereas the 2AFC evaluation is aspect-conditioned only. See Appendix~\ref{sec:supp_dataset} for further details.}
\vspace{-5pt}
\label{tab:supp_dataset_stats}
\centering

\footnotesize
\setlength{\tabcolsep}{6pt}
\renewcommand{\arraystretch}{1}
\begin{subtable}[t]{0.46\linewidth}
\caption{Odd-one-out dataset \arxiv{($\mathcal{D}$)}.}
\label{tab:supp_dataset_stats_ooo}
\centering
\begin{tabular}{@{}lrrr@{}}
\toprule
\textbf{Split} & \textbf{Triplet} & \textbf{Triplet-Aspect} & \textbf{Human votes} \\
\midrule
Train       & 22{,}157 & 234{,}662 & 938{,}960 \\
Validation  &     185  &   1{,}947 & 7{,}496 \\
Test        &  2{,}000 &  20{,}782 & 98{,}039 \\
\midrule
Total       & 24{,}342 & 257{,}391 & 1{,}044{,}495 \\
\bottomrule
\end{tabular}
\end{subtable}
\hfill
\begin{subtable}[t]{0.50\linewidth}
\caption{\arxiv{External-algorithm 2AFC set ($\mathcal{D}_\text{ext}$).}}
\label{tab:supp_dataset_stats_2afc}
\centering
\begin{tabular}{@{}lrrr@{}}
\toprule
\textbf{Source} & \textbf{Triplet} & \textbf{Triplet-Aspect} & \textbf{Human Votes} \\
\midrule
Img edit            &  256 & 1{,}633 & 4{,}838 \\
Img composite        &  294 &   839 & 2{,}947 \\
NVS     &  180 &   650 & 2{,}444 \\
Im-to-3D          &  397 & 1{,}649 & 6{,}020 \\
\midrule
Total                    & 1{,}127 & 4{,}771 & 16{,}249 \\
\bottomrule
\end{tabular}
\end{subtable}
\vspace{-20pt}
\end{table}

\begin{figure}
    \centering
    \includegraphics[width=\linewidth]{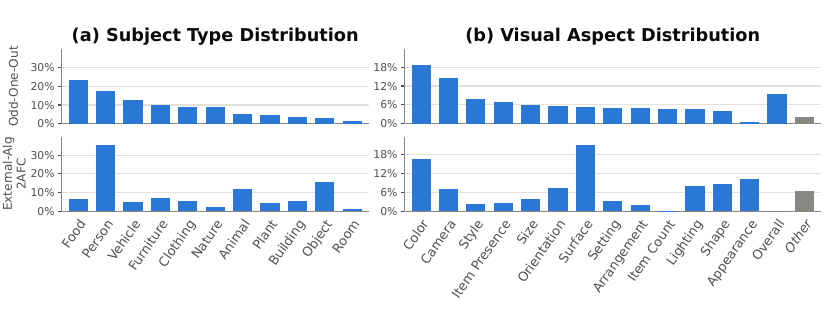}
    \vspace{-35pt}
    \caption{\arxiv{\textbf{Data diversity.} Left shows subject type distribution across the image triplets, and right categorizes the properties captured by our visual aspects. Top and bottom rows represent the odd-one-out and external-algorithm 2AFC dataset, respectively. Subject types are determined via CLIP image-text similarity, and visual aspects via lexical lookups. Further details are in Appendix~\ref{sec:supp_dataset}.}}
    \vspace{-15pt}
    \label{fig:data_diversity}
\end{figure}

\vspace{-15pt}
\subsection{\arxiv{External-algorithm} 2AFC dataset}
\label{sec:dataset_2afc}
To test generalization to different image variations and content, we curate an additional \textit{\arxiv{external-algorithm} evaluation} set from the outputs of other \arxiv{unseen} computer vision algorithms.
\arxiv{We curate this set to demonstrate our method's capability to evaluate outputs from vision algorithms under various visual aspects. Because these algorithms typically introduce targeted visual changes, such as altering a specific object or camera viewpoint, their evaluation inherently demands the interpretability of an aspect-conditioned similarity metric.}

Unlike our main dataset, this setting is inherently \emph{reference-based}: each query comes with a reference image (either the algorithm input or the ground-truth target) and multiple candidate outputs. We therefore adopt a 2-alternative-forced-choice (2AFC) protocol: given an aspect $\mathbf{c}$, annotators are shown the reference $\mathbf{x}_\text{ref}$ and two randomly sampled outputs $(\mathbf{x}_0, \mathbf{x}_1)$, and select which output is more similar to the reference under $\mathbf{c}$, with multiple votes across annotators averaged as $y\in [0, 1]$. Similar to the odd-one-out dataset, we collect votes across 5 annotators and remove those with $\geq 3$ ``can't tell'' votes. We also remove datapoints that have a tie, as they are not useful for evaluation.

This produces an evaluation dataset $\mathcal{D}_\text{ext} = \{(\mathbf{x}_\text{ref}, \mathbf{x}_0, \mathbf{x}_1, \mathbf{c}, y)\}$.
The algorithms and domains used in this \arxiv{external-algorithm} set differ from the training data.%

\arxiv{An overview of dataset statistics and diversity is in Table~\ref{tab:supp_dataset_stats} and Figure~\ref{fig:data_diversity}, respectively. Additional diversity analysis and data curation details are in Appendix~\ref{sec:supp_dataset}.}

\section{Learning Text-Guided Perceptual Similarity}
\label{sec:method}

\begin{figure}
    \centering
    \includegraphics[width=1.\linewidth]{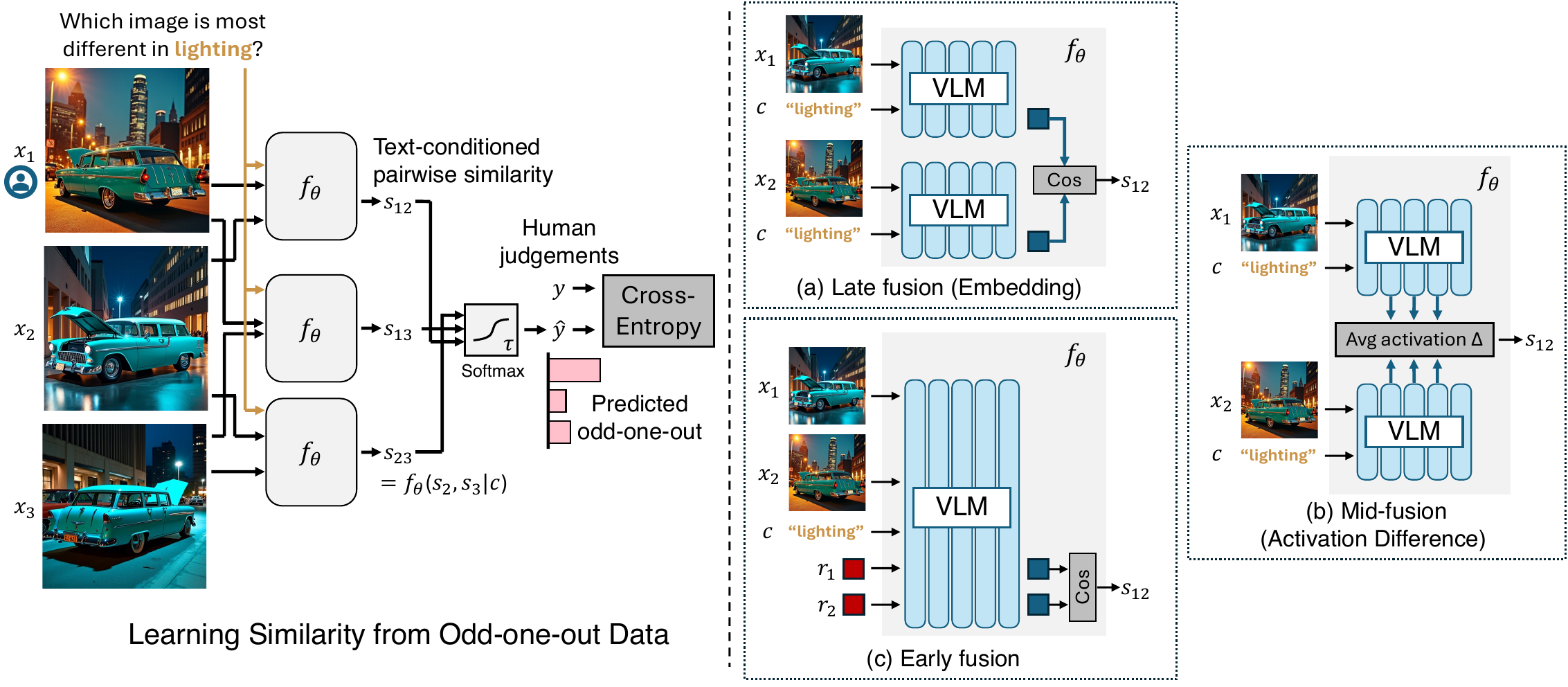}
    \vspace{-18pt}
    \caption{\textbf{(Left) Learning pairwise similarity from odd-one-out data.} 
    \arxiv{To supervise our pairwise similarity model $f_\theta$ using odd-one-out data, we convert the model's pairwise similarity scores directly into an odd-one-out prediction. Pairwise similarity (e.g., $s_{23}$) acts as the logit for the remaining image ($\textbf{x}_1$) being the least similar, odd-one-out image. These three logits are passed through a temperature-scaled ($\tau$) softmax and trained against human annotations via cross-entropy loss. \textbf{(Right) Fusion strategies for $f_\theta$.} We fine-tune a base VLM using three approaches: (a) \textit{Late fusion} embeds the images separately and only compares the final activation; (b) \textit{Mid-fusion} uses internal feature activations between separately embedded images; (c) \textit{Early fusion} feeds both images and the aspect condition as a single sequence into a VLM. Because the VLM’s architecture is not invariant to the ordering of the images, we introduce two learnable, shared-weight register tokens acting as image readouts. By restricting each register to attend exclusively to its respective image, their cosine similarity guarantees symmetry and identity by construction. Implementation details of fusion strategies are in Appendix~\ref{sec:supp_implementation}.}}
    \label{fig:learning objectives}
    \vspace{-10pt}
\end{figure}

We learn text-prompted similarity $f_\theta(\mathbf{x}_1, \mathbf{x}_2 \mid \mathbf{c})$ to be consistent with the annotators' choices in the odd-one-out dataset $\mathcal{D}$. We show the objective in Figure~\ref{fig:learning objectives} \arxiv{(left)}. Suppose that for a triplet, all annotators selected $\mathbf{x}_k$ to be the odd-one-out. This implies that the similarity between the other two images, $f_\theta(\mathbf{x}_i, \mathbf{x}_j \mid \mathbf{c})$, where $\{i, j\} = \{1, 2, 3\} \setminus \{k\}$, should be more similar to each other than any pair involving $\mathbf{x}_k$. Formally, $f_\theta(\mathbf{x}_i, \mathbf{x}_j \mid \mathbf{c}) > f_\theta(\mathbf{x}_k, \mathbf{x}_l \mid \mathbf{c}), \hspace{2mm} \forall l \hspace{.5mm} \in \{i, j\}.$
\arxiv{We follow the softmax choice model in prior odd-one-out task~\cite{muttenthaler2023humanalignment} that leverages this fact to convert predicted pairwise similarities to an odd-one-out prediction.}
\arxiv{In our case, the similarity is text-conditioned, and we define} $\bar{s}_k = s_{ij} = f_\theta(\mathbf{x}_i, \mathbf{x}_j \mid \mathbf{c})$ and temperature $\tau$ > 0,
and we model the predicted likelihood of $k$ being chosen, comparing it to the human vote $\mathbf{y}$ with a cross-entropy loss.

\begin{equation}
\label{eq:triplet_likelihood}
\hat{y}_k = p_\theta(k \mid \mathbf{X}, \mathbf{c})
\;=\; 
\frac{\exp \bigl( \bar{s}_k / \tau\bigr)}
{\sum_{m=1}^3 \exp \bigl( \bar{s}_m / \tau\bigr)}, 
\hspace{4mm} \mathcal{L}(\theta)
\;=\;
- \mathbb{E}_{(\mathbf{X},\,\mathbf{c},\,\mathbf{y}) \sim \mathcal{D}}
\Biggl[\, \sum_{k=1}^{3} y_k \log \hat{y}_k \,\Biggr].
\end{equation}

\myparagraph{\arxiv{Fusion strategies.}}
\arxiv{While Vision-Language Models (VLMs) can be prompted zero-shot to judge text-conditioned image similarity, their predictions diverge substantially from human perception (Section~\ref{sec:experiments}). To bridge this gap, we fine-tune a base VLM to predict a symmetric, scalar similarity score. %
To extract a pairwise similarity score, we explore three fusion strategies that differ in how early the two compared images interact within the network, as shown in Figure~\ref{fig:learning objectives} (right), all trained end-to-end with Equation~\ref{eq:triplet_likelihood}. Late fusion independently encodes the images and text prompt, computing cosine similarity over their final embeddings, which enables highly efficient, cacheable retrieval. Mid fusion adapts the LPIPS~\cite{zhang2018lpips} strategy to VLMs, comparing internal feature activations across multiple layers. Finally, early fusion packs both images and text into a single sequence, enabling direct patch-level correspondence via full joint attention. As our experiments show in Section~\ref{sec:experiments}, early fusion achieves the highest overall performance, while late fusion offers efficiency for large-scale retrieval. Mid fusion, though effective, provides no distinct advantage over either. More fusion implementation and training details are provided in Appendix~\ref{sec:supp_implementation}.}

\begin{figure}
    \centering
    \includegraphics[width=1.0\linewidth]{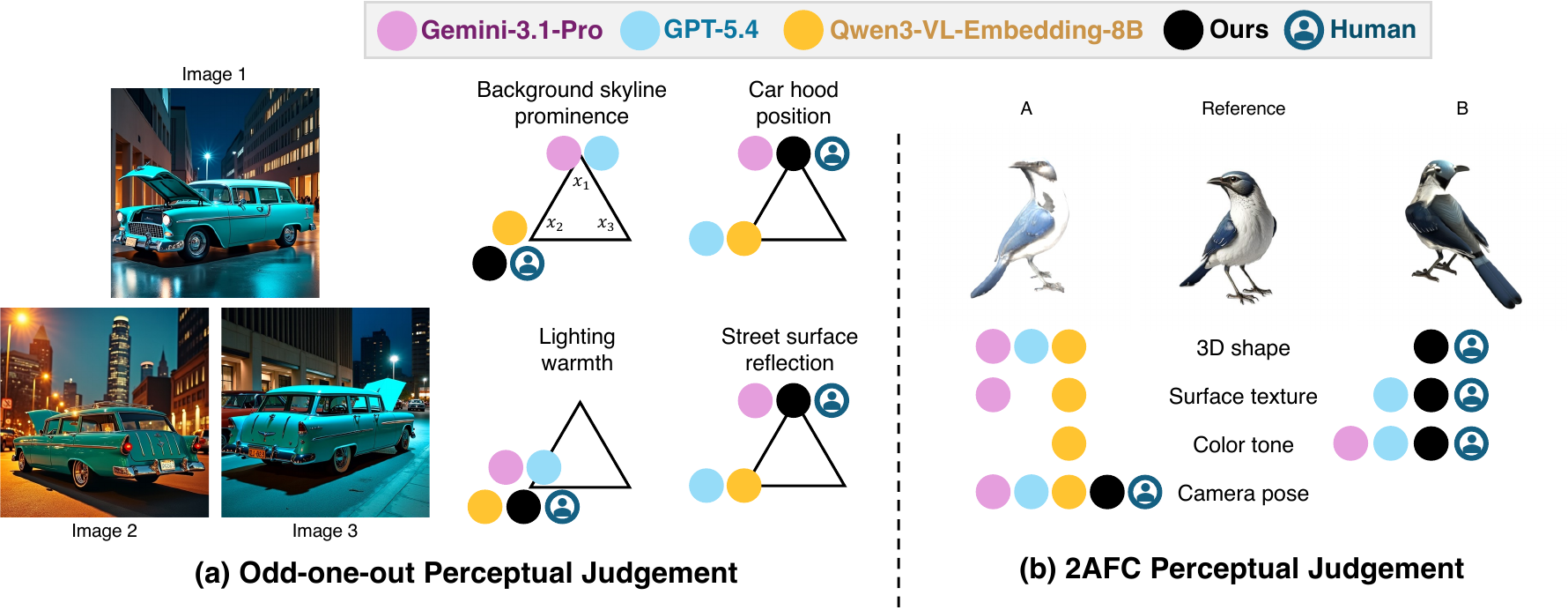}
    \vspace{-10pt}
    \caption{\textbf{\updated{Do humans agree with the model's choice?}} We compare aspect-conditioned similarity among VLMs, our late-fusion model, and humans, denoted by different dot colors and patterns. (Left) Odd-one-out: the dot location indicates each method's selection of the most different image for the aspect. (Right) 2AFC: the dot location indicates the image is selected as more similar to the reference. Note that our method (black) agrees with the human judgment, whereas VLMs often do not.}
    \label{fig:qual_model_human_agreement.pdf}
\end{figure}

\begin{figure}
    \centering
    \includegraphics[width=1.0\linewidth]
        {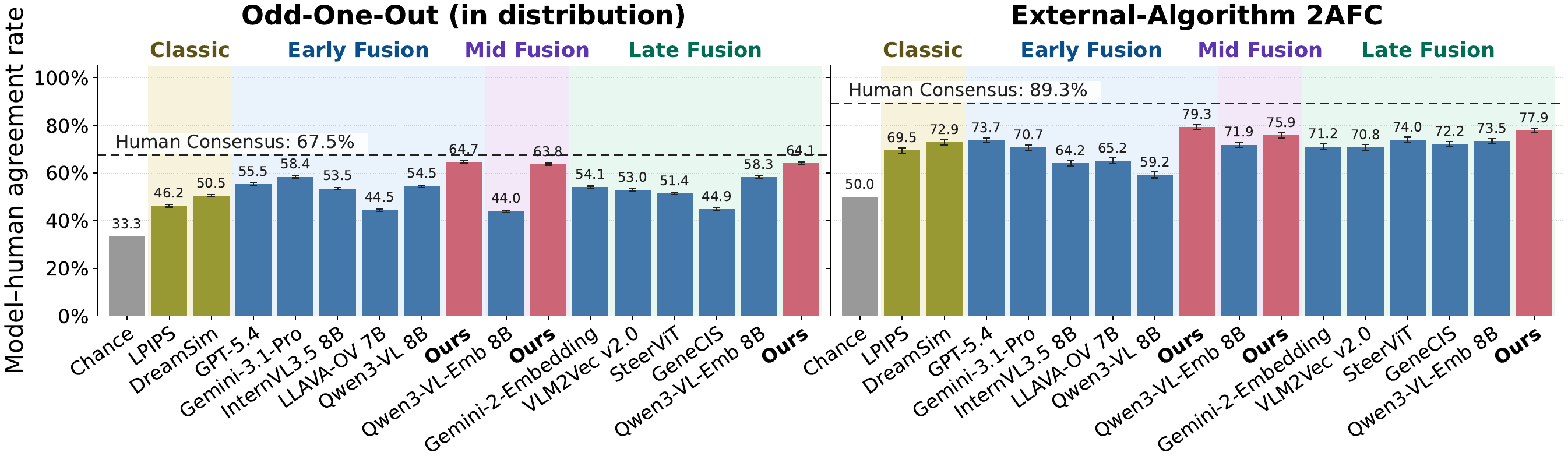}
        \vspace{-10pt}
    \caption{\textbf{How well do models agree with human perceptual judgement?} We report model-human agreement rate across recent vision language models and text-image embedding methods, categorized as classic, early fusion, mid fusion, and late fusion. \arxiv{We show results on the held-out test split of our odd-one-out set $\mathcal{D}$ (left) and our external-algorithm 2AFC set $\mathcal{D}_\text{ext}$ (right)}. Error bar denotes $\pm 2$ standard error across test examples. We find that our method (red bars) consistently outperforms baselines in both in-domain odd-one-out tests and out-of-domain 2AFC tests. Our early fusion model achieves slightly better model-human-agreement than the other two variants. The dotted line denotes the \arxiv{human consensus rate}, providing a reference for agreement between human annotators. Extended comparisons are detailed in Table~\ref{tab:big_results}.}
    \label{fig:plt_human_agreement}

\end{figure}

\section{Experiments}
\label{sec:experiments}

We evaluate on the odd-one-out and \arxiv{external-algorithm} 2AFC test sets. Our dataset improves VLMs on other perceptual similarity benchmarks.
We discuss applications of our methods, such as aspect-conditioned retrieval and compositional retrieval, and show qualitative examples.

\begin{figure}
    \centering
    \includegraphics[width=0.85\linewidth]{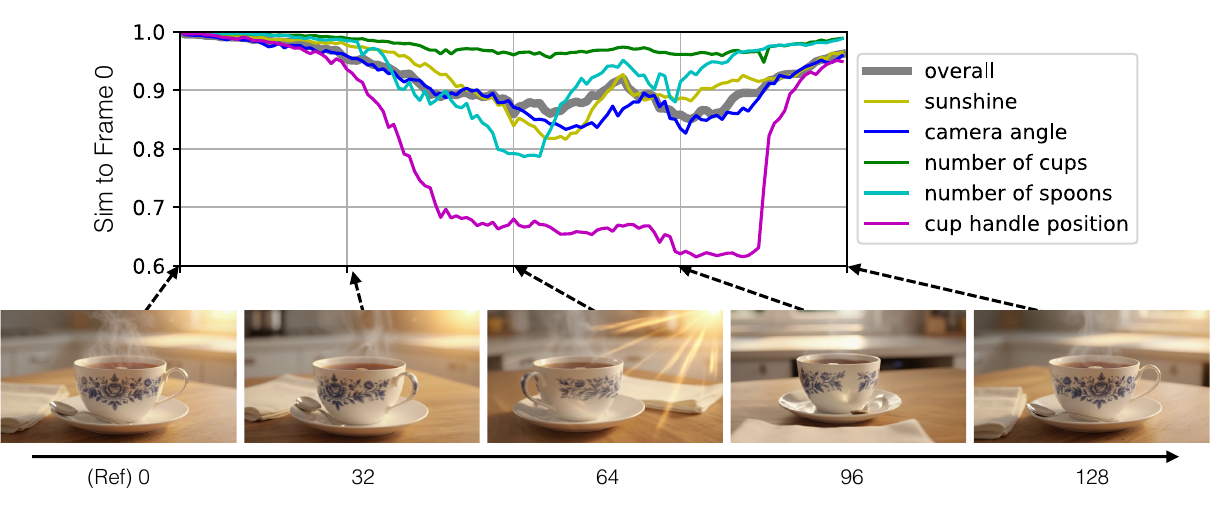}
    \vspace{-10pt}
    \caption{\textbf{Video example.} Similarity between video frames to the initial frame. While the overall similarity is dominated by the main factors (camera angle, as well as sunshine), the metric can be steered to different attributes, such as the cup handle position.}
    \label{fig:video}
    \vspace{-15pt}
\end{figure}

\myparagraph{Metrics.}
For each aspect in a triplet, we report \updated{models' agreement with humans}~\cite{zhang2018lpips}, measuring how often a random rater agrees with the model's top choice. For an odd-one-out triplet with human vote distribution $\mathbf{y} = (y_1, y_2, y_3)$ and model prediction $\hat{k} = \arg\max_k \hat{y}_k$, agreement is $y_{\hat{k}}$--the probability humans vote the same as model prediction. Averaging all aspect conditions over the test set gives the expected rater--model agreement. The 2AFC analogue is defined identically over two choices. %

To contextualize absolute agreement numbers, we compute \updated{\textit{human consensus}}. It is defined as $y_1^2+ y_2^2+ y_3^2$, the probability that the two human responses agree if we independently sample two responses from human votes $\mathbf{y}$. We average this quantity across aspect conditions.
Likewise, the 2AFC analogue is defined identically over two choices.

\myparagraph{Baselines} are organized below:
\begin{itemize}[leftmargin=12pt,itemsep=0pt,topsep=0pt]
    \item \textbf{Classic}: LPIPS~\cite{zhang2018lpips}, DreamSim~\cite{fu2023dreamsim}. Apply same similarity score regardless of the queried aspect.
    \item \textbf{Late fusion (embeddings)}: text-image embedding models including text-conditioned encoders (SteerViT~\cite{ruthardt2026steervit}), late mixing of text and image embeddings (GeneCIS~\cite{vaze2023genecis}), VLM-finetuned embedding models (VLM2Vec-2.0~\cite{meng2025vlm2vecv2}, Qwen3-VL-Embedding~\cite{li2026qwen3vlembedding}), and proprietary models (Gemini-2-Embedding~\cite{geminiteam2023gemini}).
    \item \textbf{Early fusion}: VLMs including open-weight models (Qwen3-VL~\cite{bai2025qwen3vl}, InternVL3.5~\cite{wang2025internvl35}, and LLaVA-OneVision~\cite{li2024llavaonevision}) and proprietary models (GPT-5.4 and Gemini 3.1 Pro). Each model is given both images and the text prompt and asked to return \arxiv{an integer value from 0 to 10 in text. Unlike our early fusion model (Section~\ref{sec:method}), these baselines lack image-order invariance. We address this by averaging their scores across both orderings.} 
    \item \textbf{Mid-fusion (activation difference)}: we find that Qwen3-VL series is the best performing open-weight backbone in both late and early fusion, and focus on this family for the mid-fusion analysis.
\end{itemize}
More implementation details of the baselines are in Appendix~\ref{sec:supp_baselines}.

We present our main results in Figure~\ref{fig:qual_model_human_agreement.pdf} and~\ref{fig:plt_human_agreement}. On our odd-one-out task, a substantial gap exists between the human consensus reference ($67.5\%$, with $33.3\%$ chance) and baseline model predictions across both tasks. LPIPS ($46.2\%$) and DreamSim ($50.5\%$) are aspect-agnostic, and merely providing an overall similarity judgement is not sufficient. Among baselines, Gemini-3.1-Pro achieves the highest model-human agreement on the odd-one-out task at $58.4\%$. 
Among open-weight models, the embedding-based Qwen3-VL-Embed performs best on both tasks, and we therefore adopt it as the backbone for our approach. Fine-tuning on our data substantially narrows the gap to the human consensus reference, reducing it from $9.1\%$ to $2.8\%$ on the odd-one-out task.

\begin{figure}[!t]
    \centering
    \includegraphics[width=1.0\linewidth]{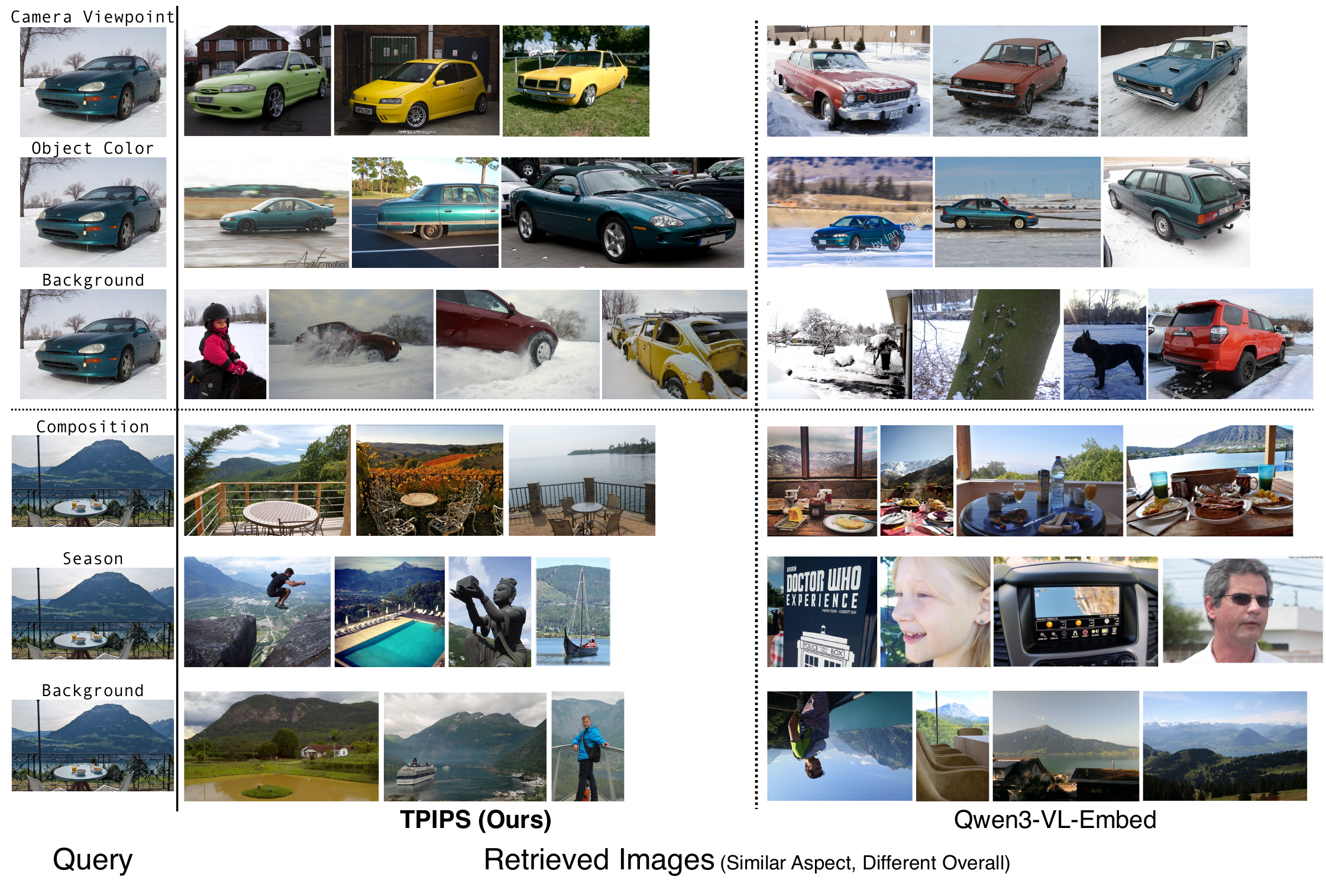}
    \vspace{-20pt}
    \caption{\updated{\textbf{Aspect-conditioned Image Retrieval.} Given a query on the left, our method retrieves images with respect to different visual aspects. Each row represents an aspect. Compared to the base model (Qwen3-VL-8B-Embedding), our fine-tuned late-fusion model consistently retrieves images that better match the specified aspects. To encourage retrieval diversity, we adjust the ranking score by subtracting a fraction of the overall similarity from the aspect similarity. We also include retrieval results without this adjustment in Appendix~\ref{sec:supp_reading_table}.}}
    \label{fig:retrieval_multi_factor}
    \vspace{-5pt}
\end{figure}

\begin{figure}[!h]
    \centering
    \includegraphics[width=0.95\linewidth]{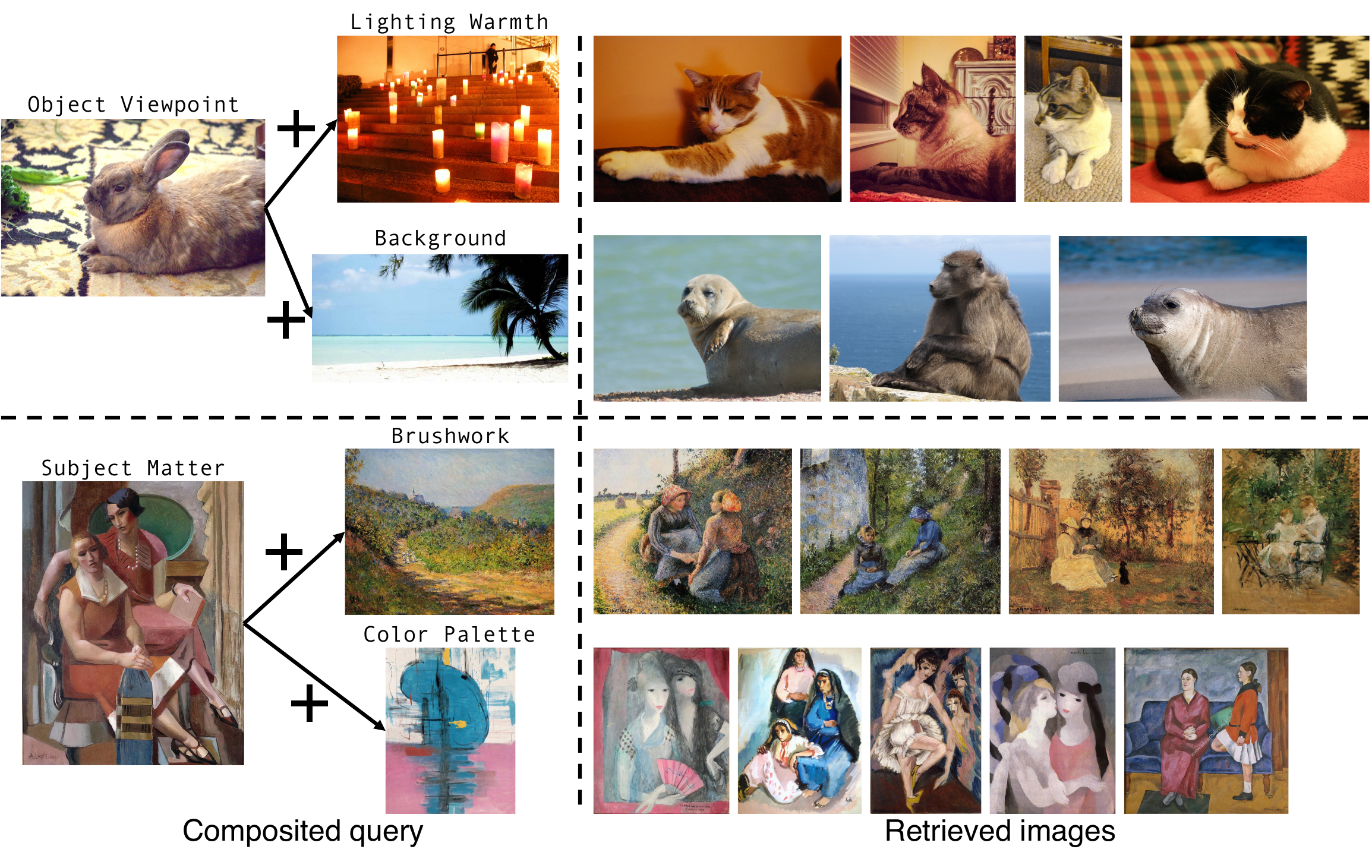}
    \vspace{-10pt}
    \caption{\textbf{Compositional retrieval.} Each query is specified with different aspects, and we directly sum up similarity scores from two queries to rank the data. For example, combining (1) the painting of two sitting people conditioned on subject matter and (2) Monet's painting conditioned on brush work yields Impressionist paintings of two sitting people. When we swap out the second query to be color-palette specific, we can retrieve portraits of two people in the same pink and blue color tone.} %
    \label{fig:retrieval_multi_query}
\end{figure}

\myparagraph{\arxiv{Generalization.}}
\arxiv{As shown in Figure~\ref{fig:plt_human_agreement}, the gains shown in the odd-one-out test} 
translate to the \arxiv{external-algorithm} 2AFC distribution, reducing \arxiv{the human consensus gap} from $15.8\%$ to $10.0\%$, above baselines.
\arxiv{These results show that our method, compared with baselines like frontier VLMs, can serve as a more human-aligned perceptual metric for vision and graphics algorithms. In Appendix~\ref{sec:supp_generalization}, we also show that our method generalizes to other benchmarks on unconditional perceptual similarity, attribute-specific retrieval, and text-conditional retrieval on real-world complex scenes.}

\myparagraph{Early, mid, and late-fusion.}
Among the baseline models, no single fusion strategy consistently dominates. Mid-fusion performs relatively poorly for off-the-shelf baselines, as such models are not designed to produce informative mid-level features directly. When we fine-tune all fusion strategies under the same Qwen3-VL-8B-Embedding base model, early fusion consistently outperforms both middle and late fusion, albeit by a modest margin. This suggests that allowing the model to interact with both images from the earliest layers is beneficial for perceptual similarity judgements. On the other hand, the late fusion model achieves similar performance but enables efficient retrieval by precomputing embeddings. We will present our results mainly with the late fusion model unless specified. We analyze the effect of dataset scale and model size in more detail in Appendix~\ref{sec:supp_ablations}.

\myparagraph{Qualitative video example.}
In Fig~\ref{fig:video}, we show a synthesized video of a camera rotating around a teacup, and calculate our metric from each frame to the initial frame. The main factor that vary is the ``camera angle'', as well as ``sunshine'' that appears in the middle of the video. As such, the ``overall'' similarity mirrors these two main factors. However, the metric can be prompted to completely different factors. The ``number of cups'' stays nearly constant, while ``number of spoons'' drops when the spoon disappears behind the cup. More dramatically, prompting for the ``cup handle position'' causes a large perturbation, as the handle moves from the right side to the left.

\myparagraph{Aspect-conditioned retrieval.}
Figure~\ref{fig:retrieval_multi_factor} illustrates how our method can retrieve perceptually similar images under different aspect conditions given the same query. We index the OpenImages~\cite{kuznetsova2020openimages} training set (1.7M images) and retrieve nearest neighbors conditioned on each aspect independently. \updated{We encourage the retrieval for each aspect to have more different-looking images by adjusting the similarity score to $f_\theta(\mathbf{x}_1, \mathbf{x}_2 \mid \mathbf{c_\text{aspect}}) - 0.3*f_\theta(\mathbf{x}_1, \mathbf{x}_2 \mid \mathbf{c_\text{overall}})$, where $\mathbf{c_\text{aspect}}$ and $\mathbf{c_\text{overall}}$ represent aspect and overall condition, respectively.} The results clearly demonstrate that perceptual similarity is context-dependent. The same query yields meaningfully different nearest neighbors depending on the aspect condition. We provide retrieval results without the similarity adjustments in Appendix~\ref{sec:supp_reading_table}.

\myparagraph{Compositional retrieval.}
Our method supports compositional queries by linearly combining similarity scores across multiple queries conditioned on different aspect conditions. We validate this on OpenImages again, and we also show results on the WikiArt dataset~\cite{artgan2018} (80K images), where aspects such as subject matter, brushwork, and color palette capture independently meaningful dimensions of visual similarity. As shown in Figure~\ref{fig:retrieval_multi_query}, composing queries across different aspects produces retrievals that jointly satisfy all specified conditions, and swapping a single query component leads to interpretably different results. We provide more qualitative results in Appendix~\ref{sec:supp_reading_table}.

\begin{figure}[!h]
    \centering
    \includegraphics[width=\linewidth]{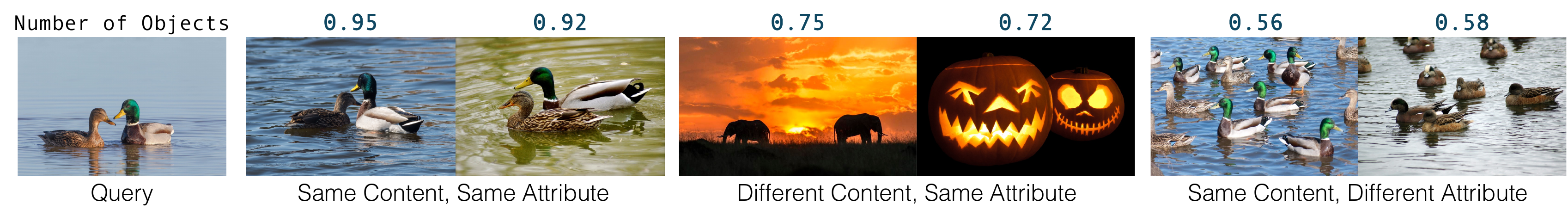}
    \vspace{-20pt}
    \caption{\arxiv{\textbf{Limitation.} Given the reference image and aspect condition \texttt{Number of Objects} on the left, we show similarity scores of images from our late-fusion method on the right. Our model cannot completely isolate a specific aspect from the object identity, giving higher similarity scores to other paired ducks over other paired objects. However, our method can still assign much lower similarity to an image of different number of ducks.}}
    \label{fig:limitation}
    \vspace{-10pt}
\end{figure}

\section{Discussion, Broader Impacts, and Limitations}
\label{sec:discussion}
Our metric is human-aligned, aspect-conditioned, and it can also serve as a differentiable feedback signal, as recent work has leveraged VLM feedback to train image editing~\cite{kumari2026npedit, luo2025dualprocess} and generation~\cite{xu2023imagereward, xue2025dancegrpo} models. Our method can also aid in interpretability systems that automatically discover and characterize neural features~\cite{shaham2024maia, surkov2024sae}. 
\arxiv{Our metric clarifies exactly what visual concepts an image captures, making evaluation much more transparent. As generative systems see wider use, this practical interpretability is helpful for meaningful public oversight.}

\arxiv{One recurring limitation is concept entanglement, where the model cannot completely isolate a specific aspect from the object identity, as shown in Figure~\ref{fig:limitation}.} Our VLM-based metric is substantially more expensive to compute than LPIPS or DreamSim. \updated{Also, our current embedding-based retrieval requires separate feature encoding and indexing for each aspect condition.} In addition, our annotator pool is not representative of all viewers; perception varies across people, and our metric inherits that bias. Our method is also limited by training data diversity. We source prompts from FLUX-Reason-6M~\cite{fang2025fluxreason}, which has more aesthetically curated images. Our aspect proposals are VLM-generated, and we can systematically miss aspects that VLMs cannot capture. %
Data-related issues could be addressed with more diverse image sources and human-annotated aspects in future work. That said, as VLMs continue to improve, our pipeline naturally benefits — and we view our human perceptual data as a complementary signal toward building more human-aligned models, offering a new angle for collecting fine-grained human perceptual judgments.

\myparagraph{Acknowledgments.} We thank Nupur Kumari, Kangle Deng, and Antonio Torralba for helpful discussions and feedback on drafts. Sheng-Yu Wang is supported by the Google PhD Fellowship. The project was partly supported by Adobe Inc., the Packard Fellowship, the Institute of Information \& Communications Technology Planning \& Evaluation (IITP) grant funded by the Korean government (MSIT) (No. RS-2024-00457882, National AI Research Lab Project), NSF IIS-2239076, NSF ISS-2403303, and NSF IIS-2403305.

\medskip

{
\small
\bibliographystyle{unsrt}
\bibliography{main}

@inproceedings{lee2024nvembed,
  title={{NV-Embed}: Improved Techniques for Training {LLM}s as Generalist Embedding Models},
  author={Lee, Chankyu and Roy, Rajarshi and Xu, Mengyao and Raiman, Jonathan and Shoeybi, Mohammad and Catanzaro, Bryan and Ping, Wei},
  booktitle={International Conference on Learning Representations (ICLR)},
  year={2025}
}

@inproceedings{hu2022lora,
  title={{LoRA}: Low-Rank Adaptation of Large Language Models},
  author={Hu, Edward J. and Shen, Yelong and Wallis, Phillip and Allen-Zhu, Zeyuan and Li, Yuanzhi and Wang, Shean and Wang, Lu and Chen, Weizhu},
  booktitle={International Conference on Learning Representations (ICLR)},
  year={2022}
}

@article{wang2004ssim,
  title={Image quality assessment: from error visibility to structural similarity},
  author={Wang, Zhou and Bovik, Alan C and Sheikh, Hamid R and Simoncelli, Eero P},
  journal={IEEE Transactions on Image Processing},
  volume={13},
  number={4},
  pages={600--612},
  year={2004},
  publisher={IEEE}
}

@article{zhang2011fsim,
  title={{FSIM}: A feature similarity index for image quality assessment},
  author={Zhang, Lin and Zhang, Lei and Mou, Xuanqin and Zhang, David},
  journal={IEEE Transactions on Image Processing},
  volume={20},
  number={8},
  pages={2378--2386},
  year={2011},
  publisher={IEEE}
}

@article{mantiuk2011hdrvdp2,
  title={{HDR-VDP-2}: A calibrated visual metric for visibility and quality predictions in all luminance conditions},
  author={Mantiuk, Rafat and Kim, Kil Joong and Rempel, Allan G and Heidrich, Wolfgang},
  journal={ACM Transactions on Graphics (Proc. SIGGRAPH)},
  volume={30},
  number={4},
  pages={40:1--40:14},
  year={2011}
}

@inproceedings{krizhevsky2012alexnet,
  title={{ImageNet} classification with deep convolutional neural networks},
  author={Krizhevsky, Alex and Sutskever, Ilya and Hinton, Geoffrey E},
  booktitle={Advances in Neural Information Processing Systems (NeurIPS)},
  year={2012}
}

@inproceedings{simonyan2015vgg,
  title={Very deep convolutional networks for large-scale image recognition},
  author={Simonyan, Karen and Zisserman, Andrew},
  booktitle={International Conference on Learning Representations (ICLR)},
  year={2015}
}

@inproceedings{he2016resnet,
  title={Deep residual learning for image recognition},
  author={He, Kaiming and Zhang, Xiangyu and Ren, Shaoqing and Sun, Jian},
  booktitle={Proceedings of the IEEE Conference on Computer Vision and Pattern Recognition (CVPR)},
  pages={770--778},
  year={2016}
}

@inproceedings{dosovitskiy2021vit,
  title={An image is worth 16x16 words: Transformers for image recognition at scale},
  author={Dosovitskiy, Alexey and Beyer, Lucas and Kolesnikov, Alexander and Weissenborn, Dirk and Zhai, Xiaohua and Unterthiner, Thomas and Dehghani, Mostafa and Minderer, Matthias and Heigold, Georg and Gelly, Sylvain and Uszkoreit, Jakob and Houlsby, Neil},
  booktitle={International Conference on Learning Representations (ICLR)},
  year={2021}
}

@article{oquab2024dinov2,
  title={{DINOv2}: Learning robust visual features without supervision},
  author={Oquab, Maxime and Darcet, Timoth{\'e}e and Moutakanni, Th{\'e}o and Vo, Huy and Szafraniec, Marc and Khalidov, Vasil and Fernandez, Pierre and Haziza, Daniel and Massa, Francisco and El-Nouby, Alaaeldin and Assran, Mahmoud and Ballas, Nicolas and Galuba, Wojciech and Howes, Russell and Huang, Po-Yao and Li, Shang-Wen and Misra, Ishan and Rabbat, Michael and Sharma, Vasu and Synnaeve, Gabriel and Xu, Hu and J{\'e}gou, Herv{\'e} and Mairal, Julien and Labatut, Patrick and Joulin, Armand and Bojanowski, Piotr},
  journal={Transactions on Machine Learning Research (TMLR)},
  year={2024}
}

@inproceedings{johnson2016perceptualloss,
  title={Perceptual losses for real-time style transfer and super-resolution},
  author={Johnson, Justin and Alahi, Alexandre and Fei-Fei, Li},
  booktitle={European Conference on Computer Vision (ECCV)},
  pages={694--711},
  year={2016},
  organization={Springer}
}

@inproceedings{gatys2016styletransfer,
  title={Image style transfer using convolutional neural networks},
  author={Gatys, Leon A and Ecker, Alexander S and Bethge, Matthias},
  booktitle={Proceedings of the IEEE Conference on Computer Vision and Pattern Recognition (CVPR)},
  pages={2414--2423},
  year={2016}
}

@inproceedings{amir2021understanding,
  title={Understanding and simplifying perceptual distances},
  author={Amir, Dan and Weiss, Yair},
  booktitle={Proceedings of the IEEE/CVF Conference on Computer Vision and Pattern Recognition (CVPR)},
  pages={12226--12235},
  year={2021}
}

@inproceedings{zhang2018lpips,
  title={The unreasonable effectiveness of deep features as a perceptual metric},
  author={Zhang, Richard and Isola, Phillip and Efros, Alexei A and Shechtman, Eli and Wang, Oliver},
  booktitle={Proceedings of the IEEE Conference on Computer Vision and Pattern Recognition (CVPR)},
  pages={586--595},
  year={2018}
}

@inproceedings{prashnani2018pieapp,
  title={{PieAPP}: Perceptual image-error assessment through pairwise preference},
  author={Prashnani, Ekta and Cai, Hong and Mostofi, Yasamin and Sen, Pradeep},
  booktitle={Proceedings of the IEEE Conference on Computer Vision and Pattern Recognition (CVPR)},
  pages={1808--1817},
  year={2018}
}

@article{ding2020dists,
  title={Image quality assessment: Unifying structure and texture similarity},
  author={Ding, Keyan and Ma, Kede and Wang, Shiqi and Simoncelli, Eero P},
  journal={IEEE Transactions on Pattern Analysis and Machine Intelligence (TPAMI)},
  volume={44},
  number={5},
  pages={2567--2581},
  year={2020},
  publisher={IEEE}
}

@inproceedings{fu2023dreamsim,
  title={{DreamSim}: Learning new dimensions of human visual similarity using synthetic data},
  author={Fu, Stephanie and Tamir, Netanel Y and Sundaram, Shobhita and Chai, Lucy and Zhang, Richard and Dekel, Tali and Isola, Phillip},
  booktitle={Advances in Neural Information Processing Systems (NeurIPS)},
  year={2023}
}

@inproceedings{sargent2026vlic,
  title={VLIC: Vision-Language Models As Perceptual Judges for Human-Aligned Image Compression},
  author={Sargent, Kyle and Gao, Ruiqi and Henzler, Philipp and Herrmann, Charles and Holynski, Aleksander and Fei-Fei, Li and Wu, Jiajun and Zhang, Jason Y},
  booktitle={Proceedings of the IEEE/CVF Conference on Computer Vision and Pattern Recognition},
  year={2026}
}

@inproceedings{nguyen2026relational,
  title={Relational Visual Similarity},
  author={Nguyen, Thao and Mo, Sicheng and Singh, Krishna Kumar and Wang, Yilin and Shi, Jing and Kolkin, Nicholas and Shechtman, Eli and Lee, Yong Jae and Li, Yuheng},
  booktitle={Proceedings of the IEEE/CVF Conference on Computer Vision and Pattern Recognition},
  year={2026}
}

@article{garces2014illustrationsim,
author = {Garces, Elena and Agarwala, Aseem and Gutierrez, Diego and Hertzmann, Aaron},
title = {A similarity measure for illustration style},
year = {2014}
}

@inproceedings{somepalli2024csd,
    title={Measuring Style Similarity in Diffusion Models},
    author={Somepalli, Gowthami and Gupta, Anubhav and Gupta, Kamal and Palta, Shramay and Goldblum, Micah and Geiping, Jonas and Shrivastava, Abhinav and Goldstein, Tom},
    year = {2024},
    booktitle={European Conference on Computer Vision (ECCV)}
}

@inproceedings{muttenthaler2023humanalignment,
  title={Human alignment of neural network representations},
  author={Muttenthaler, Lukas and Dippel, Jonas and Linhardt, Lorenz and Vandermeulen, Robert A and Kornblith, Simon},
  booktitle={International Conference on Learning Representations (ICLR)},
  year={2023}
}

@article{hebart2019things,
  title={{THINGS}: A database of 1,854 object concepts and more than 26,000 naturalistic object images},
  author={Hebart, Martin N and Dickter, Adam H and Kidder, Alexis and Kwok, Wan Y and Corriveau, Anna and Van Wicklin, Caitlin and Baker, Chris I},
  journal={PLOS ONE},
  volume={14},
  number={10},
  pages={e0223792},
  year={2019}
}

@inproceedings{ghildyal2022shifttolerantlpips,
  title={Shift-tolerant perceptual similarity metric},
  author={Ghildyal, Abhijay and Liu, Feng},
  booktitle={European Conference on Computer Vision (ECCV)},
  pages={91--107},
  year={2022},
  organization={Springer}
}

@inproceedings{ghazanfari2024lipsim,
  title={{LipSim}: A provably robust perceptual similarity metric},
  author={Ghazanfari, Sara and Araujo, Alexandre and Krishnamurthy, Prashanth and Khorrami, Farshad and Garg, Siddharth},
  booktitle={International Conference on Learning Representations (ICLR)},
  year={2024}
}

@article{kettunen2019elpips,
  title={{E-LPIPS}: Robust perceptual image similarity via random transformation ensembles},
  author={Kettunen, Markus and Härkönen, Erik and Lehtinen, Jaakko},
  journal={arXiv preprint arXiv:1906.03973},
  year={2019}
}

@article{tversky1977features,
  title={Features of similarity},
  author={Tversky, Amos},
  journal={Psychological Review},
  volume={84},
  number={4},
  pages={327--352},
  year={1977}
}

@article{medin1993respects,
  title={Respects for similarity},
  author={Medin, Douglas L and Goldstone, Robert L and Gentner, Dedre},
  journal={Psychological Review},
  volume={100},
  number={2},
  pages={254--278},
  year={1993},
  publisher={American Psychological Association},
  doi={10.1037/0033-295X.100.2.254}
}

@inproceedings{amid2015mvte,
  title={Multiview triplet embedding: Learning attributes in multiple maps},
  author={Amid, Ehsan and Ukkonen, Antti},
  booktitle={International Conference on Machine Learning (ICML)},
  pages={1472--1480},
  year={2015},
  organization={PMLR}
}

@inproceedings{veit2017csn,
  title={Conditional similarity networks},
  author={Veit, Andreas and Belongie, Serge and Karaletsos, Theofanis},
  booktitle={Proceedings of the IEEE Conference on Computer Vision and Pattern Recognition (CVPR)},
  pages={830--838},
  year={2017}
}

@inproceedings{vasileva2018typeaware,
  title={Learning type-aware embeddings for fashion compatibility},
  author={Vasileva, Mariya I and Plummer, Bryan A and Dusad, Krishna and Rajpal, Shreya and Kumar, Ranjitha and Forsyth, David},
  booktitle={European Conference on Computer Vision (ECCV)},
  pages={390--405},
  year={2018}
}

@article{thong2019cooperative,
  title={Cooperative embeddings for instance, attribute and category retrieval},
  author={Thong, William and Snoek, Cees GM and Smeulders, Arnold WM},
  journal={arXiv preprint arXiv:1904.01421},
  year={2019}
}

@inproceedings{tan2019learningsimilarity,
  title={Learning similarity conditions without explicit supervision},
  author={Tan, Reuben and Vasileva, Mariya I and Saenko, Kate and Plummer, Bryan A},
  booktitle={Proceedings of the IEEE/CVF International Conference on Computer Vision (ICCV)},
  pages={10373--10382},
  year={2019}
}

@inproceedings{kawarada2025dior,
  title={Training-free Conditional Image Embedding Framework Leveraging Large Vision Language Models},
  author={Kawarada, Masayuki and Yamada, Kosuke and Tejero-de-Pablos, Antonio and Inoue, Naoto},
  booktitle={Proceedings of the IEEE/CVF Winter Conference on Applications of Computer Vision},
  pages={7636--7646},
  year={2026}
}

@inproceedings{li2025promptableembeddings,
  title={Highlighting what matters: Promptable embeddings for attribute-focused image retrieval},
  author={Li, Siting and Gao, Xiang and Du, Simon Shaolei},
  booktitle={Advances in Neural Information Processing Systems (NeurIPS)},
  year={2025}
}

@inproceedings{vaze2023genecis,
  title={{GeneCIS}: A benchmark for general conditional image similarity},
  author={Vaze, Sagar and Carion, Nicolas and Misra, Ishan},
  booktitle={Proceedings of the IEEE/CVF Conference on Computer Vision and Pattern Recognition (CVPR)},
  pages={6862--6872},
  year={2023}
}

@article{liu2025conditional,
  title={Conditional representation learning for customized tasks},
  author={Liu, Honglin and Sun, Chao and Hu, Peng and Li, Yunfan and Peng, Xi},
  journal={Advances in Neural Information Processing Systems},
  year={2025}
}

@article{hsieh2025focallens,
  title={Focallens: Instruction tuning enables zero-shot conditional image representations},
  author={Hsieh, Cheng-Yu and Vasu, Pavan Kumar Anasosalu and Faghri, Fartash and Vemulapalli, Raviteja and Li, Chun-Liang and Krishna, Ranjay and Tuzel, Oncel and Pouransari, Hadi},
  journal={arXiv preprint arXiv:2504.08368},
  year={2025}
}

@article{chen2025omniattribute,
  title={Omni-Attribute: Open-vocabulary Attribute Encoder for Visual Concept Personalization},
  author={Chen, Tsai-Shien and Siarohin, Aliaksandr and Qian, Guocheng Gordon and Wang, Kuan-Chieh Jackson and Nemchinov, Egor and Haji-Ali, Moayed and Guler, Riza Alp and Menapace, Willi and Skorokhodov, Ivan and Kag, Anil and others},
  journal={arXiv preprint arXiv:2512.10955},
  year={2025}
}

@inproceedings{wang2025oak,
  title={Open ad-hoc categorization with contextualized feature learning},
  author={Wang, Zilin and Mo, Sangwoo and Yu, Stella X and Behpour, Sima and Ren, Liu},
  booktitle={Proceedings of the IEEE/CVF Conference on Computer Vision and Pattern Recognition (CVPR)},
  year={2025}
}

@inproceedings{kwon2024ictc,
  title={Image clustering conditioned on text criteria},
  author={Kwon, Sehyun and Park, Jaeseung and Kim, Minkyu and Cho, Jaewoong and Ryu, Ernest K and Lee, Kangwook},
  booktitle={International Conference on Learning Representations (ICLR)},
  year={2024}
}

@inproceedings{vo2019tirg,
  title={Composing text and image for image retrieval-an empirical odyssey},
  author={Vo, Nam and Jiang, Lu and Sun, Chen and Murphy, Kevin and Li, Li-Jia and Fei-Fei, Li and Hays, James},
  booktitle={Proceedings of the IEEE/CVF Conference on Computer Vision and Pattern Recognition (CVPR)},
  pages={6439--6448},
  year={2019}
}

@inproceedings{liu2021cirr,
  title={Image retrieval on real-life images with pre-trained vision-and-language models},
  author={Liu, Zheyuan and Rodriguez-Opazo, Cristian and Teney, Damien and Gould, Stephen},
  booktitle={Proceedings of the IEEE/CVF International Conference on Computer Vision (ICCV)},
  pages={2125--2134},
  year={2021}
}

@inproceedings{saito2023pic2word,
  title={{Pic2Word}: Mapping pictures to words for zero-shot composed image retrieval},
  author={Saito, Kuniaki and Sohn, Kihyuk and Zhang, Xiang and Li, Chun-Liang and Lee, Chen-Yu and Saenko, Kate and Pfister, Tomas},
  booktitle={Proceedings of the IEEE/CVF Conference on Computer Vision and Pattern Recognition (CVPR)},
  pages={19305--19314},
  year={2023}
}

@inproceedings{baldrati2023searle,
  title={Zero-shot composed image retrieval with textual inversion},
  author={Baldrati, Alberto and Agnolucci, Lorenzo and Bertini, Marco and Del Bimbo, Alberto},
  booktitle={Proceedings of the IEEE/CVF International Conference on Computer Vision (ICCV)},
  pages={15338--15347},
  year={2023}
}

@inproceedings{radford2021clip,
  title={Learning transferable visual models from natural language supervision},
  author={Radford, Alec and Kim, Jong Wook and Hallacy, Chris and Ramesh, Aditya and Goh, Gabriel and Agarwal, Sandhini and Sastry, Girish and Askell, Amanda and Mishkin, Pamela and Clark, Jack and others},
  booktitle={International conference on machine learning},
  pages={8748--8763},
  year={2021},
  organization={PmLR}
}

@inproceedings{zhai2023siglip,
  title={Sigmoid loss for language image pre-training},
  author={Zhai, Xiaohua and Mustafa, Basil and Kolesnikov, Alexander and Beyer, Lucas},
  booktitle={Proceedings of the IEEE/CVF International Conference on Computer Vision (ICCV)},
  pages={11975--11986},
  year={2023}
}

@inproceedings{cherti2023openclip,
  title={Reproducible scaling laws for contrastive language-image learning},
  author={Cherti, Mehdi and Beaumont, Romain and Wightman, Ross and Wortsman, Mitchell and Ilharco, Gabriel and Gordon, Cade and Schuhmann, Christoph and Schmidt, Ludwig and Jitsev, Jenia},
  booktitle={Proceedings of the IEEE/CVF Conference on Computer Vision and Pattern Recognition (CVPR)},
  pages={2818--2829},
  year={2023}
}

@inproceedings{li2022blip,
  title={{BLIP}: Bootstrapping language-image pre-training for unified vision-language understanding and generation},
  author={Li, Junnan and Li, Dongxu and Xiong, Caiming and Hoi, Steven},
  booktitle={International Conference on Machine Learning (ICML)},
  pages={12888--12900},
  year={2022},
  organization={PMLR}
}

@inproceedings{li2023blip2,
  title={{BLIP-2}: Bootstrapping language-image pre-training with frozen image encoders and large language models},
  author={Li, Junnan and Li, Dongxu and Savarese, Silvio and Hoi, Steven},
  booktitle={International Conference on Machine Learning (ICML)},
  year={2023}
}

@article{li2024llavaonevision,
  title={{LLaVA-OneVision}: Easy visual task transfer},
  author={Li, Bo and Zhang, Yuanhan and Guo, Dong and Zhang, Renrui and Li, Feng and Zhang, Hao and Zhang, Kaichen and Li, Yanwei and Liu, Ziwei and Li, Chunyuan},
  journal={arXiv preprint arXiv:2408.03326},
  year={2024}
}

@inproceedings{chen2024internvl,
  title={Internvl: Scaling up vision foundation models and aligning for generic visual-linguistic tasks},
  author={Chen, Zhe and Wu, Jiannan and Wang, Wenhai and Su, Weijie and Chen, Guo and Xing, Sen and Zhong, Muyan and Zhang, Qinglong and Zhu, Xizhou and Lu, Lewei and others},
  booktitle={Proceedings of the IEEE/CVF conference on computer vision and pattern recognition},
  pages={24185--24198},
  year={2024}
}

@article{bai2023qwenvl,
  title={{Qwen-VL}: A versatile vision-language model for understanding, localization, text reading, and beyond},
  author={Bai, Jinze and Bai, Shuai and Yang, Shusheng and Wang, Shijie and Tan, Sinan and Wang, Peng and Lin, Junyang and Zhou, Chang and Zhou, Jingren},
  journal={arXiv preprint arXiv:2308.12966},
  year={2023}
}

@article{yang2025qwen3,
  title={Qwen3 technical report},
  author={Yang, An and Li, Anfeng and Yang, Baosong and Zhang, Beichen and Hui, Binyuan and Zheng, Bo and Yu, Bowen and Gao, Chang and Huang, Chengen and Lv, Chenxu and others},
  journal={arXiv preprint arXiv:2505.09388},
  year={2025}
}

@article{wang2024qwen2vl,
  title={{Qwen2-VL}: Enhancing vision-language model's perception of the world at any resolution},
  author={Wang, Peng and Bai, Shuai and Tan, Sinan and Wang, Shijie and Fan, Zhihao and Bai, Jinze and Chen, Keqin and Liu, Xuejing and Wang, Jialin and Ge, Wenbin and others},
  journal={arXiv preprint arXiv:2409.12191},
  year={2024}
}

@article{bai2025qwen3vl,
  title={Qwen3-vl technical report},
  author={Bai, Shuai and Cai, Yuxuan and Chen, Ruizhe and Chen, Keqin and Chen, Xionghui and Cheng, Zesen and Deng, Lianghao and Ding, Wei and Gao, Chang and Ge, Chunjiang and others},
  journal={arXiv preprint arXiv:2511.21631},
  year={2025}
}

@misc{qwen2026qwen35,
  title={{Qwen3.5}: Unified vision-language foundation with early-fusion multimodal training},
  author={{Qwen Team}},
  year={2026},
  howpublished={\url{https://github.com/QwenLM/Qwen3.5}},
  note={Technical blog post}
}

@misc{wu2024deepseekvl2,
      title={DeepSeek-VL2: Mixture-of-Experts Vision-Language Models for Advanced Multimodal Understanding},
      author={Zhiyu Wu and Xiaokang Chen and Zizheng Pan and Xingchao Liu and Wen Liu and Damai Dai and Huazuo Gao and Yiyang Ma and Chengyue Wu and Bingxuan Wang and Zhenda Xie and Yu Wu and Kai Hu and Jiawei Wang and Yaofeng Sun and Yukun Li and Yishi Piao and Kang Guan and Aixin Liu and Xin Xie and Yuxiang You and Kai Dong and Xingkai Yu and Haowei Zhang and Liang Zhao and Yisong Wang and Chong Ruan},
      year={2024},
      eprint={2412.10302},
      archivePrefix={arXiv},
      primaryClass={cs.CV},
      url={https://arxiv.org/abs/2412.10302},
}

@inproceedings{
alayrac2022flamingo,
title={Flamingo: a Visual Language Model for Few-Shot Learning},
author={Jean-Baptiste Alayrac and Jeff Donahue and Pauline Luc and Antoine Miech and Iain Barr and Yana Hasson and Karel Lenc and Arthur Mensch and Katherine Millican and Malcolm Reynolds and Roman Ring and Eliza Rutherford and Serkan Cabi and Tengda Han and Zhitao Gong and Sina Samangooei and Marianne Monteiro and Jacob Menick and Sebastian Borgeaud and Andrew Brock and Aida Nematzadeh and Sahand Sharifzadeh and Mikolaj Binkowski and Ricardo Barreira and Oriol Vinyals and Andrew Zisserman and Karen Simonyan},
booktitle={Advances in Neural Information Processing Systems},
editor={Alice H. Oh and Alekh Agarwal and Danielle Belgrave and Kyunghyun Cho},
year={2022},
url={https://openreview.net/forum?id=EbMuimAbPbs}
}

@article{geminiteam2023gemini,
  title={Gemini: A family of highly capable multimodal models},
  author={{Gemini Team, Google}},
  journal={arXiv preprint arXiv:2312.11805},
  year={2023}
}

@article{openai2023gpt4v,
  title={{GPT-4V(ision)} system card},
  author={{OpenAI}},
  journal={OpenAI Technical Report},
  year={2023}
}

@inproceedings{jiang2025vlm2vec,
  title={{VLM2Vec}: Training vision-language models for massive multimodal embedding tasks},
  author={Jiang, Ziyan and Meng, Rui and Yang, Xinyi and Yavuz, Semih and Zhou, Yingbo and Chen, Wenhu},
  booktitle={International Conference on Learning Representations (ICLR)},
  year={2025}
}

@article{meng2025vlm2vecv2,
  title={{VLM2Vec-V2}: Advancing multimodal embedding for videos, images, and visual documents},
  author={Meng, Rui and Jiang, Ziyan and Liu, Ye and Su, Mingyi and Yang, Xinyi and Fu, Yuepeng and Qin, Can and Chen, Zeyuan and Xu, Ran and Xiong, Caiming and others},
  journal={arXiv preprint arXiv:2507.04590},
  year={2025}
}

@article{li2026qwen3vlembedding,
  title={{Qwen3-VL-Embedding} and {Qwen3-VL-Reranker}: A unified framework for state-of-the-art multimodal retrieval and ranking},
  author={Li, Mingxin and Zhang, Yanzhao and Long, Dingkun and Chen, Keqin and Song, Sibo and Bai, Shuai and Yang, Zhibo and Xie, Pengjun and Yang, An and Liu, Dayiheng and Zhou, Jingren and Lin, Junyang},
  journal={arXiv preprint},
  year={2026}
}

@article{wang2025internvl35,
  title={Internvl3. 5: Advancing open-source multimodal models in versatility, reasoning, and efficiency},
  author={Wang, Weiyun and Gao, Zhangwei and Gu, Lixin and Pu, Hengjun and Cui, Long and Wei, Xingguang and Liu, Zhaoyang and Jing, Linglin and Ye, Shenglong and Shao, Jie and others},
  journal={arXiv preprint arXiv:2508.18265},
  year={2025}
}

@article{ruthardt2026steervit,
  title={Steerable Visual Representations},
  author={Ruthardt, Jona and Gaur, Manu and Ramanan, Deva and Tapaswi, Makarand and Asano, Yuki M},
  journal={arXiv preprint arXiv:2604.02327},
  year={2026}
}

@article{liu2025geditbench,
  title={GEditBench v2: A Human-Aligned Benchmark for General Image Editing},
  author={Jiang, Zhangqi and Sun, Zheng and Zeng, Xianfang and Yang, Yufeng and Zhang, Xuanyang and Wu, Yongliang and Cheng, Wei and Yu, Gang and Yang, Xu and Wen, Bihan},
  journal={arXiv preprint arXiv:2603.28547},
  year={2026}
}

@article{deng2025bagel,
  title={Emerging properties in unified multimodal pretraining},
  author={Deng, Chaorui and Zhu, Deyao and Li, Kunchang and Gou, Chenhui and Li, Feng and Wang, Zeyu and Zhong, Shu and Yu, Weihao and Nie, Xiaonan and Song, Ziang and others},
  journal={arXiv preprint arXiv:2505.14683},
  year={2025}
}

@misc{bfl2025flux2,
  title={{FLUX.2}: Frontier visual intelligence},
  author={{Black Forest Labs}},
  year={2025},
  howpublished={\url{https://bfl.ai/blog/flux-2}}
}

@article{meituan2025longcatimage,
  title={{LongCat-Image} technical report},
  author={{Meituan LongCat Team}},
  journal={arXiv preprint arXiv:2512.07584},
  year={2025}
}

@article{wu2025omnigen2,
  title={{OmniGen2}: Towards instruction-aligned multimodal generation},
  author={Wu, Chenyuan and others},
  journal={arXiv preprint arXiv:2506.18871},
  year={2025}
}

@misc{qwen2025imageedit,
  title={{Qwen-Image-Edit}: Image editing with higher quality and efficiency},
  author={{Qwen Team}},
  year={2025},
  howpublished={\url{https://qwenlm.github.io/blog/qwen-image-edit}}
}

@article{liu2025step1xedit,
  title={{Step1X-Edit}: A practical framework for general image editing},
  author={{Step1X-Image Team}},
  journal={arXiv preprint arXiv:2504.17761},
  year={2025}
}

@misc{flux2024,
  title={{FLUX.1}: Open-weight rectified flow transformers for text-to-image generation},
  author={{Black Forest Labs}},
  year={2024},
  howpublished={\url{https://blackforestlabs.ai/announcing-black-forest-labs/}}
}

@article{fang2025fluxreason,
  title={Flux-reason-6m \& prism-bench: A million-scale text-to-image reasoning dataset and comprehensive benchmark},
  author={Fang, Rongyao and Yu, Aldrich and Duan, Chengqi and Huang, Linjiang and Bai, Shuai and Cai, Yuxuan and Wang, Kun and Liu, Si and Liu, Xihui and Li, Hongsheng},
  journal={arXiv preprint arXiv:2509.09680},
  year={2025}
}

@inproceedings{zhang2025iclight,
  title={Scaling in-the-wild training for diffusion-based illumination harmonization and editing by imposing consistent light transport},
  author={Zhang, Lvmin and Rao, Anyi and Agrawala, Maneesh},
  booktitle={International Conference on Learning Representations (ICLR)},
  year={2025}
}

@article{liu2025dreamlight,
  title={{DreamLight}: Towards harmonious and consistent image relighting},
  author={Liu, Yong and Xiao, Wenpeng and Wang, Qianqian and others},
  journal={arXiv preprint arXiv:2506.14549},
  year={2025}
}

@article{chen2023diffharmonization,
  title={Zero-shot image harmonization with generative model prior},
  author={Chen, Jianqi and Zhang, Yilan and Zou, Zhengxia and Chen, Keyan and Shi, Zhenwei},
  journal={IEEE Transactions on Multimedia},
  volume={27},
  pages={4494--4507},
  year={2025},
  publisher={IEEE}
}

@inproceedings{barron2022mipnerf360,
  title={{Mip-NeRF~360}: Unbounded anti-aliased neural radiance fields},
  author={Barron, Jonathan T. and Mildenhall, Ben and Verbin, Dor and Srinivasan, Pratul P. and Hedman, Peter},
  booktitle={Conference on Computer Vision and Pattern Recognition (CVPR)},
  year={2022}
}

@inproceedings{mildenhall2020nerf,
  title={NeRF: Representing Scenes as Neural Radiance Fields for View Synthesis},
  author={Ben Mildenhall and Pratul P. Srinivasan and Matthew Tancik and Jonathan T. Barron and Ravi Ramamoorthi and Ren Ng},
  year={2020},
  booktitle={ECCV},
}

@inproceedings{tancik2023nerfstudio,
  title={{Nerfstudio}: A modular framework for neural radiance field development},
  author={Tancik, Matthew and Weber, Ethan and Ng, Evonne and Li, Ruilong and Yi, Brent and Kerr, Justin and Wang, Terrance and Kristoffersen, Alexander and Austin, Jake and Salahi, Kamyar and Ahuja, Abhik and McAllister, David and Kanazawa, Angjoo},
  booktitle={ACM SIGGRAPH Conference Proceedings},
  year={2023}
}

@article{kerbl20233dgs,
  title={{3D} {Gaussian} splatting for real-time radiance field rendering},
  author={Kerbl, Bernhard and Kopanas, Georgios and Leimk{\"u}hler, Thomas and Drettakis, George},
  journal={ACM Transactions on Graphics (SIGGRAPH)},
  year={2023}
}

@article{mueller2022instantngp,
  title={Instant neural graphics primitives with a multiresolution hash encoding},
  author={M{\"u}ller, Thomas and Evans, Alex and Schied, Christoph and Keller, Alexander},
  journal={ACM Transactions on Graphics (SIGGRAPH)},
  year={2022}
}

@inproceedings{xiang2024trellis,
  title={Structured 3d latents for scalable and versatile 3d generation},
  author={Xiang, Jianfeng and Lv, Zelong and Xu, Sicheng and Deng, Yu and Wang, Ruicheng and Zhang, Bowen and Chen, Dong and Tong, Xin and Yang, Jiaolong},
  booktitle={Proceedings of the IEEE/CVF conference on computer vision and pattern recognition},
  pages={21469--21480},
  year={2025}
}

@article{xu2024instantmesh,
  title={{InstantMesh}: Efficient {3D} mesh generation from a single image with sparse-view large reconstruction models},
  author={Xu, Jiale and Cheng, Weihao and Gao, Yiming and Wang, Xintao and Gao, Shenghua and Shan, Ying},
  journal={arXiv preprint arXiv:2404.07191},
  year={2024}
}

@article{tencent2025hunyuan3d,
  title={{Hunyuan3D 2.1}: From images to high-fidelity {3D} assets with production-ready {PBR} material},
  author={{Tencent Hunyuan3D Team}},
  journal={arXiv preprint arXiv:2506.15442},
  year={2025}
}

@article{stepfun2025step1x3d,
  title={{Step1X-3D}: Towards high-fidelity and controllable generation of textured {3D} assets},
  author={{StepFun Step1X-3D Team}},
  journal={arXiv preprint arXiv:2505.07747},
  year={2025}
}

@article{liu2019roberta,
  title={{RoBERTa}: A robustly optimized {BERT} pretraining approach},
  author={Liu, Yinhan and Ott, Myle and Goyal, Naman and Du, Jingfei and Joshi, Mandar and Chen, Danqi and Levy, Omer and Lewis, Mike and Zettlemoyer, Luke and Stoyanov, Veselin},
  journal={arXiv preprint arXiv:1907.11692},
  year={2019}
}

@article{kuznetsova2020openimages,
  title={The {Open Images Dataset V4}: Unified image classification, object detection, and visual relationship detection at scale},
  author={Kuznetsova, Alina and Rom, Hassan and Alldrin, Neil and Uijlings, Jasper and Krasin, Ivan and Pont-Tuset, Jordi and Kamali, Shahab and Popov, Stefan and Malloci, Matteo and Kolesnikov, Alexander and Duerig, Tom and Ferrari, Vittorio},
  journal={International Journal of Computer Vision (IJCV)},
  year={2020}
}

@inproceedings{loshchilov2017sgdr,
  title={{SGDR}: Stochastic gradient descent with warm restarts},
  author={Loshchilov, Ilya and Hutter, Frank},
  booktitle={International Conference on Learning Representations (ICLR)},
  year={2017}
}

@article{hendrycks2021many,
  title={The Many Faces of Robustness: A Critical Analysis of Out-of-Distribution Generalization},
  author={Dan Hendrycks and Steven Basart and Norman Mu and Saurav Kadavath and Frank Wang and Evan Dorundo and Rahul Desai and Tyler Zhu and Samyak Parajuli and Mike Guo and Dawn Song and Jacob Steinhardt and Justin Gilmer},
  journal={ICCV},
  year={2021}
}

@article{artgan2018,
  title={Improved ArtGAN for Conditional Synthesis of Natural Image and Artwork},
  author={Tan, Wei Ren and Chan, Chee Seng and Aguirre, Hernan and Tanaka, Kiyoshi},
  journal={IEEE Transactions on Image Processing},
  volume    = {28},
  number    = {1},
  pages     = {394--409},
  year      = {2019},
  url       = {https://doi.org/10.1109/TIP.2018.2866698},
  doi       = {10.1109/TIP.2018.2866698}
}

@inproceedings{zhao2021calibrate,
  title={Calibrate before use: Improving few-shot performance of language models},
  author={Zhao, Zihao and Wallace, Eric and Feng, Shi and Klein, Dan and Singh, Sameer},
  booktitle={International conference on machine learning},
  pages={12697--12706},
  year={2021},
  organization={Pmlr}
}

@inproceedings{zheng2024large,
title={Large Language Models Are Not Robust Multiple Choice Selectors},
author={Chujie Zheng and Hao Zhou and Fandong Meng and Jie Zhou and Minlie Huang},
booktitle={International Conference on Learning Representations (ICLR)},
year={2024}
}

@inproceedings{wang2024large,
  title={Large language models are not fair evaluators},
  author={Wang, Peiyi and Li, Lei and Chen, Liang and Cai, Zefan and Zhu, Dawei and Lin, Binghuai and Cao, Yunbo and Kong, Lingpeng and Liu, Qi and Liu, Tianyu and others},
  booktitle={Proceedings of the 62nd Annual Meeting of the Association for Computational Linguistics (Volume 1: Long Papers)},
  pages={9440--9450},
  year={2024}
}

@inproceedings{kumari2026npedit,
  title     = {Learning an Image Editing Model without Image Editing Pairs},
  author    = {Kumari, Nupur and Wang, Sheng-Yu and Zhao, Nanxuan and
               Nitzan, Yotam and Li, Yuheng and Singh, Krishna Kumar and
               Zhang, Richard and Shechtman, Eli and Zhu, Jun-Yan and Huang, Xun},
  booktitle = {ICLR},
  year      = {2026}
}

@article{luo2025dualprocess,
  title   = {Dual-Process Image Generation},
  author  = {Luo, Grace and Granskog, Jonathan and Holynski, Aleksander and Darrell, Trevor},
  journal = {arXiv preprint arXiv:2506.01955},
  year    = {2025}
}

@article{xu2023imagereward,
  title   = {{ImageReward}: Learning and Evaluating Human Preferences
             for Text-to-Image Generation},
  author  = {Xu, Jiazheng and Liu, Xiao and Wu, Yuchen and Tong, Yuxuan and
             Li, Qinkai and Ding, Ming and Tang, Jie and Dong, Yuxiao},
  journal = {arXiv preprint arXiv:2304.05977},
  year    = {2023}
}

@inproceedings{kirstain2023pickapic,
    title={Pick-a-Pic: An Open Dataset of User Preferences for Text-to-Image Generation},
    author={Yuval Kirstain and Adam Polyak and Uriel Singer and Shahbuland Matiana and Joe Penna and Omer Levy},
    booktitle={Neural Information Processing Systems (NeurIPS)},
    year={2023}
}

@article{wu2023human,
  title={Human Preference Score v2: A Solid Benchmark for Evaluating Human Preferences of Text-to-Image Synthesis},
  author={Wu, Xiaoshi and Hao, Yiming and Sun, Keqiang and Chen, Yixiong and Zhu, Feng and Zhao, Rui and Li, Hongsheng},
  journal={arXiv preprint arXiv:2306.09341},
  year={2023}
}

@inproceedings{wu2026editreward,
title={EditReward: A Human-Aligned Reward Model for Instruction-Guided Image Editing},
author={Keming Wu and Sicong Jiang and Max Ku and Ping Nie and Minghao Liu and Wenhu Chen},
booktitle={International Conference on Learning Representations (ICLR)},
year={2026}
}

@article{xue2025dancegrpo,
  title   = {{DanceGRPO}: Unleashing {GRPO} on Visual Generation},
  author  = {Xue, Zeyue and Wu, Jie and Gao, Yu and Kong, Fangyuan and
             Zhu, Lingting and Chen, Mengzhao and Liu, Zhiheng and Liu, Wei and
             Guo, Qiushan and Huang, Weilin and Luo, Ping},
  journal = {arXiv preprint arXiv:2505.07818},
  year    = {2025}
}

@inproceedings{shaham2024maia,
  title     = {A Multimodal Automated Interpretability Agent},
  author    = {Rott Shaham, Tamar and Schwettmann, Sarah and Wang, Franklin and
               Rajaram, Achyuta and Hernandez, Evan and Andreas, Jacob and Torralba, Antonio},
  booktitle = {ICML},
  year      = {2024}
}

@article{surkov2024sae,
  title   = {One-Step is Enough: Sparse Autoencoders for
             Text-to-Image Diffusion Models},
  author  = {Surkov, Viacheslav and Wendler, Chris and Mari, Antonio and
             Terekhov, Mikhail and Deschenaux, Justin and West, Robert and
             Gulcehre, Caglar and Bau, David},
  journal = {arXiv preprint arXiv:2410.22366},
  year    = {2024}
}

@inproceedings{wei2026qare,
  title={Towards Text-Guided Attribute-Disentangled Multimodal Representation Learning},
  author={Wei, Yibing and Katakol, Sudeep and Brack, Manuel and Lin, Jinhong and Bai, Haoyue and Li, Yu-Teng and Zhang, Richard and Shechtman, Eli and Ravi, Hareesh and Kale, Ajinkya},
  booktitle={CVPR},
  year={2026}
}
}

\appendix

\section{Dataset collection and curation}
\label{sec:supp_dataset}

We extend Section~\ref{sec:dataset}  and provide the full curation pipeline behind the synthetic odd-one-out dataset used for training and in-distribution evaluation (Section~\ref{sec:supp_dataset_ooo}), and the vision-algorithm 2AFC set used for \arxiv{evaluating generalization} (Section~\ref{sec:supp_dataset_2afc}). Throughout, we use the term \emph{aspect} for the text condition $\mathbf{c}$ (e.g., \emph{lighting}, \emph{color tone}, \emph{scene geometry}).

\myparagraph{Annotation crowdsourcing.} We collect annotations on Amazon Mechanical Turk and Prolific. Reward per task is set to target an effective hourly rate of \$10/hour. Sentinel checks are applied to every session. A sentinel triplet is constructed by duplicating one image from a real triplet, making the correct odd-one-out visually unambiguous. Sessions that fail the sentinels are returned, and the corresponding responses are excluded.

\myparagraph{Per-aspect ``can't tell'' filter.} To handle the case where the visual aspect condition is ambiguous, our annotation interface includes a ``can't tell'' option in addition to the image choices. We collect 5 annotations per triplet, discard the ``can't tell'' vote, and only train and test on the image votes. To ensure sufficient votes for all visual aspects in the test set, we discard any aspect if 3 or more annotators mark ``can't tell'' in the test set. The filter is applied independently per aspect. A triplet may keep some aspects and lose others, and we drop the triplet only when no aspect remains.

\myparagraph{\arxiv{Data diversity analysis details.}}
\arxiv{In the main text, Figure~\ref{fig:data_diversity} reports dataset diversity along subject and aspect axes. Subject categories are assigned via CLIP image-text matching using eleven broad labels (e.g., person, animal, object) using ``{\menlo A photo of <label>}'' as template. We match the text with the average image embedding of three images for an odd-one-out triplet, or against the reference image for a 2AFC sample. We categorize aspect condition text into different visual properties using a deterministic head-word extraction rule. The deciding head is typically the final token (e.g., \texttt{car paint color} maps to ``Color'' category). We apply exceptions to ignore common suffixes (e.g., extracting \texttt{texture} from \texttt{texture detail} to map to ``Surface'', or \texttt{shadow} from \texttt{shadow strength} to map to ``Lighting''), and to parse count patterns (e.g., mapping phrases like \texttt{number of people at table} or \texttt{cookie count on plate} to an ``Item Count'' category).}

\subsection{Odd-one-out dataset}
\label{sec:supp_dataset_ooo}

\myparagraph{Image generation.} We use the FLUX-Reason-6M~\cite{fang2025fluxreason} caption pool as base prompts. For each base prompt, an instruction-tuned LLM (Qwen3-4B-Instruct-2507~\cite{yang2025qwen3}) authors three controlled prompt variations along a small set of pre-specified axes (e.g., color, lighting, composition), so that within a triplet the three prompts deliberately differ on at least one axis. The three prompts are then rendered by FLUX.1-dev~\cite{flux2024} with a shared random seed, 50 sampling steps, and guidance scale of~$4.0$. Below are the instructions to generate the prompt variation:

\begin{tcolorbox}[
  breakable, enhanced,
  colback=gray!8, colframe=gray!50,
  title={\bfseries\small System Prompt},
  boxrule=0.4pt, left=6pt, right=6pt
]
\begin{Verbatim}[fontsize=\small, breaklines=true, breakanywhere=true, breaksymbolleft={}]
You are an expert at generating perceptual variations for odd-one-out perceptual studies.

TASK: Given a base text prompt, generate THREE CLEARLY DISTINGUISHABLE text-to-image prompts (0, 1, 2) that form an interesting odd-one-out triplet.

GOAL: These images will be shown to humans who must identify which image is "most different" from the others. For each visual FACTOR you vary, TWO images should be SIMILAR and ONE should be the "ODD" one (clearly different).

CRITICAL DESIGN PRINCIPLES:
1. Vary 5-7 visual factors across the triplet
2. For EACH factor: two images share a similar value, one image is distinctly different
3. DISTRIBUTE the "odd" role across images - don't make one image odd for all factors!
   - Example: Image 0 might be odd for "lighting" and "pose"
   - Image 1 might be odd for "background" and "texture"  
   - Image 2 might be odd for "camera angle" and "color"
4. This creates an INTERESTING perceptual puzzle where each image is "odd" in some ways but "similar" in others

VISUAL FACTORS TO VARY (use contextual names, not category names):
   - Color & Illumination → "lighting warmth", "shadow direction", "brightness level"
   - Material & Texture → "sofa fabric", "surface glossiness", "wood grain"
   - Geometry & Shape & Pose → "cat pose", "flower arrangement", "building height"
   - Background & Environment → "sky condition", "room clutter", "weather"
   - Composition & Spatial Arrangement → "object spacing", "framing tightness"
   - Camera → "camera distance", "viewing angle", "depth of field"
   - Photographic Style → "color saturation", "film grain", "contrast level"

REQUIREMENTS:
1. All three must describe the same subject(s) or scene
2. Use VIVID, SPECIFIC descriptors - avoid "slightly" or "somewhat"
3. Each prompt should produce an image DISTINGUISHABLE AT A GLANCE
4. Factor names must be contextual: "cat pose" not "Geometry & Shape"
5. Keep factor names short and understandable by non-experts

EXAMPLE for "A golden retriever in a park":
{
  "metadata": {
    "num_variations": 3,
    "factors": [
      {"name": "dog pose", "odd_image": 0, "similar_value": "sitting upright", "odd_value": "lying down"},
      {"name": "lighting", "odd_image": 1, "similar_value": "bright sunny", "odd_value": "overcast diffuse"},
      {"name": "background trees", "odd_image": 2, "similar_value": "lush green", "odd_value": "autumn orange"},
      {"name": "grass length", "odd_image": 0, "similar_value": "short manicured", "odd_value": "tall wild"},
      {"name": "camera angle", "odd_image": 1, "similar_value": "eye level", "odd_value": "low angle looking up"},
      {"name": "depth of field", "odd_image": 2, "similar_value": "sharp throughout", "odd_value": "blurred background"}
    ]
  },
  "triplets": [
    {"id": 0, "prompt_or_edit_or_video_instruction": "A golden retriever lying down in tall wild grass in a sunny park with lush green trees, eye level shot, sharp focus throughout, bright natural lighting"},
    {"id": 1, "prompt_or_edit_or_video_instruction": "A golden retriever sitting upright in short manicured grass in a park under overcast diffuse light, low angle looking up, lush green trees, sharp focus throughout"},
    {"id": 2, "prompt_or_edit_or_video_instruction": "A golden retriever sitting upright in short manicured grass in a sunny park with autumn orange trees, eye level shot, blurred background, bright natural lighting"}
  ]
}

OUTPUT FORMAT: Return ONLY valid JSON with the structure shown above.
Remember: Make each image "odd" for roughly 2-3 factors, creating an interesting perceptual puzzle!
\end{Verbatim}
\end{tcolorbox}

\myparagraph{Aspect proposal and refinement.} Each triplet inherits the proposed aspects from its prompt-variation step. Because the prompt-to-image mapping is not faithful, we additionally pass the rendered triplet and its candidate aspect list to GPT-5.2~\cite{openai2023gpt4v} and ask it to drop aspects that are not visibly varying across the three images and to add visually salient aspects that the prompts did not specify. The refinement instruction is:

\begin{tcolorbox}[
  breakable, enhanced,
  colback=gray!8, colframe=gray!50,
  title={\bfseries\small System Prompt},
  boxrule=0.4pt, left=6pt, right=6pt
]
\begin{Verbatim}[fontsize=\small, breaklines=true, breakanywhere=true, breaksymbolleft={}]
You are analyzing three horizontally concatenated images (left, middle, right).

INITIAL AXES OF VARIATION:
<list of aspects inserted here>

TASK: Refine the initial axes based on the images. Keep most axes, remove only irrelevant ones, and add any new axes for visual differences that are actually present.

RULES (must be strictly followed):
1. Each axis must be a single, concise noun phrase.
2. Do NOT use the words "and", "or", "not", parentheses, asterisks, commas, or any other symbols in an axis.
3. Always name axes contextually for the subject (e.g., "cat pose", "sky brightness", "table material").
4. Always include axes for pose, orientation, geometry, size, or viewpoint if subjects vary.
5. Include axes for color, lighting, material, texture, background, composition, camera, or photographic style if relevant.
6. Output only axes that actually vary between the images.
7. Do not hallucinate axes that are not visually evident.
8. Provide 5-10 axes sorted by importance, most important first.

OUTPUT FORMAT: Numbered list
1. [Most important axis]
2. [Second axis]
3. ...
\end{Verbatim}
\end{tcolorbox}

The output is an over-complete refined aspect list, which is then pruned by the can't-tell filter from human annotation.

\begin{figure}[h]
    \centering
    \includegraphics[width=1.0\linewidth]{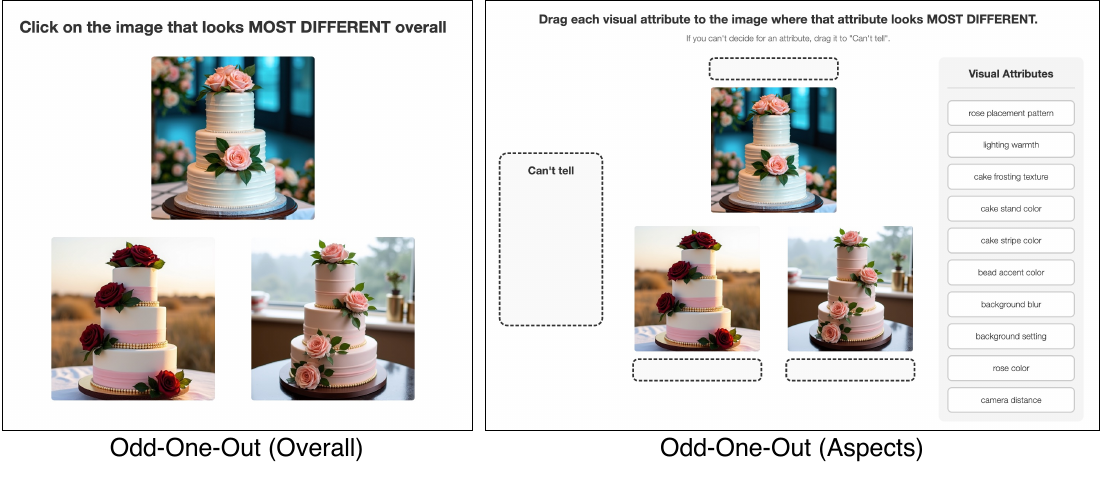}
    \caption{\textbf{Odd-one-out annotation interface.} Annotators are shown three images. (Left) Annotators click the most different-looking image overall. (Right) Annotators are given a list of aspects as draggable chips, and assign each aspect either to the image they perceive as the odd-one-out or to ``can't tell''.}
    \label{fig:supp_ui_ooo}
\end{figure}

\myparagraph{Annotation interface.} Annotators see three triangle-arranged images and a list of refined aspects rendered as draggable chips. For each aspect, they drag the chip onto the image they consider the odd-one-out, or onto a ``can't tell'' bin (Figure~\ref{fig:supp_ui_ooo}). Each triplet is annotated by 5 workers; after the can't-tell filter, the remaining votes form a categorical distribution $\mathbf{y} \in \Delta^{3}$ over the three image positions. We additionally include one ``overall'' aspect per triplet by asking an unconditional question ``which image looks most different overall?''. We treat ``overall'' as a regular aspect condition during training and evaluation.

\begin{table}[h]
\caption{\textbf{Algorithms and aspects used in the 2AFC set.} We report the algorithms used for our 2AFC set and the number of triplet occurrences for each algorithm, along with the aspect list used.}
\label{tab:supp_2afc_algos}
\centering
\footnotesize
\setlength{\tabcolsep}{6pt}
\renewcommand{\arraystretch}{1.15}
\begin{tabular}{@{}>{\raggedright\arraybackslash}m{0.1\linewidth} >{\raggedright\arraybackslash}m{0.43\linewidth} >{\raggedright\arraybackslash}m{0.42\linewidth}@{}}
\toprule
\textbf{Source} & \textbf{Algorithm (\# triplets it appears in)} & \textbf{Aspect list} \\
\midrule
Image editing      & BAGEL (67), FLUX2-dev (83), LongCat-Image-Edit (72), OmniGen2 (92), Qwen-Image-Edit (64), Qwen-Image-Edit-2511 (69), Step1X-Edit~v1.2 (65) & 6--10 aspects predicted by GPT-5.2 per sample and human-pruned  \\ \hdashline

Image compositing     & IC-Light (197), DreamLight (195), Diff-Harmonization (196) & \textit{(fg-ref)} facial features, hair texture, clothing detail; \textit{(bg-ref)} lighting direction, lighting warmth, color tone, illumination \\ \hdashline

Novel view synthesis  & Splatfacto (120), Nerfacto (120), Instant-NGP (120) & scene geometry, scene sharpness, texture detail, color tone, lighting brightness \\ \hdashline

Single-image 3D       & TRELLIS.2-4B (197), InstantMesh (199), Hunyuan3D-2.1 (199), Step1X-3D (199) & 3D shape, surface texture, material, color tone, camera pose, object size \\
\bottomrule
\end{tabular}
\end{table}

\subsection{\arxiv{External-algorithm 2AFC} evaluation set}
\label{sec:supp_dataset_2afc}

\begin{figure}[t]
    \centering
    \includegraphics[width=1.0\linewidth]{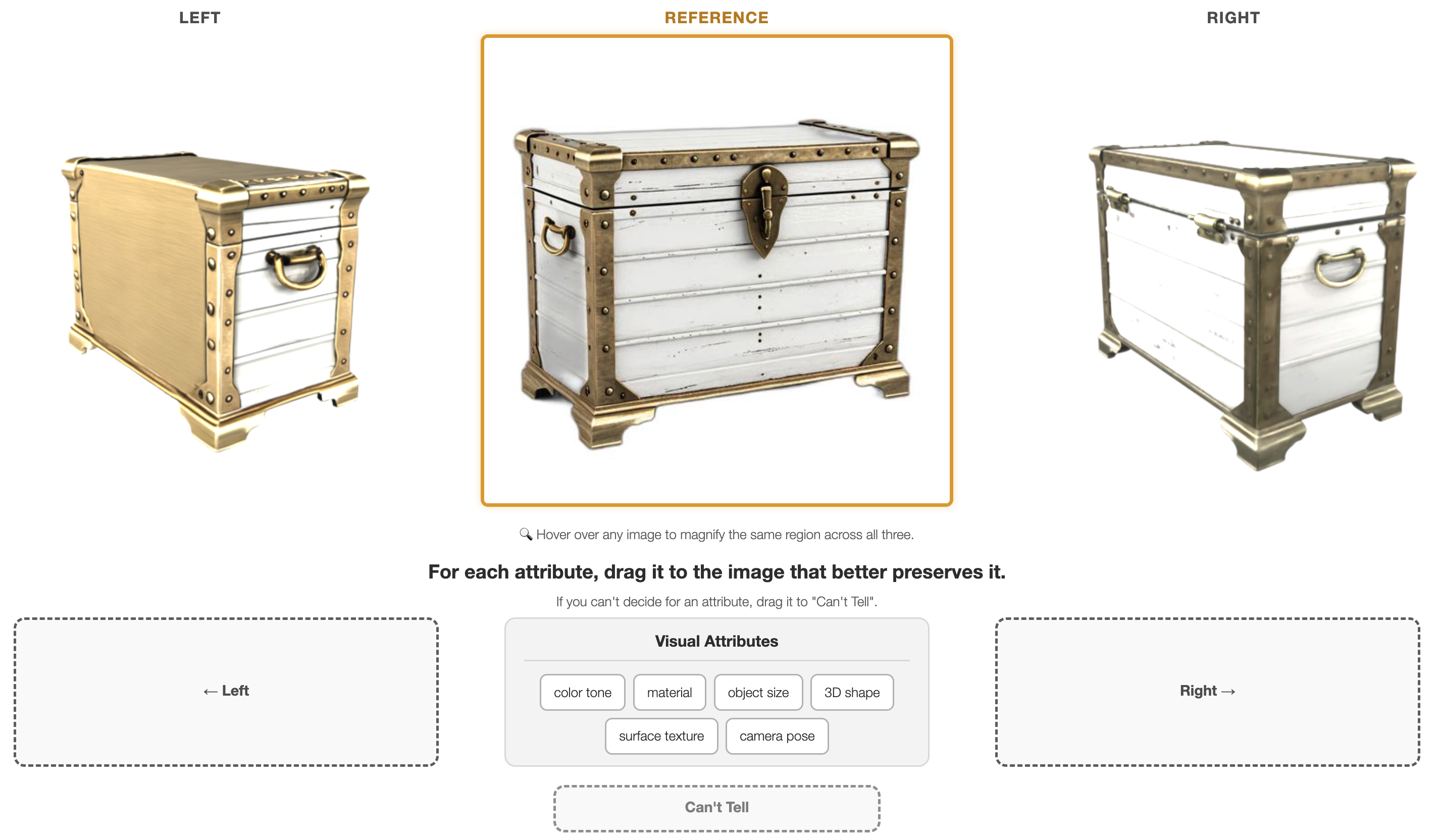}
    \caption{\textbf{2AFC annotation interface.} Annotators see a reference image (center) and two candidates on two sides and select, for each aspect, which candidate is more similar to the reference, or ``can't tell''.}
    \label{fig:supp_ui_2afc}
\end{figure}

The 2AFC set is curated from outputs of real vision algorithms across four sources: image editing, image compositing, novel view synthesis (NVS), and single-image 3D. Each source contributes a reference image $\mathbf{x}_\text{ref}$ and outputs from multiple algorithms; a triplet pairs $\mathbf{x}_\text{ref}$ with two outputs from \emph{different} algorithms, and the annotator picks the output more similar to $\mathbf{x}_\text{ref}$ for each aspect. Algorithms and aspect lists per source are summarized in Table~\ref{tab:supp_2afc_algos}.

\myparagraph{Image editing.} Sources are drawn from GEditBench v2~\cite{liu2025geditbench}, which includes 16 methods. We keep the 7 methods: BAGEL~\cite{deng2025bagel}, FLUX2-dev~\cite{bfl2025flux2}, LongCat-Image-Edit~\cite{meituan2025longcatimage}, OmniGen2~\cite{wu2025omnigen2}, Qwen-Image-Edit~\cite{qwen2025imageedit}, Qwen-Image-Edit-2511~\cite{qwen2025imageedit}, Step1X-Edit~v1.2~\cite{liu2025step1xedit}. The rest are dropped either because they crop or resize the input on at least some samples or because they are near-duplicate intermediate releases of one of the kept models. We further discard four edit categories from GEditBench v2 (chart editing, in-image text translation, text editing, and line-to-image) since they contain visual content out of our interest (e.g., text renders, charts, line sketches). For each sample, GPT-5.2 names the \emph{target} aspect that the edit instruction explicitly changes (e.g., \emph{background setting} for the instruction ``replace the background with a modern office'') and 6--10 \emph{preserved} aspects that should remain identical to the reference; both lists are then human-pruned. Both target and preserved aspects are kept in the annotation list. Each triplet pairs the source photograph with two edits from different, randomly selected methods of the same instruction. We drop triplets whose human majority votes are the same image for every aspect, since these are usually dominated by one method barely modifying the input.

\myparagraph{Image compositing.} We use 10 foreground images of portraits crossed with 30 background reference photographs and run three methods on each pair: IC-Light (text-foreground-background-conditioned model)~\cite{zhang2025iclight}, DreamLight~\cite{liu2025dreamlight}, and zero-shot Diff-Harmonization~\cite{chen2023diffharmonization}. Each foreground-background pair yields two reference modes: \emph{foreground-reference} (the foreground is $\mathbf{x}_\text{ref}$ and the aspects describe subject preservation) and \emph{background-reference} (the background photograph is $\mathbf{x}_\text{ref}$, and the aspects describe background preservation). We take random subsets from this collection to ensure the dataset size matches the other tasks.

\myparagraph{Novel view synthesis.} We use 6 scenes from Mip-NeRF~360~\cite{barron2022mipnerf360} and 3 nerfstudio methods~\cite{tancik2023nerfstudio}: Splatfacto (3D Gaussian Splatting~\cite{kerbl20233dgs}), Nerfacto~\cite{mildenhall2020nerf}, and Instant-NGP~\cite{mueller2022instantngp}. All methods receive the same training views and render the same held-out poses; the reference image is the held-out ground-truth photograph at the test pose.

\myparagraph{Single-image 3D.} We use inputs from 3D Arena and 4 methods: TRELLIS.2~\cite{xiang2024trellis}, InstantMesh~\cite{xu2024instantmesh}, Hunyuan3D-2.1~\cite{tencent2025hunyuan3d}, and Step1X-3D~\cite{stepfun2025step1x3d}. Each method's mesh is re-rendered in Blender at 4 spaced-out azimuth angles per object; the reference image is the source photograph. Since there is no consistent canonical view for each method, camera poses from each method are randomly different from the reference. Hence, we also annotate and evaluate camera pose similarity in this setting.

\myparagraph{Annotation interface.} Each triplet is presented as $[\mathbf{x}_\text{cand A} \mid \mathbf{x}_\text{ref} \mid \mathbf{x}_\text{cand B}]$, with the reference in the center. Per triplet, annotators are shown the aspect list and select, for each aspect, which candidate is more similar to the reference, or ``can't tell'' (Figure~\ref{fig:supp_ui_2afc}). Each (triplet, aspect) is annotated by 5 workers, and we apply the ``can't-tell'' filter as before. Following Section~\ref{sec:dataset_2afc}, we additionally drop aspects whose post-filter vote is exactly tied. Unlike the odd-one-out evaluation, the 2AFC evaluation does not include an ``overall'' aspect: every comparison is explicitly conditioned on a visual aspect.

\section{Additional results}
\label{sec:supp_additional_results}

Extending Section~\ref{sec:experiments}, we provide \arxiv{additional generalization results in Section~\ref{sec:supp_generalization},} additional analysis in Section~\ref{sec:supp_reading_table}, and additional ablations in Section~\ref{sec:supp_ablations}.

\begin{table}
\centering
\caption{\textbf{Generalization to generic perceptual similarity:} \emph{BAPPS}~\cite{zhang2018lpips} and \emph{NIGHTS}~\cite{fu2023dreamsim} on low and mid-level similarity, respectively. Standard errors are in {\color{gray}{gray}}. Best is \textbf{bold}, runner-up is \underline{underlined}. Ours is a fine-tuned late fusion model from Qwen3-VL-8B-Emb., which shows consistent improvement, even without training on either dataset directly. Our dataset subset with only overall conditions reported for comparison.}
\label{tab:bapps_nights}
\begin{tabular}{lccc}
\toprule
\multirow{2}{*}{Method} & \multicolumn{3}{c}{Dataset} \\ 
\cmidrule(lr){2-4} 
& BAPPS & NIGHTS & Ours--overall \\
\midrule
LPIPS            & \textbf{68.4}{\color{gray}\scriptsize$\pm$0.2} & 76.4{\color{gray}\scriptsize$\pm$1.0}  & 53.9{\color{gray}\scriptsize$\pm$0.8}\\
DreamSim         & \underline{68.3}{\color{gray}\scriptsize$\pm$0.2} & \textbf{96.2}{\color{gray}\scriptsize$\pm$0.5}  & \underline{65.1}{\color{gray}\scriptsize$\pm$0.7}\\
Qwen3-VL-8B-Emb. & 65.2{\color{gray}\scriptsize$\pm$0.2} & 87.9{\color{gray}\scriptsize$\pm$0.8}  & 60.1{\color{gray}\scriptsize$\pm$0.7}\\ 
\cdashline{1-4} 
Ours    & 67.3{\color{gray}\scriptsize$\pm$0.2} & \underline{94.0}{\color{gray}\scriptsize$\pm$0.6}  & \textbf{66.5}{\color{gray}\scriptsize$\pm$0.7}\\
\bottomrule
\end{tabular}
\end{table}

\begin{table}
\centering
\caption{\arxiv{\textbf{Generalization to compositional image retrieval:} GeneCIS~\cite{vaze2023genecis} Focus Attribute and Focus Object tasks. Best is \textbf{bold}, runner-up is \underline{underlined}. CRL~\cite{liu2025conditional} reports results only on the object-conditioned GeneCIS tasks and thus does not report Focus Attribute.}}
\label{tab:genecis}
\begin{tabular}{lcccccc}
\toprule
\multirow{2}{*}{Method} & \multicolumn{3}{c}{Focus Attribute} & \multicolumn{3}{c}{Focus Object} \\
\cmidrule(lr){2-4} \cmidrule(lr){5-7}
& R@1 & R@2 & R@3 & R@1 & R@2 & R@3 \\
\midrule
GeneCIS
& 19.0
& 31.0
& 41.5
& 14.7
& 25.9
& 36.1 \\

CRL (Pretrained)
& -- & -- & --
& 15.4
& 26.7
& 35.8 \\

CRL (Fine-tuned)
& -- & -- & --
& \underline{19.7}
& \textbf{32.7}
& \textbf{41.3} \\

Qwen3-VL-8B-Emb.
& \underline{30.85}
& \underline{46.10}
& \underline{56.20}
& 18.11
& 28.67
& 36.22 \\

\cdashline{1-7}
TPIPS
& \textbf{31.35}
& \textbf{46.95}
& \textbf{57.25}
& \textbf{19.95}
& \underline{30.61}
& \underline{38.16} \\
\bottomrule
\end{tabular}
\end{table}

\begin{table*}[t]
\centering
\caption{\textbf{Generalization to QARE-Bench (Synthetic)~\cite{wei2026qare} dataset}. The benchmark measures disentanglement on objects, background, and style on a small set of 48 images. \arxiv{Best is \textbf{bold}, runner-up is \underline{underlined}.} Our method is competitive with the provided benchmarks from QARE, with the highest mAP among 8B models and close to the 12B model.}
\resizebox{\textwidth}{!}{
\begin{tabular}{ccllrrrrrrr}
\toprule

\multirow{2}{*}{\shortstack[c]{\textbf{Post-trained/}\\\textbf{Zero-shot}}} &
\bf \multirow{2}{*}{\shortstack[c]{\textbf{Query-}\\\textbf{able}}} & \multicolumn{3}{c}{\bf Model} & \multicolumn{4}{c}{\textbf{mAP ($\uparrow$)}} & \multirow{2}{*}{\bf AIS ($\downarrow$)} \\ \cmidrule(lr){3-5} \cmidrule(lr){6-9}
& & \bf Method &
\bf Backbone &
\bf Params &
\bf obj &
\bf sty &
\bf bg &
\bf all &
 \\

\midrule

\multirow{4}{*}{\shortstack[l]{Zero-shot}}
& \multirow{4}{*}{\xmark}

& \multirow{4}{*}{\shortstack[l]{Vision\\Encoder}}
& CLIP
& --
& 9.4 & 13.1 & 8.8 & 4.5 & 1.00 \\

&
&
& SigLIP
& --
& 10.0 & 11.0 & 10.1 & 4.4 & 1.00 \\

&
&
& DINOv2
& --
& 13.5 & 6.8 & 10.0 & 4.2 & 1.00 \\

&
&
& DINOv3
& --
& 12.1 & 7.1 & 11.2 & 4.1 & 1.00 \\

\midrule

\multirow{11}{*}{\shortstack[l]{Zero-shot}}
& \multirow{11}{*}{\cmark}
& \multirow{11}{*}{TF-QARE~\cite{wei2026qare}}

& Qwen2-VL
& 2B
& 8.7 & 20.5 & 37.1 & 22.1 & \underline{0.63} \\

&
&
& Qwen2-VL
& 7B
& 69.7 & 73.9 & \textbf{91.7} & 78.4 & 0.68 \\

\cdashline{4-10}

&
&
& Qwen2.5-VL
& 3B
& 38.7 & 45.6 & \underline{91.5} & 58.6 & 0.78 \\

&
&
& Qwen2.5-VL
& 7B
& \underline{83.9} & 56.9 & 90.1 & 77.0 & 0.73 \\

&
&
& Qwen2.5-VL
& 32B
& 79.0 & 55.2 & \textbf{91.7} & 75.3 & 0.81 \\

\cdashline{4-10}

&
&
& InternVL3
& 1B
& 47.8 & 23.5 & 65.6 & 45.6 & 0.74 \\

&
&
& InternVL3
& 2B
& 46.9 & 58.0 & 90.2 & 65.0 & 0.75 \\

&
&
& InternVL3
& 8B
& 78.0 & 56.8 & \textbf{91.7} & 75.5 & \textbf{0.55} \\

&
&
& InternVL3
& 14B
& \textbf{85.8} & 55.4 & \textbf{91.7} & 77.6 & 0.78 \\

\cdashline{4-10}

&
&
& Gemma3
& 4B
& 55.6 & 70.4 & 83.9 & 70.0 & 0.88 \\

&
&
& Gemma3
& 12B
& 82.9 & 75.4 & \textbf{91.7} & \underline{83.3} & 0.88 \\

\midrule

\multirow{3}{*}{\shortstack[l]{Post-trained}}
& \multirow{3}{*}{\cmark}

& VLM2VecV1~\cite{jiang2025vlm2vec}
& Qwen2-VL
& 7B
& 8.9 & 29.6 & 11.6 & 16.7 & 0.97 \\

&
& VLM2VecV2~\cite{meng2025vlm2vecv2}
& Qwen2-VL
& 2B
& 7.9 & 27.0 & 11.2 & 15.4 & 0.82 \\

&
& Qwen3-VL-Emb
& Qwen3-VL-Emb
& 8B
& 56.5 & \textbf{90.6} & 89.8 & 78.9 & 0.64 \\

&
& \textbf{Ours}
& Qwen3-VL-Emb
& 8B
& 81.6 & \underline{81.4} & \textbf{91.7} & \textbf{84.9} & \underline{0.63} \\

\bottomrule
\end{tabular}
}
\label{tab:qare_main}
\end{table*}

\subsection{\arxiv{Generalization to other benchmarks}}
\label{sec:supp_generalization}
\arxiv{To test whether our method generalizes beyond the training distribution, we provide additional evaluations as follows.}

\myparagraph{\arxiv{Generic perceptual similarity.}}
Our proposed method extends image similarity to aspect-conditioning. Here, we test how well the method performs on generic image similarity on previous perceptual similarity datasets. In Table~\ref{tab:bapps_nights}, we report results on BAPPS~\cite{zhang2018lpips} and NIGHTS~\cite{fu2023dreamsim}, two established perceptual similarity benchmarks designed to train and evaluate LPIPS~\cite{zhang2018lpips} and DreamSim~\cite{fu2023dreamsim}, respectively. Unlike our setting, these benchmarks are not aspect-conditioned and collect only overall perceptual similarities.  Notably, training exclusively on our dataset and inferencing with the ``overall'' condition yields consistent improvements over the base VLM across both benchmarks, closing much of the gap to methods trained in-domain. This suggests that our dataset captures perceptual structure that transfers to other domains.

\myparagraph{\arxiv{Conditional retrieval on natural images.}}
\arxiv{To test whether our method generalizes to real-world natural images, we evaluate our method on compositional image retrieval using the Focus Attribute and Focus Object subsets of the GeneCIS benchmark~\cite{vaze2023genecis}, which features complex, real-world scenes.
While Focus Object images are evaluated at native resolution, Focus Attribute images are cropped by bounding boxes that are sometimes too small, so we dilate the annotated bounding boxes by 0.3 and upscale the crops to a minimum of 256 vision tokens. We use the annotated attribute (e.g., \texttt{color}) for Focus Attribute, and the phrase \texttt{scene context and presence of <object>} for Focus Object. Our method doesn't apply to the GeneCIS ``Change'' subsets because they focus on edit transformations. Table~\ref{tab:genecis} shows the results. Without any GeneCIS fine-tuning or score adjustments, our embedding method improves over its Qwen3-VL-Embedding-8B backbone at every cutoff and outperforms both GeneCIS and training-free CRL~\cite{liu2025conditional}. It also performs on par with fine-tuned CRL that is directly tuned on GeneCIS, which is the state-of-the-art on this benchmark.}

\myparagraph{\arxiv{Attribute-specific retrieval.}}
We also compare to a previous dataset, Queryable Attribute Representation Extraction (QARE)~\cite{wei2026qare} in Tab.~\ref{tab:qare_main}, which creates a small synthetic dataset of 4 different objects, rendered on 4 backgrounds and 3 artistic styles with known variations, resulting in 48 images. The benchmark queries for these 3 attributes, and measures the recall in mAP and the tightness of the embeddings in AIS. \arxiv{We ran our embedding method on this benchmark, using conditions \texttt{main subject type}, \texttt{background scene}, and \texttt{artistic medium} for object, background, and artistic style, respectively.} Though not designed for this dataset, our method performs competitively with the provided baselines.

\begin{table}[t]
\caption{\textbf{Agreement with humans.} We show the best per-task type in \textbf{bold} and second-best \underline{underlined}; gray values are standard errors across test examples. For the triplet task, our method variant is explicitly trained on odd-one-out tasks and does not apply to 2AFC tests.
}
\label{tab:big_results}
{\small
\centering
\resizebox{1.\linewidth}{!}{
\begin{tabular}{llrcccccc}
\toprule
\multirow{2}{*}{Type} & \multirow{2}{*}{Family} & \multirow{2}{*}{Size} & \multirow{2}{*}{\makecell{Odd-One-Out\\(in-distribution)}} & \multicolumn{5}{c}{\arxiv{External-Algorithm 2AFC}} \\
\cmidrule(lr){5-9}
 &  &  &  & Editing & Composite & NVS & Im-to-3D & Total \\
\midrule
\multirow{2}{*}{Classic} & LPIPS & 62M & \underline{46.2}{\color{gray}\scriptsize$\pm$0.2} & \underline{62.2}{\color{gray}\scriptsize$\pm$1.0} & \textbf{69.4}{\color{gray}\scriptsize$\pm$1.3} & \textbf{96.9}{\color{gray}\scriptsize$\pm$0.5} & \underline{65.9}{\color{gray}\scriptsize$\pm$1.0} & \underline{69.5}{\color{gray}\scriptsize$\pm$0.6} \\
 & DreamSim & 258M & \textbf{50.5}{\color{gray}\scriptsize$\pm$0.2} & \textbf{63.7}{\color{gray}\scriptsize$\pm$1.0} & \underline{67.0}{\color{gray}\scriptsize$\pm$1.4} & \underline{96.6}{\color{gray}\scriptsize$\pm$0.5} & \textbf{75.8}{\color{gray}\scriptsize$\pm$0.9} & \textbf{72.9}{\color{gray}\scriptsize$\pm$0.5} \\
\midrule
\multirow{13}{*}{Triplet} & GPT-5.4 & -- & 57.6{\color{gray}\scriptsize$\pm$0.2} & \underline{69.4}{\color{gray}\scriptsize$\pm$1.0} & \textbf{78.6}{\color{gray}\scriptsize$\pm$1.1} & \textbf{96.6}{\color{gray}\scriptsize$\pm$0.5} & \underline{78.5}{\color{gray}\scriptsize$\pm$0.8} & \textbf{77.9}{\color{gray}\scriptsize$\pm$0.5} \\
 & Gemini-3.1-Pro & -- & \underline{62.0}{\color{gray}\scriptsize$\pm$0.2} & \textbf{70.6}{\color{gray}\scriptsize$\pm$0.9} & \underline{74.6}{\color{gray}\scriptsize$\pm$1.2} & \underline{96.1}{\color{gray}\scriptsize$\pm$0.6} & 74.4{\color{gray}\scriptsize$\pm$0.9} & \underline{76.1}{\color{gray}\scriptsize$\pm$0.5} \\
\cdashline{2-9}[0.5pt/2pt]
 & \multirow{5}{*}{InternVL3.5} & 1B & 38.9{\color{gray}\scriptsize$\pm$0.2} & 50.6{\color{gray}\scriptsize$\pm$1.1} & 50.8{\color{gray}\scriptsize$\pm$1.5} & 54.5{\color{gray}\scriptsize$\pm$1.9} & 51.0{\color{gray}\scriptsize$\pm$1.1} & 51.3{\color{gray}\scriptsize$\pm$0.6} \\
 &  & 2B & 40.0{\color{gray}\scriptsize$\pm$0.2} & 52.7{\color{gray}\scriptsize$\pm$1.1} & 63.0{\color{gray}\scriptsize$\pm$1.4} & 66.2{\color{gray}\scriptsize$\pm$1.8} & 56.7{\color{gray}\scriptsize$\pm$1.1} & 57.7{\color{gray}\scriptsize$\pm$0.6} \\
 &  & 4B & 53.9{\color{gray}\scriptsize$\pm$0.2} & 60.4{\color{gray}\scriptsize$\pm$1.0} & 63.6{\color{gray}\scriptsize$\pm$1.4} & 77.1{\color{gray}\scriptsize$\pm$1.6} & 71.5{\color{gray}\scriptsize$\pm$0.9} & 67.1{\color{gray}\scriptsize$\pm$0.6} \\
 &  & 8B & 53.3{\color{gray}\scriptsize$\pm$0.2} & 61.9{\color{gray}\scriptsize$\pm$1.0} & 69.3{\color{gray}\scriptsize$\pm$1.3} & 92.4{\color{gray}\scriptsize$\pm$0.9} & 68.2{\color{gray}\scriptsize$\pm$1.0} & 69.6{\color{gray}\scriptsize$\pm$0.6} \\
 &  & 14B & 56.0{\color{gray}\scriptsize$\pm$0.2} & 63.3{\color{gray}\scriptsize$\pm$1.0} & 71.5{\color{gray}\scriptsize$\pm$1.3} & 86.5{\color{gray}\scriptsize$\pm$1.2} & 76.6{\color{gray}\scriptsize$\pm$0.9} & 72.5{\color{gray}\scriptsize$\pm$0.6} \\
\cdashline{2-9}[0.5pt/2pt]
 & \multirow{2}{*}{LLAVA-OV} & 0.5B & 45.2{\color{gray}\scriptsize$\pm$0.2} & 53.7{\color{gray}\scriptsize$\pm$1.1} & 51.3{\color{gray}\scriptsize$\pm$1.5} & 68.5{\color{gray}\scriptsize$\pm$1.8} & 51.1{\color{gray}\scriptsize$\pm$1.1} & 54.4{\color{gray}\scriptsize$\pm$0.6} \\
 &  & 7B & 53.9{\color{gray}\scriptsize$\pm$0.2} & 62.7{\color{gray}\scriptsize$\pm$1.0} & 69.7{\color{gray}\scriptsize$\pm$1.3} & 86.9{\color{gray}\scriptsize$\pm$1.2} & 77.1{\color{gray}\scriptsize$\pm$0.9} & 72.2{\color{gray}\scriptsize$\pm$0.6} \\
\cdashline{2-9}[0.5pt/2pt]
 & \multirow{3}{*}{Qwen3-VL} & 2B & 47.8{\color{gray}\scriptsize$\pm$0.2} & 58.4{\color{gray}\scriptsize$\pm$1.1} & 64.1{\color{gray}\scriptsize$\pm$1.4} & 91.5{\color{gray}\scriptsize$\pm$1.0} & 68.3{\color{gray}\scriptsize$\pm$1.0} & 67.3{\color{gray}\scriptsize$\pm$0.6} \\
 &  & 4B & 56.1{\color{gray}\scriptsize$\pm$0.2} & 64.0{\color{gray}\scriptsize$\pm$1.0} & 67.1{\color{gray}\scriptsize$\pm$1.4} & 95.4{\color{gray}\scriptsize$\pm$0.7} & 77.3{\color{gray}\scriptsize$\pm$0.9} & 73.4{\color{gray}\scriptsize$\pm$0.5} \\
 &  & 8B & 57.1{\color{gray}\scriptsize$\pm$0.2} & 63.8{\color{gray}\scriptsize$\pm$1.0} & 68.1{\color{gray}\scriptsize$\pm$1.3} & 95.1{\color{gray}\scriptsize$\pm$0.7} & \textbf{78.6}{\color{gray}\scriptsize$\pm$0.8} & 73.9{\color{gray}\scriptsize$\pm$0.5} \\
\cdashline{2-9}[0.5pt/2pt]
 & \textbf{Ours} & 8B & \textbf{65.0}{\color{gray}\scriptsize$\pm$0.2} & -- & -- & -- & -- & -- \\
\midrule
\multirow{13}{*}{Early Fusion} & GPT-5.4 & -- & 55.5{\color{gray}\scriptsize$\pm$0.2} & \underline{67.0}{\color{gray}\scriptsize$\pm$1.0} & \underline{73.8}{\color{gray}\scriptsize$\pm$1.2} & \textbf{96.8}{\color{gray}\scriptsize$\pm$0.5} & 71.3{\color{gray}\scriptsize$\pm$1.0} & \underline{73.7}{\color{gray}\scriptsize$\pm$0.5} \\
 & Gemini-3.1-Pro & -- & \underline{58.4}{\color{gray}\scriptsize$\pm$0.2} & 65.5{\color{gray}\scriptsize$\pm$1.0} & 66.5{\color{gray}\scriptsize$\pm$1.4} & 90.8{\color{gray}\scriptsize$\pm$1.0} & 70.1{\color{gray}\scriptsize$\pm$1.0} & 70.7{\color{gray}\scriptsize$\pm$0.6} \\
\cdashline{2-9}[0.5pt/2pt]
 & \multirow{5}{*}{InternVL3.5} & 1B & 45.4{\color{gray}\scriptsize$\pm$0.2} & 57.2{\color{gray}\scriptsize$\pm$1.1} & 69.8{\color{gray}\scriptsize$\pm$1.3} & 88.7{\color{gray}\scriptsize$\pm$1.1} & 63.2{\color{gray}\scriptsize$\pm$1.0} & 65.8{\color{gray}\scriptsize$\pm$0.6} \\
 &  & 2B & 46.3{\color{gray}\scriptsize$\pm$0.2} & 61.5{\color{gray}\scriptsize$\pm$1.0} & 70.7{\color{gray}\scriptsize$\pm$1.3} & 82.3{\color{gray}\scriptsize$\pm$1.4} & 72.6{\color{gray}\scriptsize$\pm$0.9} & 69.8{\color{gray}\scriptsize$\pm$0.6} \\
 &  & 4B & 53.1{\color{gray}\scriptsize$\pm$0.2} & 57.9{\color{gray}\scriptsize$\pm$1.1} & 71.1{\color{gray}\scriptsize$\pm$1.3} & 25.2{\color{gray}\scriptsize$\pm$1.6} & 74.5{\color{gray}\scriptsize$\pm$0.9} & 61.5{\color{gray}\scriptsize$\pm$0.6} \\
 &  & 8B & 53.5{\color{gray}\scriptsize$\pm$0.2} & 60.2{\color{gray}\scriptsize$\pm$1.0} & 64.6{\color{gray}\scriptsize$\pm$1.4} & 45.4{\color{gray}\scriptsize$\pm$1.9} & 75.4{\color{gray}\scriptsize$\pm$0.9} & 64.2{\color{gray}\scriptsize$\pm$0.6} \\
 &  & 14B & 54.2{\color{gray}\scriptsize$\pm$0.2} & 55.5{\color{gray}\scriptsize$\pm$1.1} & 53.9{\color{gray}\scriptsize$\pm$1.5} & 23.6{\color{gray}\scriptsize$\pm$1.6} & 73.2{\color{gray}\scriptsize$\pm$0.9} & 57.0{\color{gray}\scriptsize$\pm$0.6} \\
\cdashline{2-9}[0.5pt/2pt]
 & \multirow{2}{*}{LLAVA-OV} & 0.5B & 30.1{\color{gray}\scriptsize$\pm$0.2} & 51.0{\color{gray}\scriptsize$\pm$1.1} & 51.1{\color{gray}\scriptsize$\pm$1.5} & 65.7{\color{gray}\scriptsize$\pm$1.8} & 35.7{\color{gray}\scriptsize$\pm$1.0} & 47.7{\color{gray}\scriptsize$\pm$0.6} \\
 &  & 7B & 44.5{\color{gray}\scriptsize$\pm$0.2} & 57.7{\color{gray}\scriptsize$\pm$1.1} & 62.0{\color{gray}\scriptsize$\pm$1.4} & 69.4{\color{gray}\scriptsize$\pm$1.7} & 72.7{\color{gray}\scriptsize$\pm$0.9} & 65.2{\color{gray}\scriptsize$\pm$0.6} \\
\cdashline{2-9}[0.5pt/2pt]
 & \multirow{3}{*}{Qwen3-VL} & 2B & 51.0{\color{gray}\scriptsize$\pm$0.2} & 60.4{\color{gray}\scriptsize$\pm$1.0} & 68.7{\color{gray}\scriptsize$\pm$1.3} & 93.1{\color{gray}\scriptsize$\pm$0.9} & 74.3{\color{gray}\scriptsize$\pm$0.9} & 71.1{\color{gray}\scriptsize$\pm$0.6} \\
 &  & 4B & 55.1{\color{gray}\scriptsize$\pm$0.2} & 61.1{\color{gray}\scriptsize$\pm$1.0} & 70.4{\color{gray}\scriptsize$\pm$1.3} & 88.5{\color{gray}\scriptsize$\pm$1.2} & \underline{76.1}{\color{gray}\scriptsize$\pm$0.9} & 71.6{\color{gray}\scriptsize$\pm$0.6} \\
 &  & 8B & 54.5{\color{gray}\scriptsize$\pm$0.2} & 55.4{\color{gray}\scriptsize$\pm$1.1} & 56.5{\color{gray}\scriptsize$\pm$1.5} & 47.3{\color{gray}\scriptsize$\pm$1.9} & 69.0{\color{gray}\scriptsize$\pm$1.0} & 59.2{\color{gray}\scriptsize$\pm$0.6} \\
\cdashline{2-9}[0.5pt/2pt]
 & \textbf{Ours} & 8B & \textbf{64.7}{\color{gray}\scriptsize$\pm$0.2} & \textbf{72.5}{\color{gray}\scriptsize$\pm$0.9} & \textbf{75.6}{\color{gray}\scriptsize$\pm$1.2} & \underline{96.6}{\color{gray}\scriptsize$\pm$0.5} & \textbf{81.2}{\color{gray}\scriptsize$\pm$0.8} & \textbf{79.3}{\color{gray}\scriptsize$\pm$0.5} \\
\midrule
\multirow{3}{*}{Mid-Fusion} & \multirow{2}{*}{Qwen3-VL-Emb} & 2B & 43.9{\color{gray}\scriptsize$\pm$0.2} & 61.4{\color{gray}\scriptsize$\pm$1.0} & \textbf{72.0}{\color{gray}\scriptsize$\pm$1.3} & 96.6{\color{gray}\scriptsize$\pm$0.5} & 69.9{\color{gray}\scriptsize$\pm$1.0} & 71.0{\color{gray}\scriptsize$\pm$0.6} \\
 &  & 8B & \underline{44.0}{\color{gray}\scriptsize$\pm$0.2} & \underline{61.5}{\color{gray}\scriptsize$\pm$1.0} & \underline{71.3}{\color{gray}\scriptsize$\pm$1.3} & \underline{96.7}{\color{gray}\scriptsize$\pm$0.5} & \underline{72.7}{\color{gray}\scriptsize$\pm$0.9} & \underline{71.9}{\color{gray}\scriptsize$\pm$0.6} \\
\cdashline{2-9}[0.5pt/2pt]
 & \textbf{Ours} & 8B & \textbf{63.8}{\color{gray}\scriptsize$\pm$0.2} & \textbf{71.2}{\color{gray}\scriptsize$\pm$0.9} & 64.2{\color{gray}\scriptsize$\pm$1.4} & \textbf{96.8}{\color{gray}\scriptsize$\pm$0.5} & \textbf{78.1}{\color{gray}\scriptsize$\pm$0.8} & \textbf{75.9}{\color{gray}\scriptsize$\pm$0.5} \\
\midrule
\multirow{7}{*}{Late Fusion} & Gemini-2-Embedding & -- & 54.1{\color{gray}\scriptsize$\pm$0.2} & 61.5{\color{gray}\scriptsize$\pm$1.0} & 61.5{\color{gray}\scriptsize$\pm$1.4} & 95.8{\color{gray}\scriptsize$\pm$0.6} & 75.9{\color{gray}\scriptsize$\pm$0.9} & 71.2{\color{gray}\scriptsize$\pm$0.6} \\
 & VLM2Vec v2.0 & 2B & 53.0{\color{gray}\scriptsize$\pm$0.2} & 62.3{\color{gray}\scriptsize$\pm$1.0} & 66.6{\color{gray}\scriptsize$\pm$1.4} & 90.9{\color{gray}\scriptsize$\pm$1.0} & 73.5{\color{gray}\scriptsize$\pm$0.9} & 70.8{\color{gray}\scriptsize$\pm$0.6} \\
 & SteerViT & 86M & 51.4{\color{gray}\scriptsize$\pm$0.2} & \underline{67.9}{\color{gray}\scriptsize$\pm$1.0} & 66.0{\color{gray}\scriptsize$\pm$1.4} & \underline{96.6}{\color{gray}\scriptsize$\pm$0.5} & 75.2{\color{gray}\scriptsize$\pm$0.9} & \underline{74.0}{\color{gray}\scriptsize$\pm$0.5} \\
 & GeneCIS & 150M & 44.9{\color{gray}\scriptsize$\pm$0.2} & 62.4{\color{gray}\scriptsize$\pm$1.0} & \underline{71.8}{\color{gray}\scriptsize$\pm$1.3} & 96.3{\color{gray}\scriptsize$\pm$0.6} & 72.7{\color{gray}\scriptsize$\pm$0.9} & 72.2{\color{gray}\scriptsize$\pm$0.6} \\
\cdashline{2-9}[0.5pt/2pt]
 & \multirow{2}{*}{Qwen3-VL-Emb} & 2B & 56.7{\color{gray}\scriptsize$\pm$0.2} & 64.4{\color{gray}\scriptsize$\pm$1.0} & 70.0{\color{gray}\scriptsize$\pm$1.3} & 92.7{\color{gray}\scriptsize$\pm$0.9} & 77.9{\color{gray}\scriptsize$\pm$0.8} & 73.9{\color{gray}\scriptsize$\pm$0.5} \\
 &  & 8B & \underline{58.3}{\color{gray}\scriptsize$\pm$0.2} & 62.9{\color{gray}\scriptsize$\pm$1.0} & 68.6{\color{gray}\scriptsize$\pm$1.3} & 94.5{\color{gray}\scriptsize$\pm$0.7} & \underline{78.1}{\color{gray}\scriptsize$\pm$0.8} & 73.5{\color{gray}\scriptsize$\pm$0.5} \\
\cdashline{2-9}[0.5pt/2pt]
 & \textbf{Ours} & 8B & \textbf{64.1}{\color{gray}\scriptsize$\pm$0.2} & \textbf{71.2}{\color{gray}\scriptsize$\pm$0.9} & \textbf{72.1}{\color{gray}\scriptsize$\pm$1.3} & \textbf{96.6}{\color{gray}\scriptsize$\pm$0.5} & \textbf{80.0}{\color{gray}\scriptsize$\pm$0.8} & \textbf{77.9}{\color{gray}\scriptsize$\pm$0.5} \\
\bottomrule
\end{tabular}

}}
\end{table}

\subsection{Additional analysis}
\label{sec:supp_reading_table}

Table~\ref{tab:big_results} includes analysis from more models, measured by their agreement with humans.

\myparagraph{Triplet oracle.} Section~\ref{sec:experiments} reports evaluation on pairwise metrics. Our goal is to create an algorithm that can judge pairwise distance, using the odd-one-out triplet task as a pretext training signal. Here, though the odd-one-out task is not the primary task, as an structurally advantaged oracle, we provide all 3 images to the VLM, and see how closely our pairwise algorithm (which only sees 2 images at a time) compares.
To do so,
we ask the model: ``Which image is the most different among the three in terms of {\menlo <aspect>}?'', same as what the human annotators answered. This makes even untrained VLMs strong baselines under the triplet prompt. While Gemini-3.1-Pro reaches $62.0\%$ on odd-one-out and GPT-5.4 reaches $77.9\%$ total on 2AFC, our early fusion method still outperforms even without the structural advantage. 

In addition, we verify that finetuning a generative VLM to solve the odd-one-out task directly with the triplet yields the strongest result (see Ours-Triplet). However, such a model produces only a triplet decision and not a pairwise score, and therefore does not transfer to 2AFC evaluation and retrieval. The pairwise design is what makes the metric usable across all of these settings.

\myparagraph{Effect of model size.} As expected, for most model families, agreement with humans increases with backbone size on both the in-distribution odd-one-out test and the 2AFC total. For example, in early fusion baselines, InternVL3.5 climbs $45.4\% \to 54.2\%$ from 1B to 14B, and Qwen3-VL climbs $51.0\% \to 54.5\%$ from 2B to 8B. In our ablation studies (Section~\ref{sec:supp_ablations}), we also find that our method's performance improves with model size (Table~\ref{tab:supp_ablations}).

\begin{figure}
    \centering
    \includegraphics[width=1.0\linewidth]{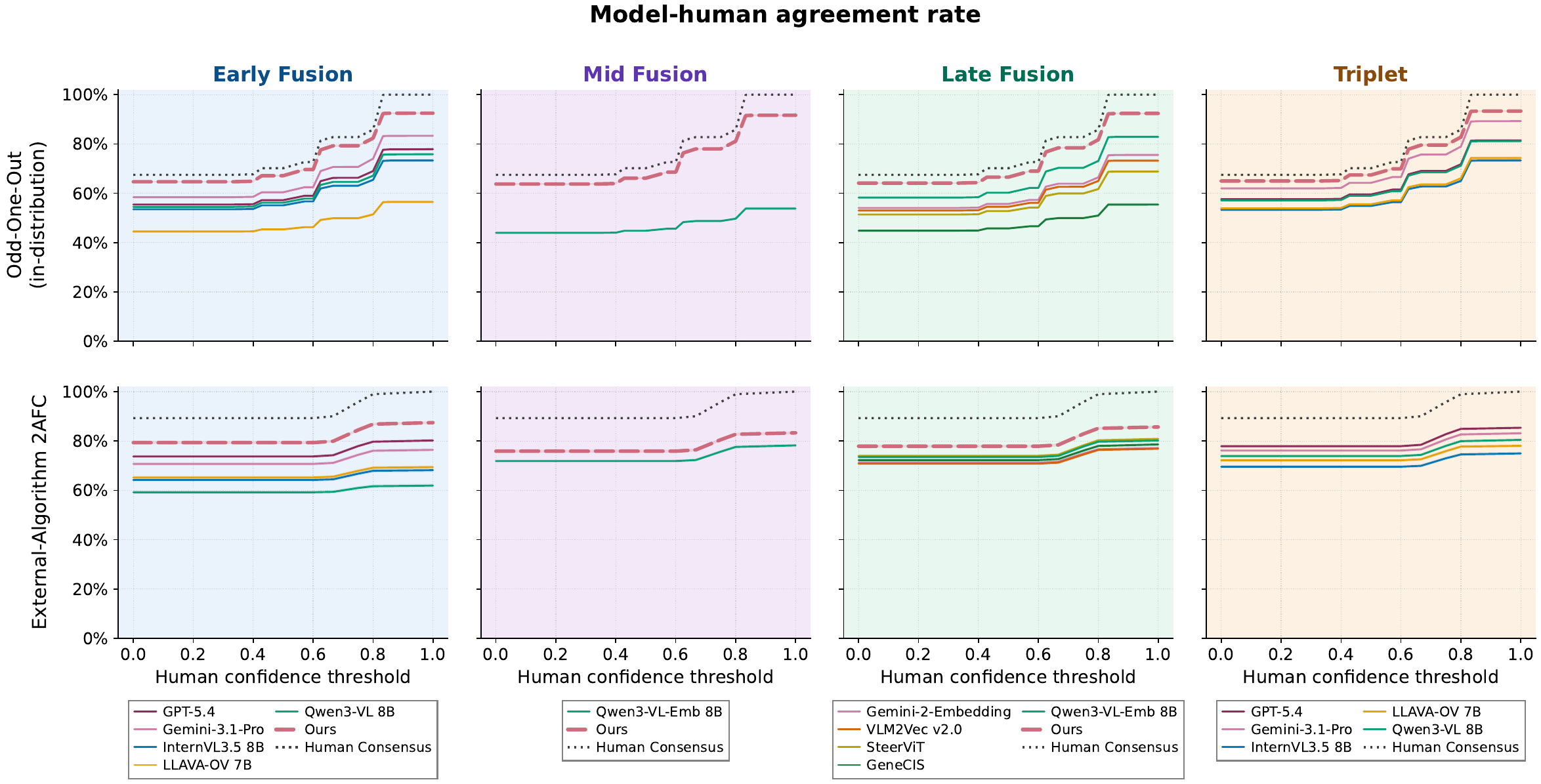}
    \caption{\textbf{Model's agreement with humans as a function of confidence threshold.} We filter by human confidence threshold (x-axis) and report model performance (y-axis) on the retained data.}
    \label{fig:supp_threshold_curves}
\end{figure}

\myparagraph{Human confidence vs.\ model's agreement with humans.} As the perceptual task is subjective, often humans do not agree $100\%$ on a given example. In Figure~\ref{fig:supp_threshold_curves}, we plot model performance against $\tau$, how often humans agree with their top selected choice. Recall that 5 annotators are used per aspect, with some removed if they select ``can't tell''. Hence, $\tau$ increases at specific values (e.g., $\tfrac{3}{5}, \tfrac{4}{5}, \tfrac{3}{4}$, etc).
As expected, at high confidence levels $\tau$, performance rises monotonically for all methods, as low-confidence predictions are filtered out. The method rankings remain consistent across the entire threshold range. Our model maintains the smallest gap to human consensus at every $\tau$, demonstrating that its advantage is not threshold-sensitive.

\begin{figure}[t]

    \centering
    \includegraphics[width=0.95\linewidth]{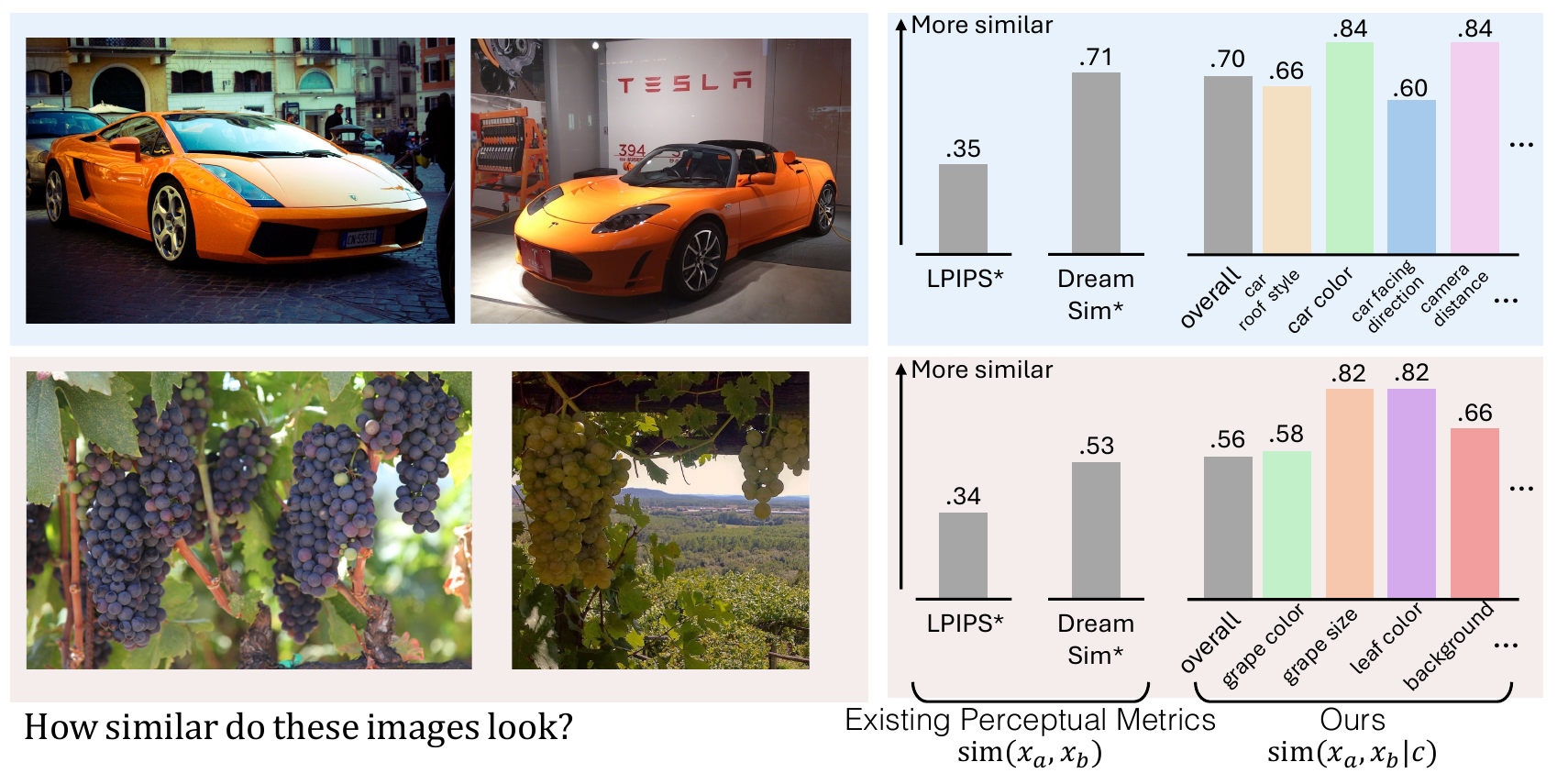}
    \vspace{-5pt}
    \caption{\textbf{Examples of text-conditioned similarity score}. Given two images, existing perceptual metrics provide only a crude, \textit{overall} similarity measure. Our method reveals a key aspect of the problem by conditioning on the \textit{sense} of similarity, denoted by the text-prompt $\mathbf{c}$. (top) For example, the cars share a more similar color and are taken at a similar camera distance, whereas their facing direction and roof style are quite different. (bottom) The grape colors are different, while the grape size and the leaf color are the same. $^{*}$We show 1-LPIPS~\cite{zhang2018lpips} and 1-DreamSim~\cite{fu2023dreamsim} to convert the distances so that higher scores represent higher similarities.}
    \label{fig:supp_tpips}
    \vspace{-10pt}
\end{figure}

\begin{figure}[t]
    \centering
    \includegraphics[width=1.0\linewidth]{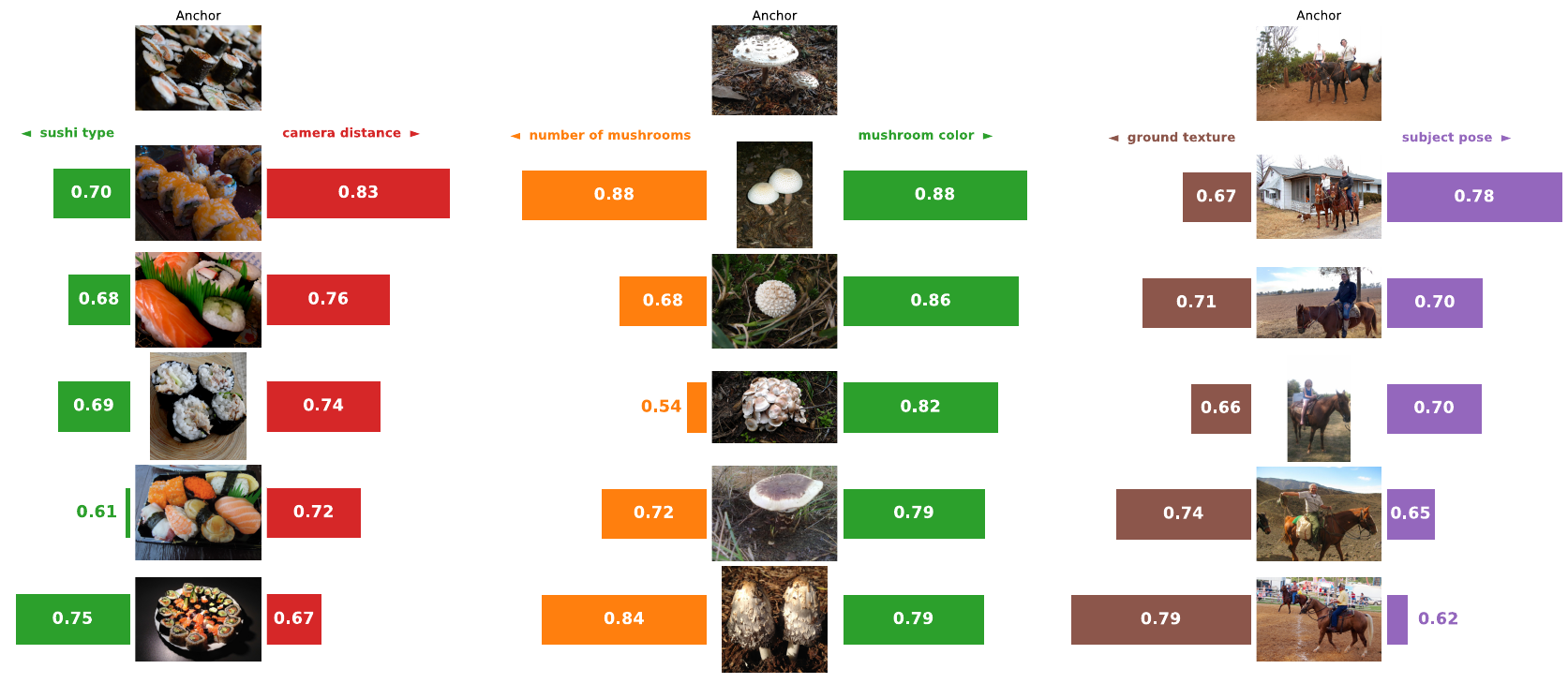}
    \caption{\textbf{Exploring senses of similarity with our metric}. Left (Sushi): The anchor is a close-up shot of sushi rolls. Prompts can prioritize similarity of shot composition (close-up vs. distant) or the sushi type, distinguishing similar rolls from dissimilar sashimi. Middle (Mushrooms): The anchor shows two white mushrooms. A color-based query sorts items from white to brown, while a count-based query reorders them according to the number of mushrooms present. Right (Horses): Similarity focuses either on the physical pose of the subjects or the specific texture and color of the ground. Similarity is reported as cosine similarity, ranging from -1 to 1.}
    \label{fig:teaser2}
\end{figure}

\begin{figure}
    \centering
    \includegraphics[width=1.0\linewidth]{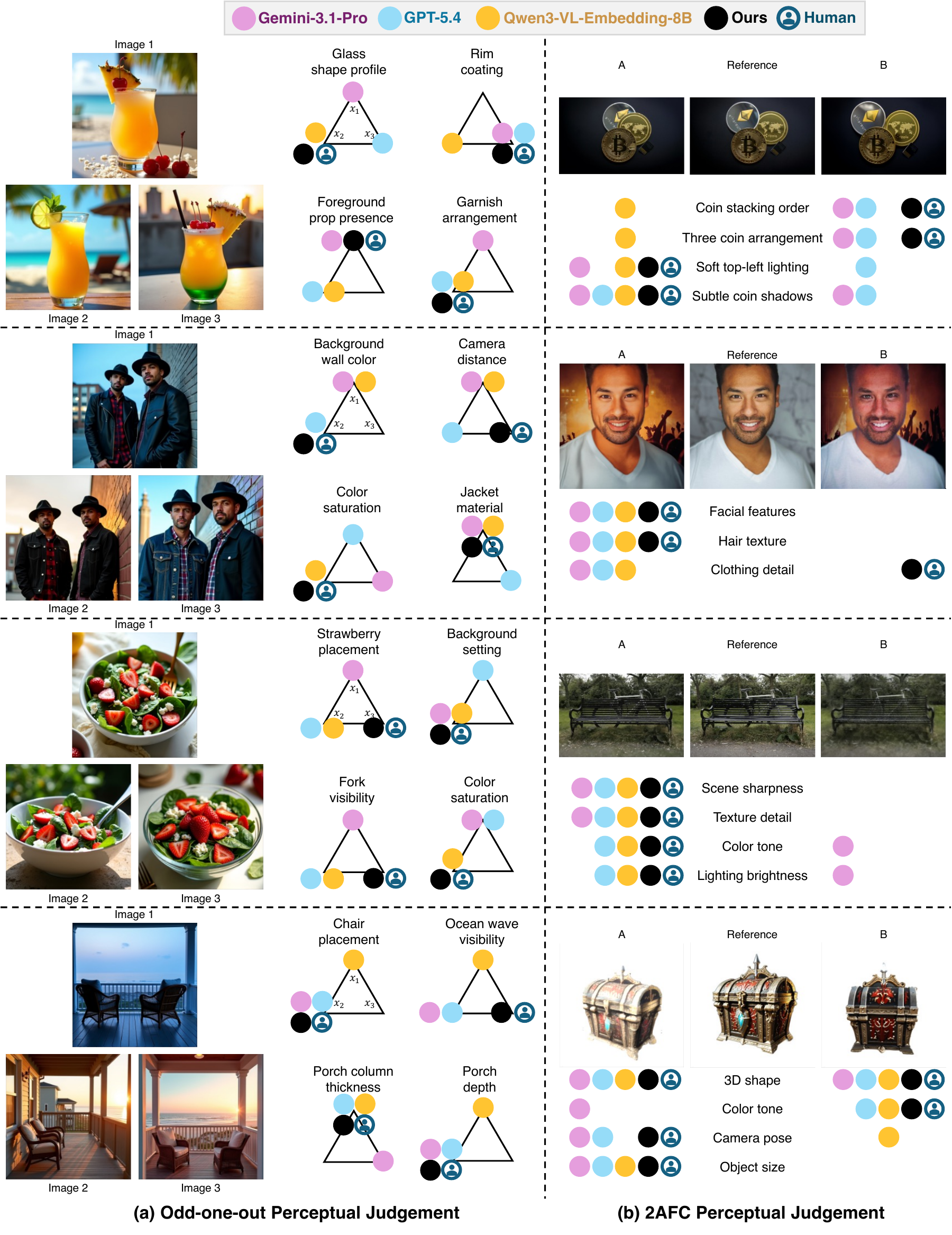}
    \caption{\textbf{Additional qualitative results.} We extend Figure~\ref{fig:qual_model_human_agreement.pdf} and provide more results for our odd-one-out and 2AFC test cases.}
    \label{fig:supp_qual}
\end{figure}

\begin{figure}
    \centering
    \includegraphics[width=1.0\linewidth]{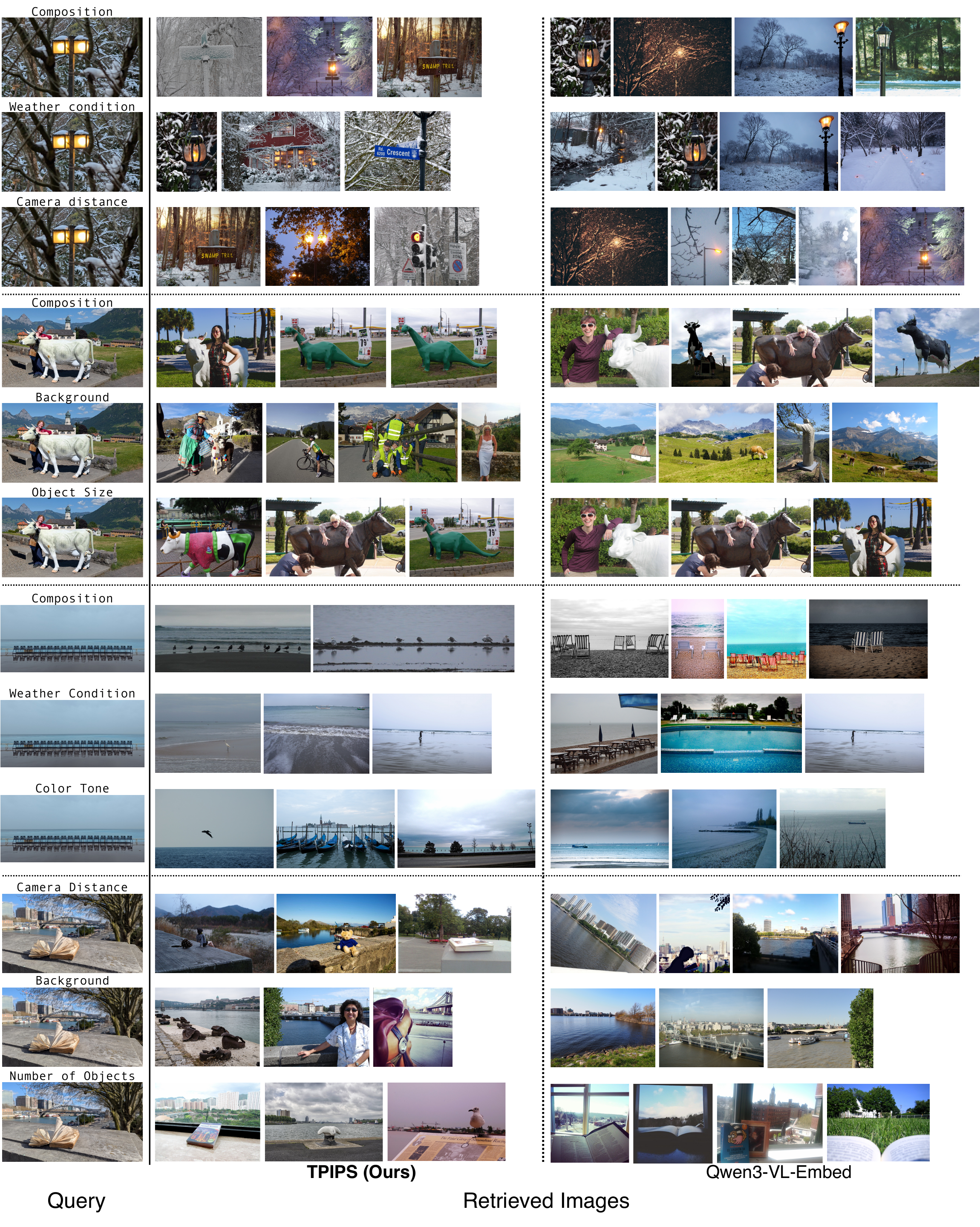}
    \caption{\textbf{Aspect-conditioned retrieval results without similarity adjustments.} We extend results from Figure~\ref{fig:retrieval_multi_factor} and show retrieval results without similarity adjustments. For each query (left), conditioning on different aspects also returns different nearest neighbors. Left shows retrievals on OpenImages~\cite{kuznetsova2020openimages} using our trained late-fusion model, where it visibly produces better-aligned retrieval images than the base model's retrieval on the right.}
    \label{fig:supp_retrieval_multi_factor_openimages}
\end{figure}

\begin{figure}
    \centering
    \includegraphics[width=1.0\linewidth]{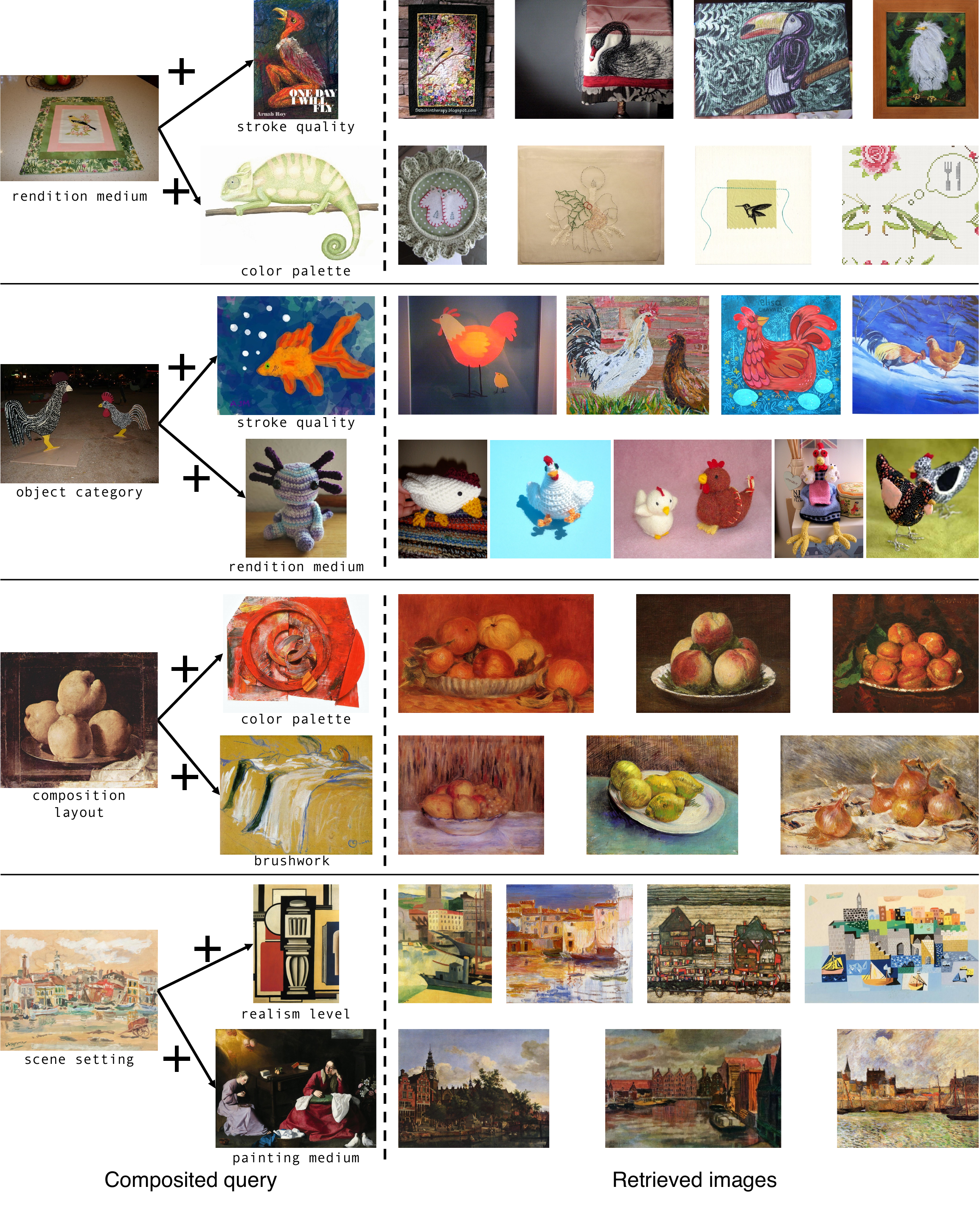}
    \caption{\textbf{Additional compositional retrieval results.} We extend results from Figure~\ref{fig:retrieval_multi_query}. Top two rows are compositional retrieval results from ImageNet-R~\cite{hendrycks2021many}, and the bottom two are from WikiArt~\cite{artgan2018}.}
    \label{fig:supp_retrieval_multi_query}
\end{figure}

\myparagraph{More qualitative results.}
To extend Section~\ref{sec:experiments}, we present additional qualitative results below.
Figure~\ref{fig:supp_tpips} and Figure~\ref{fig:teaser2} illustrate the range of aspect-conditioned similarities.
Figure~\ref{fig:supp_qual} expands the in-distribution odd-one-out and \arxiv{external-algorithm} 2AFC results.
Figure~\ref{fig:supp_retrieval_multi_factor_openimages} extends Figure~\ref{fig:retrieval_multi_factor} and provides nearest-neighbor retrieval on OpenImages~\cite{kuznetsova2020openimages} without the similarity adjustment. Conditioning the same query on different aspects returns visibly different neighbor sets, and our method still visibly outperforms the base model qualitatively in this setup. Figure~\ref{fig:supp_retrieval_multi_query} extends Figure~\ref{fig:retrieval_multi_query} and provides more results on multi-query retrieval.

\subsection{Additional ablations}
\label{sec:supp_ablations}

Table~\ref{tab:supp_ablations} reports the ablation results.

\begin{table}[t]
\caption{\textbf{Additional ablations.} Each row flips one knob of our default models. The default rows (8B and 100\%) are the default late fusion model. ``Mid-fusion default'' is the default mid-fusion model.}
\label{tab:supp_ablations}
\centering
\footnotesize
\setlength{\tabcolsep}{6pt}
\renewcommand{\arraystretch}{1.15}
\begin{tabular}{lllcc@{}}
\toprule
\textbf{Ablation} & \textbf{Family} & \textbf{Variant} & \textbf{OOO test (\%)} & \textbf{2AFC total (\%)} \\
\midrule
\multirow{2}{*}{\emph{Backbone scale}} &
\multirow{2}{*}{Late fusion} & 2B               & 63.7{\color{gray}\scriptsize$\pm$0.2} & 77.4{\color{gray}\scriptsize$\pm$0.5} \\
& & 8B (default)     & 64.1{\color{gray}\scriptsize$\pm$0.2} & 77.9{\color{gray}\scriptsize$\pm$0.5} \\
\midrule
\multirow{5}{*}{\emph{Training data fraction}} &
\multirow{5}{*}{Late fusion} & 20\%             & 62.8{\color{gray}\scriptsize$\pm$0.2} & 76.6{\color{gray}\scriptsize$\pm$0.5} \\
& & 40\%             & 63.7{\color{gray}\scriptsize$\pm$0.2} & 78.0{\color{gray}\scriptsize$\pm$0.5} \\
& & 60\%             & 64.0{\color{gray}\scriptsize$\pm$0.2} & 77.4{\color{gray}\scriptsize$\pm$0.5} \\
& & 80\%             & 64.0{\color{gray}\scriptsize$\pm$0.2} & 77.5{\color{gray}\scriptsize$\pm$0.5} \\
& & 100\% (default)  & 64.1{\color{gray}\scriptsize$\pm$0.2} & 77.9{\color{gray}\scriptsize$\pm$0.5} \\
\midrule
\multirow{3}{*}{\emph{Mid-fusion head}} &
\multirow{3}{*}{Mid-fusion} & frozen backbone        & 45.6{\color{gray}\scriptsize$\pm$0.2} & 72.9{\color{gray}\scriptsize$\pm$0.5} \\
& & no aspect conditioning for channel weights & 59.8{\color{gray}\scriptsize$\pm$0.2} & 75.1{\color{gray}\scriptsize$\pm$0.5} \\
& & default & 63.8{\color{gray}\scriptsize$\pm$0.2} & 75.9{\color{gray}\scriptsize$\pm$0.5} \\
\bottomrule
\end{tabular}
\end{table}

\myparagraph{Backbone scale.} The 2B late-fusion model trails the 8B version by only a marginal gap on both the odd-one-out ($63.7\%$ vs.\ $64.1\%$) and 2AFC ($77.4\%$ vs.\ $77.9\%$) tasks, offering a more efficient alternative with competitive performance.

\myparagraph{Data scaling.} Performance saturates early. The odd-one-out metric plateaus at $60\%$ of training data ($64.0\%$), and 2AFC reaches $78.0\%$ at just $40\%$, on par with the full-data result. 
\arxiv{We note that our method's improvement flattens as model-human agreement approaches the observed human consensus level.}

\myparagraph{Mid-fusion architecture.} We ablate mid-fusion architecture choice described in Appendix~\ref{sec:supp_fusion_strategy}. Freezing the backbone collapses odd-one-out and 2AFC performance substantially ($45.6\%$ / $72.9\%$ vs.\ $63.8\%$ / $75.9\%$ for the default), confirming that learning channel weights alone on a frozen VLM is insufficient. Removing aspect conditioning from the channel weights also degrades both metrics ($59.8\%$ / $75.1\%$), indicating that the optimal per-layer weighting is aspect-dependent and cannot be shared across conditions.

\section{Implementation details}
\label{sec:supp_implementation}

In this section, we include implementation details for each fusion strategy (Section~\ref{sec:supp_fusion_strategy}), architecture details for our early fusion model (Section~\ref{sec:supp_early_fusion}), training and hyperparameter details (Section~\ref{sec:supp_training}), and baseline details (Section~\ref{sec:supp_baselines}).
\subsection{\arxiv{Fusion strategy implementation details}}
\label{sec:supp_fusion_strategy}
\arxiv{We extend Section~\ref{sec:method} and discuss the implementation of each fusion strategy in more detail.}

\myparagraph{Late fusion (embeddings).}
We repurpose the VLM as a text-guided embedding model, following recent work on instruction-tuned VLM retrievers~\cite{jiang2025vlm2vec,lee2024nvembed}. Given an image $\mathbf{x}$ and an aspect $\mathbf{c}$, we form the prompt ``\emph{Represent the similarity of the image based on \texttt{<}$\mathbf{c}$\texttt{>}}'' and feed it together with $\mathbf{x}$ into the VLM, taking the final, normalized hidden state of the last token as the embedding $\mathbf{z} = \widehat{g}_\theta(\mathbf{x}, \mathbf{c})$. Pairwise similarity is then the cosine of the two embeddings:
$f_\theta(\mathbf{x}_1, \mathbf{x}_2 \mid \mathbf{c}) = \widehat{g}_\theta(\mathbf{x}_1, \mathbf{c})^{\!\top}\, \widehat{g}_\theta(\mathbf{x}_2, \mathbf{c})$.

This is efficient at inference time, as embeddings can be precomputed and cached. As the two images never interact within the network, the model must commit to a text-guided summary independently.

\myparagraph{Mid fusion (activation distances).}
Following LPIPS~\cite{zhang2018lpips}, we can instead compare internal activations of the VLM from the LLM decoder layers. We feed the image-prompt pair into the VLM, and for each of $L$ selected transformer layers, extract image tokens with token length $T$, unit-normalize them along the channel dimension $d$ to obtain a per-layer feature $\boldsymbol{\phi}_\ell(\mathbf{x}, \mathbf{c}) \in \mathbb{R}^{T\times d}$. The similarity is defined as the negative weighted $\ell_2$ distance between the per-layer features of the two images, averaged across image tokens, and summed over layers:
\begin{equation}
\label{eq:mid_fusion}
f_\theta(\mathbf{x}_1, \mathbf{x}_2 \mid \mathbf{c})
\;=\;
- \sum_{\ell=1}^{L} \frac{1}{T} \sum_{t=1}^{T} \bigl\lVert \mathbf{w}_{\ell}(\mathbf{c}) \odot \bigl(\boldsymbol{\phi}_{\ell}(\mathbf{x}_1, \mathbf{c})_t - \boldsymbol{\phi}_{\ell}(\mathbf{x}_2, \mathbf{c})_t\bigr)\bigr\rVert_2^2.
\end{equation}
Unlike LPIPS, which tunes a fixed set of channel weights on top of a frozen backbone, our channel weights $\mathbf{w}_\ell(\mathbf{c}) \in \mathbb{R}^{d}$ are text-conditioned per-channel weights per layer, parameterized by an MLP that takes in the averaged text token embedding in the same layer. We additionally fine-tune the VLM itself with LoRA~\cite{hu2022lora} adapters. Like late fusion, this architecture treats each image independently and fuses only at the comparison stage. Appendix~\ref{sec:supp_ablations} analyzes these design changes in more detail.

\myparagraph{Early fusion}
lets the two images interact throughout the network. Both images and prompt ``\emph{Measure the similarity between the two images based on \texttt{<}$\mathbf{c}$\texttt{>}}'' are packed into a single input sequence,
\begin{equation*}
[\, \mathbf{c} \text{ tokens} \;\|\; \mathbf{x}_1 \text{ patches} \;\|\; \mathbf{x}_2 \text{ patches} \;\|\; \mathbf{r}_1 \;\|\; \mathbf{r}_2 \,],
\end{equation*}
so that every patch of $\mathbf{x}_1$ can attend to every patch of $\mathbf{x}_2$ at every layer, enabling direct patch-level correspondence. The two learnable register tokens $\mathbf{r}_1, \mathbf{r}_2$ act as text-conditioned readouts for the two images. Their final hidden states $\mathbf{h}_1, \mathbf{h}_2$ yield the similarity score: $ f_\theta(\mathbf{x}_1, \mathbf{x}_2 \mid \mathbf{c})
 \;=\;
 \frac{\mathbf{h}_1^{\!\top} \mathbf{h}_2}{\lVert \mathbf{h}_1 \rVert \, \lVert \mathbf{h}_2 \rVert}.$

Importantly, by using careful design of the attention mask and positional encoding (as described below in Appendix~\ref{sec:supp_early_fusion}),
we satisfy two desirable properties \emph{by construction}, without any auxiliary loss: \textbf{symmetry}, $f_\theta(\mathbf{x}_1, \mathbf{x}_2 \mid \mathbf{c}) = f_\theta(\mathbf{x}_2, \mathbf{x}_1 \mid \mathbf{c})$, and \textbf{identity}, $f_\theta(\mathbf{x}, \mathbf{x} \mid \mathbf{c}) = 1$ for any $\mathbf{x}$. %

\subsection{Early-fusion architecture}
\label{sec:supp_early_fusion}

We describe the attention mask and positional encoding scheme that make the early-fusion architecture symmetric and identity-preserving by construction.

\myparagraph{Input layout.} The input sequence consists of four contiguous segments, with the aspect prompt first so that image patches can attend to it in the standard VLM fashion:
\begin{equation*}
\underbrace{\mathbf{t}_1, \ldots, \mathbf{t}_M}_{\text{aspect tokens } \mathbf{c}}
\;\|\;
\underbrace{\mathbf{p}^{(1)}_1, \ldots, \mathbf{p}^{(1)}_N}_{\text{image } \mathbf{x}_1 \text{ patches}}
\;\|\;
\underbrace{\mathbf{p}^{(2)}_1, \ldots, \mathbf{p}^{(2)}_N}_{\text{image } \mathbf{x}_2 \text{ patches}}
\;\|\;
\underbrace{\mathbf{r}_1, \mathbf{r}_2}_{\text{registers}},
\end{equation*}
where $\mathbf{t}_m$ are the aspect text token embeddings, $\mathbf{p}^{(k)}_n$ denotes the $n$-th patch embedding of image $\mathbf{x}_k$ (we assume both images have the same number of patches $N$), and $\mathbf{r}_1, \mathbf{r}_2$ are shared-parameter learnable register vectors shared across all examples.

\myparagraph{Positional encoding.} 
We use M-RoPE~\cite{wang2024qwen2vl}, which assigns each token a three-dimensional position $(t,h,w)$.
For \emph{vision} tokens we override the backbone default so that \emph{both} images share the same temporal index and matching spatial indices at corresponding grid locations:
\begin{equation*}
\mathrm{pos}\bigl(\mathbf{p}^{(1)}_{h,w}\bigr) = \mathrm{pos}\bigl(\mathbf{p}^{(2)}_{h,w}\bigr) = (0,\, h,\, w),
\end{equation*}
where $(h,w)$ follow each image's patch grid after preprocessing.
Hence, the two images are distinguished only by token content and by the attention mask, \textit{not} by the image ordering.

The aspect text is assigned sequential temporal positions $t=1,2,\ldots$ in sequence order, with $(h,w)=(0,0)$, while reserving $t=0$ for all vision patches---this differs from default two-image preprocessing in which the second image often receives a larger temporal index.

\myparagraph{Registers.}
The two registers use the \emph{same} learnable embedding vector and the \emph{same} M-RoPE coordinates $(t_{\mathrm{reg}},0,0)$ with $t_{\mathrm{reg}}$ set immediately after the maximum temporal index among all preceding tokens.
They are therefore not distinguished by positional encoding; only their attention regions differ, as illustrated below.

\myparagraph{Attention mask.} The attention mask is defined by the following rules:
\begin{enumerate}[leftmargin=1.5em, itemsep=1pt, topsep=2pt]
    \item Aspect tokens are causal among themselves and hence attend to no image patches or registers, preserving the pretrained text-only behavior of the VLM.
    \item Image patches attend to all aspect tokens and to all image patches of \emph{both} images, encouraging comparisons between two sets of image features. They do not attend to the registers.
    \item Registers are restricted: $\mathbf{r}_k$ attends to the aspect tokens and to the patches of image $\mathbf{x}_k$ only. This encourages $\mathbf{r}_k$ to output an image-specific representation.
\end{enumerate}
The image-image block of the mask is symmetric, and the register block is symmetric in the analogous sense: $\mathbf{r}_1$'s access pattern to $\mathbf{x}_1$ mirrors $\mathbf{r}_2$'s access pattern to $\mathbf{x}_2$.

\myparagraph{Symmetry property.} The two registers $\mathbf{r}_1, \mathbf{r}_2$ share the same learned embedding and the same positional encoding, and the attention mask treats the two image slots symmetrically. So if we swap $\mathbf{x}_1$ and $\mathbf{x}_2$, at every attention layer $\mathbf{r}_1$ now pools from $\mathbf{x}_2$ exactly as $\mathbf{r}_2$ did before (and vice versa); the two register hidden states are simply exchanged. Since cosine similarity is symmetric in its arguments,
\begin{equation*}
f_\theta(\mathbf{x}_1, \mathbf{x}_2, \mathbf{c}) = f_\theta(\mathbf{x}_2, \mathbf{x}_1, \mathbf{c}).
\end{equation*}

\myparagraph{Identity property.} 
When $\mathbf{x}_1 = \mathbf{x}_2 = \mathbf{x}$, the two image segments of the input sequence are token-for-token identical and the attention mask is symmetric, so the two segments produce identical hidden states at every transformer layer. Since $\mathbf{r}_1$ and $\mathbf{r}_2$ are tied and attend to identical contexts via structurally identical masks, $\mathbf{h}_1 = \mathbf{h}_2$. Hence
\begin{equation*}
f_\theta(\mathbf{x}, \mathbf{x}, \mathbf{c}) = \frac{\mathbf{h}_1^{\!\top} \mathbf{h}_1}{\lVert \mathbf{h}_1 \rVert^2} = 1.
\end{equation*}

\subsection{Training and hyperparameter selection}
\label{sec:supp_training}

\myparagraph{Backbones.} The late-fusion, mid-fusion, and early-fusion variants of our model all use Qwen3-VL-Embedding-8B~\cite{li2026qwen3vlembedding} as the backbone. The triplet generative variant uses Qwen3-VL-8B-Instruct~\cite{bai2025qwen3vl}.

\myparagraph{Adaptation.} We apply LoRA~\cite{hu2022lora} to the language tower with $\text{lora\_alpha} = 2 \cdot \text{lora\_r}$ and dropout $0.05$. The vision tower is frozen for all reported models, as we find that tuning the vision tower doesn't improve performance.

\myparagraph{Optimizer and schedule.} We optimize using AdamW (weight decay $0.01$, gradient clipping $1.0$). The learning-rate schedule is a linear warmup over $3\%$ of total steps followed by cosine decay to zero~\cite{loshchilov2017sgdr}. The triplet variant uses two parameter groups: backbone LoRA at the listed learning rate and the classification head at $5\times$ that rate.

\myparagraph{Data and per-step batching.} All four models are trained on the unfiltered training split (22{,}157 triplets) for one epoch. Each aspect in a triplet is treated as one training example, so a triplet annotated for $k$ aspects contributes $k$ optimization examples per epoch. Per-device batch size is $1$; the effective batch size is per-device batch size times gradient accumulation times eight GPUs, since our training is done on 8-GPU nodes. %

\myparagraph{Permutation augmentation for triplet models.} To alleviate position bias of vision language models~\cite{zhao2021calibrate,zheng2024large,wang2024large}, when training triplet models, the three images of every triplet are randomly permuted per batch item. At inference, we average over all six permutations and map probabilities back to the original positions.

\myparagraph{Model prompts.} The aspect string $\mathbf{c}$ is inserted into the following prompts. For late fusion and mid fusion (one image per forward pass):

\begin{quote}\small\itshape
``Represent the similarity of the image based on $\langle\,\mathbf{c}\,\rangle$.''
\end{quote}

For early fusion (two images per forward pass):

\begin{quote}\small\itshape
``Measure the similarity between the two images based on $\langle\,\mathbf{c}\,\rangle$.''
\end{quote}

The ``overall'' aspect drops the ``based on $\langle\,\mathbf{c}\,\rangle$'' suffix.

\myparagraph{Hyperparameter sweep.} We run an independent Bayesian search per family with validation human agreement as the target. The search spaces and the selected configurations are reported in Table~\ref{tab:supp_sweep} and Table~\ref{tab:supp_optimal_hparams}, respectively.

\begin{table}[t]
\caption{\textbf{Per-family hyperparameter search spaces.} Most hyperparameters are sampled from discrete choices, and the learning rate is sampled from a continuous log-uniform distribution.}
\label{tab:supp_sweep}
\centering
\footnotesize
\setlength{\tabcolsep}{5pt}
\renewcommand{\arraystretch}{1.15}
\resizebox{1.\linewidth}{!}{
\begin{tabular}{@{}lcccc@{}}
\toprule
\textbf{Hyperparameter} & \textbf{Late fusion} & \textbf{Early fusion} & \textbf{Triplet head} & \textbf{Mid fusion} \\
\midrule
Backbone                          & Qwen3-VL-Emb-8B & Qwen3-VL-Emb-8B & Qwen3-VL-8B-Instruct & Qwen3-VL-Emb-8B \\
LoRA rank $r$                     & \{8, 16, 32, 64\} & \{8, 16, 32, 64\} & \{8, 16, 32, 64\} & \{8, 16, 32, 64\} \\
Learning rate (log-uniform)       & $[10^{-5}, 5\!\times\!10^{-4}]$ & $[10^{-5}, 5\!\times\!10^{-4}]$ & $[10^{-5}, 5\!\times\!10^{-4}]$ & $[10^{-5}, 5\!\times\!10^{-4}]$ \\
Effective batch size              & \{32, 64, 128, 256\} & \{32, 64, 128, 256\} & \{32, 64, 128, 256\} & \{32, 64, 128, 256\} \\
Similarity temperature $\tau$     & \{0.05, 0.1, 0.2\} & \{0.05, 0.1, 0.2\} & --- & log-uniform $[10^{-3}, 1.0]$ \\
\# probe layers                   & --- & --- & --- & \{3, 5, 8, 10, 14, 20\} \\
\bottomrule
\end{tabular}}
\end{table}

\begin{table}[t]
\caption{\textbf{Final hyperparameters} used for every reported number.}
\label{tab:supp_optimal_hparams}
\centering
\footnotesize
\setlength{\tabcolsep}{5pt}
\renewcommand{\arraystretch}{1.15}
\begin{tabular}{@{}lcccc@{}}
\toprule
\textbf{Hyperparameter} & \textbf{Late fusion} & \textbf{Early fusion} & \textbf{Triplet head} & \textbf{Mid fusion} \\
\midrule
LoRA rank $r$                     & 16 & 16 & 16 & 16 \\
Learning rate                     & $5\!\times\!10^{-5}$ & $1\!\times\!10^{-4}$ & $3\!\times\!10^{-5}$ & $1\!\times\!10^{-4}$ \\
Effective batch size              & 128 & 128 & 32 & 32 \\
Similarity temperature $\tau$     & 0.05 & 0.05 & --- & 0.0025 \\
\# probe layers                   & --- & --- & --- & 20 \\
Epochs                            & 1 & 1 & 1 & 1 \\
\bottomrule
\end{tabular}
\end{table}

\subsection{Baselines}
\label{sec:supp_baselines}

We provide more details of the baselines reported in Section~\ref{sec:experiments}. We organize baselines by how the two images interact, plus a separate triplet family that consumes three concatenated images at once. All baselines use greedy decoding (temperature $0$). To avoid position bias of VLMs~\cite{zhao2021calibrate,zheng2024large,wang2024large}, for triplet and pairwise-score baselines we apply permutation averaging---6 orderings for odd-one-out, 2 for 2AFC---and map results back to the original positions. Embedding baselines do not need permutation averaging because cosine similarity is symmetric.

\myparagraph{Triplet baselines.} The three input images are concatenated horizontally and the model is asked which side is the odd one out. We use the prompt:

\begin{quote}\small\itshape
``I provide three horizontally concatenated images (left, middle, right). Which image is the most different among the three in terms of  $\langle\,\mathbf{c}\,\rangle$? You need to choose: left, middle, right. Answer:''
\end{quote}

For 2AFC the three positions become reference (center) and two candidates (left, right):

\begin{quote}\small\itshape
``I provide three horizontally concatenated images. The center image is the reference. Which candidate (left or right) is more similar to the reference in terms of $\langle\,\mathbf{c}\,\rangle$? Answer with one word: left or right. Answer:''
\end{quote}

When the aspect condition is ``overall'', we remove ``in terms of $\langle\,\mathbf{c}\,\rangle$'' in these prompts.

For open-weight models, we read logits at the answer position for the three (resp.\ two) target tokens and softmax. Models include InternVL3.5~\cite{wang2025internvl35}, Qwen3-VL~\cite{bai2025qwen3vl}, and LLaVA-OneVision~\cite{li2024llavaonevision}. API models (GPT-5.4~\cite{openai2023gpt4v}, Gemini~3.1 Pro~\cite{geminiteam2023gemini}) use text parsing.

\myparagraph{Pairwise 0--10 score baselines.} The same VLMs are also evaluated in a pairwise mode, where each pair of images is rated for similarity on an integer scale of $0$ (completely different) to $10$ (identical). The prompt is:

\begin{quote}\small\itshape
``I show you two images placed side by side (left and right). Rate how similar these two images are in terms of $\langle\,\mathbf{c}\,\rangle$ on a scale of 0 to 10, where 0 means completely different and 10 means identical. Answer with a single number from 0 to 10. Answer:''
\end{quote}

When the aspect is ``overall'', we swap out ``in terms of $\langle\,\mathbf{c}\,\rangle$'' to ``overall'' in the prompt.
We convert the response to a soft score either as $\mathbb{E}[d] = \sum_{d=0}^{10} d \cdot P(d)$ over digit-token logprobs. For API calls where we can only obtain the generated text, each text response is converted to a one-hot delta distribution over choices. The three pairwise scores per triplet are converted to a 3-way odd-one-out distribution by softmaxing the complementary-pair score (so the highest similarity between two images implies the third is the odd one out); for 2AFC we softmax the two reference--candidate scores.

\myparagraph{Late-fusion (embedding) baselines.} Each image is embedded independently, optionally conditioned on a short aspect prompt; pairwise similarity is the cosine of the two embeddings. The 3-way odd-one-out and 2AFC distributions are formed by the same softmax aggregation as the pairwise-score baselines.

\textit{Qwen3-VL-Embedding}~\cite{li2026qwen3vlembedding} (2B and 8B). The natively multimodal encoder version of Qwen3-VL with last-token pooling and L2-normalized output. We use the prompt ``\textit{Represent the visual similarity of this image in terms of $\langle\,\mathbf{c}\,\rangle$.}''.

\textit{VLM2Vec~v2.0}~\cite{jiang2025vlm2vec, meng2025vlm2vecv2}. A LoRA adapter on top of Qwen2-VL-2B-Instruct~\cite{wang2024qwen2vl}, used as released. We use the same prompt as Qwen3-VL-Embedding and pool the last non-pad hidden state.

\textit{SteerViT}~\cite{ruthardt2026steervit}. A DINOv2-based ViT~\cite{oquab2024dinov2} augmented with steerable cross-attention that consumes a short aspect string. SteerViT's text encoder is RoBERTa~\cite{liu2019roberta} rather than an instruction-tuned LLM, so the verbose embedding prompt above is shortened to the aspect token only (e.g., ``\textit{lighting}''). For the ``overall'' aspect we pass no text, which falls back to the unconditional DINOv2 backbone.

\textit{GeneCIS}~\cite{vaze2023genecis}. A CLIP ViT-B/16~\cite{radford2021clip} fine-tuned on conditional image-similarity triplets mined from existing vision-language datasets. As with SteerViT, CLIP's text encoder is short-text-friendly rather than instruction-tuned, so we feed the aspect token directly (e.g., ``\textit{lighting}'').

\textit{Gemini Embedding 2}~\cite{geminiteam2023gemini}. The natively multimodal embedding model, used via the official API. We interleave the aspect text and the image bytes inside a single content object so the embedding is aspect-conditioned.

\myparagraph{Mid-fusion baseline.} For completeness we report a frozen Qwen3-VL-Embedding (2B and 8B) used in the LPIPS-style mid-fusion configuration: pairwise similarity is the negative aspect-conditioned $\ell_2$ distance over selected language-tower layers (Equation~\ref{eq:mid_fusion}). This is the same architectural class as our trained mid-fusion variant but without the trained channel-weight head, and isolates the contribution of training.

\section{Additional details}
\label{sec:supp_additional_details}
\subsection{Compute cost}
\label{sec:supp_compute}

All training and inference reported in this paper were run on eight NVIDIA A100 80GB GPUs. We describe the runtime cost below.

\myparagraph{Dataset construction.} Curating the odd-one-out dataset took around 24 GPU days for FLUX image generation and around 8 GPU days for the Qwen3 text prompt variations. Curating the 2AFC dataset took around 20 GPU hours for image compositing, 5 GPU hours for NVS, and 30 GPU hours for image-to-3D. We directly downloaded image editing results from GEditBench v2~\cite{liu2025geditbench}.

\myparagraph{Final model training.} Per-run wall-clock on eight A100s: late fusion 4.5 hours, mid fusion 9 hours, early fusion 8 hours, triplet head 7 hours.

\subsection{Licenses}
\label{sec:supp_licenses}

Below we list the licenses of code, data, and models we used for this project.

\begin{itemize}[leftmargin=1.4em, itemsep=1pt, topsep=2pt]
    \item Qwen3-VL and Qwen3-VL-Embedding: Apache 2.0.
    \item InternVL3.5: Apache 2.0.
    \item LLaVA-OneVision: Apache 2.0.
    \item VLM2Vec v2.0: Apache 2.0.
    \item SteerViT: MIT.
    \item GeneCIS: CC BY-NC 4.0.
    \item FLUX.1-dev: FLUX.1 [dev] Non-Commercial License v1.1.1.
    \item FLUX-Reason-6M: Apache 2.0.
    \item GEditBench v2: MIT.
    \item Mip-NeRF~360: Apache 2.0.
    \item NerfStudio: Apache 2.0.
    \item IC-Light: Apache 2.0.
    \item DreamLight: CC BY 4.0.
    \item Diff-Harmonization: Apache 2.0.
    \item TRELLIS.2: MIT.
    \item InstantMesh: Apache 2.0.
    \item Hunyuan3D-2.1: Tencent Hunyuan 3D 2.1 Community License.
    \item Step1X-3D: Apache 2.0.

\end{itemize}

\subsection{Change log}
\myparagraph{v1} Initial release.

\myparagraph{v2} Added generalization section on additional benchmarks in Appendix~\ref{sec:supp_generalization} and Table~\ref{tab:genecis}. Updated numbers in Table~\ref{tab:qare_main} on the QARE benchmark. Reframed the vision-algorithm 2AFC set as an \emph{external-algorithm} generalization set, and clarified its motivation. Added a data diversity analysis in Figure~\ref{fig:data_diversity}. Moved fusion-strategy implementation details to Appendix~\ref{sec:supp_implementation}. Discussed the concept-entanglement limitation in Figure~\ref{fig:limitation}. Minor changes otherwise.

\clearpage

\clearpage

\end{document}